\documentclass[letterpaper,12pt]{article}
\usepackage{amsfonts}
\usepackage{amssymb, amsmath, amsthm,  graphicx}
\usepackage[margin=1in]{geometry}
\usepackage[bottom]{footmisc}
\usepackage{commath}
\usepackage[utf8]{inputenc}
\usepackage{natbib}
\usepackage{booktabs, multirow}
\usepackage{url}
\usepackage[normalem]{ulem}
\usepackage{lscape}
\useunder{\uline}{\ul}{}
\usepackage{accents}
\usepackage[flushleft]{threeparttable}
\usepackage{adjustbox}
\usepackage{tikz}
\usepackage{bigstrut}
\usepackage{setspace}
\graphicspath{ {figures/} }
\usepackage{subfig}
\usepackage{enumitem}
\usepackage{float}
\usepackage{placeins}
\usepackage{setspace}
\DeclareMathOperator*{\argmin}{argmin}

\newtheorem{thm}{Theorem}

\newtheorem{lemma}{Lemma}
\newtheorem{corollary}{Corollary}
\newtheorem{assum}{Assumption}
\newtheorem{assumwfactors}{Assumption}

\newcommand{\1}{\mbox{$\mathrm{1\hspace*{-2.5pt}l}$\,}}

\newcommand{\blind}{0}

\begin{document}

\if0\blind
{
\title{Deep Learning Based Residuals in Nonlinear Factor Models: Precision Matrix Estimation of Returns with Low Signal-To-Noise Ratio}
\author{\textsc{Mehmet Caner\thanks{
			North Carolina State University, Nelson Hall, Department of Economics, NC 27695. Email: mcaner@ncsu.edu. }}
	\and \textsc {Maurizio Daniele%
		\thanks{ETH Zürich, KOF Swiss Economic Institute, 8092 Zurich, Switzerland. Email: daniele@kof.ethz.ch}}
}

\date{\today}

\maketitle
} \fi

\if1\blind
{
\title{Deep Learning Based Residuals in non-linear Factor Models: Precision Matrix Estimation of Returns with Very Low Signal-To-Noise Ratio}
\date{\today}

\maketitle
} \fi

\bigskip
\begin{abstract}

\noindent 

This paper introduces a consistent estimator and rate of convergence for the precision matrix of asset returns in large portfolios using a non-linear factor model within the deep learning framework. Our estimator remains valid even in low signal-to-noise ratio environments typical for financial markets and is compatible with weak factors. 
Our theoretical analysis establishes uniform bounds on expected estimation risk based on deep neural networks for an expanding number of assets. Additionally, we provide a new consistent data-dependent estimator of error covariance in deep neural networks. 
Our models demonstrate superior accuracy in extensive simulations and the empirics.\\

\end{abstract}

\newpage

\section{Introduction} \label{sec_intro}

The Great Financial Crisis 2008/09 and the COVID-19 Crisis 2020/21 revealed major problems and disadvantages of existing econometric models for economic forecasting and policy analysis. Especially during periods with high uncertainties these models commonly cause large prediction errors and lead to misleading policy recommendations. Recent research developments indicate that machine learning methods suit better in these circumstances: they can measure non-linear structures and sudden changes in the relations between economic variables. Deep neural networks, i.e.,\ artificial neural networks with many hidden layers, as part of the most important machine learning techniques nowadays, received increasing attention in economics over the recent years. Their usefulness has been shown on several complex machine learning problems that concern e.g.,\ natural language processing and image recognition. A detailed overview of applications of deep neural networks on observed data can be found e.g., in \cite{Schmidhuber2015} and the literature cited therein.  In an comprehensive empirical paper, \cite{gkx2020} compare the out-of-sample performance of various machine learning techniques in predicting the portfolio returns for the US market and show that most often neural network methods provide the best results.
More recently, \cite{Gu2021} develop a non-linear conditional asset pricing model based on an extended autoencoder incorporating latent factors that depending on asset characteristics and illustrate its superiority in terms of the lowest pricing errors compared to standard methods commonly used in the finance literature.

In this paper, we provide a consistent estimator of the precision matrix of asset returns in large portfolios based on deep learning residuals formed from non-linear factor models, so that we can get better out-of-sample results for metrics like Sharpe-Ratio, variance of the portfolio.
In light of the advantages of deep neural networks in terms of prediction accuracy compared to traditional econometric models testified in various empirical studies and research fields, there is an increasing interest in analyzing their large sample properties. Important theoretical contributions that deal with the approximation properties of deep neural networks in the nonparametric regression framework are provided e.g.,\ by \cite{sh2020}, \cite{Farrell2021} and \cite{kl2021}. For a general class of smooth non-linear functions \cite{Farrell2021} make a major contribution and provide upper risk bounds for deep learning estimators.
Within a subcase for the class of composition based functions, which include (generalized) additive models, \cite{sh2020} shows that sparse deep neural networks do not suffer from the curse of dimensionality in contrast to traditional nonparametric estimation methods and achieve minimax rate of convergence in expected risk under the squared loss. Recently, \cite{jslh2023} provide risk bounds on deep learning linear function estimators. Notably, their derived constant on the risk bounds is smaller compared to existing literature, enabling the establishment of bounds for approximately sparse factor structures. Their analysis relies on a rather abstract treatment of sparsity-approximate factors and 
finds more practical utility in domains like image processing and video analysis, where low-dimensional latent structures are prevalent.

	
One of the main disadvantages of machine learning techniques is that the typical application assumes  a high signal-to-noise ratio. However, in asset pricing this is not the case. Table 3.1 of \cite{n2021} clearly illustrates that typical datasets on asset returns exhibit a low or very low signal-to-noise ratio. In our paper, we introduce a model in the framework of deep neural networks, which is also appropriate in a setting with very low signal-to-noise ratio and can be used to obtain a consistent estimator of the precision matrix of the returns, even with a growing number of assets in the portfolio.
In order to establish this estimator, we investigate the theoretical properties of feedforward multilayer neural networks (multilayer perceptron) with rectifier linear unit (ReLu) activation function. Our modeling framework builds upon the research contributions of \cite{Farrell2021} and \cite{sh2020} offers convenient theoretical generalizations and extensions. 

Specifically, we contribute to the deep learning literature with uniform results on the expected estimation risk. In our first contribution, we analyze the properties of the deep neural network estimator in case of incorporating a potential set of $J$ response variables compared to the current deep learning research that refers to univariate responses. Our results provide bounds on the expected risk and show that these upper bounds are uniform over the $J$ variables. Hence, our model framework offers considerable extensions in terms of functional composition of the true underlying model and estimation. In fact, we allow for different unknown functional forms in the nonparametric regression model across each variable $J$.
The advantage of concentrating on the theoretical framework of \cite{Farrell2021} lies in its straightforward applicability to factor models. In fact, we further contribute by building a bridge between our deep learning setting and non-linear factor models. One of the key issues in financial markets are the non-linearities due to interaction terms involving the covariates or factors explaining asset returns, as discussed in Section 3.6 of \cite{n2021}. Our deep neural network factor model (DNN-FM) estimator and its theory can handle this type of non-linearities, which is also confirmed by the superior performance in extensive simulations and the empirical application.

Moreover, we extend our deep learning results to an additive model setting, similar to the framework in \cite{sh2020} and analyze the implications of imposing sparsity in the neural networks. It is important to note that our sparse deep neural network only incorporates a sparse structure in the parameters in the hidden layers. Hence, the input variables (factors) do not exhibit any sparsity. We provide the theory for our sparse deep neural network factor model (SDNN-FM) estimator, which shows that the SDNN-FM approach can avoid the curse of dimensionality arising from an increasing number of included factors and testify its strong performance compared to alternative machine learning techniques in simulations and our out-of-sample portfolio forecasting exercise.
A non-linear additive structure for explaining asset returns is also used in \cite{fnw2020}. The authors propose an adaptive group lasso structure and show in an empirical study based on data of the US stock market that the return to risk ratio of a portfolio constructed by a non-linear factor model can be considerably higher compared to the one resulting from a linear factor model.

In order to obtain an estimate of the data covariance matrix based on the deep neural network factor model (DNN-FM) and its sparse version (SDNN-FM), we require an efficient estimate for the residual covariance matrix. For this purpose, as our second contribution, we develop a novel data-dependent covariance matrix estimator of the innovations in deep neural networks. The estimator refers to a flexible adaptive thresholding technique, which is robust to outliers in the errors. We elaborate on the consistency of the estimator in $l_2$-norm under rather mild assumptions. 
Furthermore, since the precision matrix of the returns of a high-dimensional portfolio is paramount for understanding various financial metrics, such as the Sharpe ratio, we provide consistency and rate of convergence results for the precision matrix estimated based on our DNN-FM and SDNN-FM. As our third contribution, we show that we obtain these consistency and rate of convergence results under a very low signal-to-noise ratio assumption. Specifically, we allow that the minimum eigenvalue of the covariance matrix of the non-linear factors is zero or local to zero and show its connection to a very low signal-to-noise ratio in the data. Furthermore, instead of using the Sherman-Morrison-Woodbury formula, which is commonly applied in the literature to obtain the precision matrix of the returns, we provide a new formula that uses matrix theory and is in line with a setting where the minimum eigenvalue of the covariance matrix of the factors is zero or local to zero. Moreover, we illustrate the direct connection between the low signal-to-noise ratio assumption and the weak factor setting. Hence, our deep neural network framework is suitable for measuring weakly influential factors.

Linear factor models are helpful in understanding the behavior of asset returns. \cite{fan2011} and \cite{fan2013} use linear observed and unobserved factors in a large portfolio of assets, respectively. They show that linear factor models can be combined with a sparse error covariance matrix to consistently estimate the precision matrix of asset returns even in high dimensions. These two papers are benchmarks and contributed to merging of factor model with high dimensional statistics literature in an important way. \cite{ffx2016}  show that in case of "almost" block diagonal covariance matrix of errors, there can be a sparse precision matrix of the errors, which is applicable in linear factor models. Recently, \cite{caner2019} and \cite{caner2022} estimate the precision matrix of returns via nodewise regression and with linear factor model based residual nodewise regression techniques, respectively. \cite{caner2023} proposed a new machine learning technique that may be suitable for buy and sell decisions for a single asset in a high dimensional generalized linear model with structured sparsity patterns.

Traditional factor models, as in \cite{Fama1993} or \cite{fan2011} assume rigid linear relations between the factors and the underlying observed variables. Especially during crisis periods, however, economic time series follow non-linear relations and are subject to sudden changes. Hence, the linearity assumption of standard factor models would be inappropriate to measure time series with these patterns. The structure of our deep neural network factor model (DNN-FM)  and its sparse version (SDNN-FM) mitigate this limitation and allow for measuring non-linear and complex dynamic relations between the factors and economic variables. Our theoretical elaboration provides convergence results for the covariance and precision matrix estimators based on the deep neural network. In that general framework the rate of convergence of our estimators are not affected by number of assets in a portfolio unless that number is exponentially growing. Moreover, the convergence results depend on both the number of factors and the smoothness of the functions.
In the subcase of non-linear additive functions we show that the convergence rate is not affected by the number of included factors. 

The favorable large sample properties of the DNN-FM and SDNN-FM are confirmed by our Monte Carlo study based on various simulation designs. Specifically, both approaches consistently determine the true underlying function connecting the factors and observable variables, as the number of periods increases. Further, the estimators for the covariance and precision matrix of the returns are as well consistent. Compared to competing approaches, as e.g.,\ the traditional static factor model, which are sensitive to an increase of the number of factors $d$ (i.e.,\ their error rate deteriorate as $d$ increases), the convergence rates of the DNN-FM and SDNN-FM estimators are stable with respect to a raising number of factors.

\begin{figure}[!t]
	\begin{minipage}{.45\textwidth}
		\centering
		\captionsetup{width=1\linewidth}
		\subfloat[First sample and $J = 50$]{\includegraphics[width=1\linewidth]{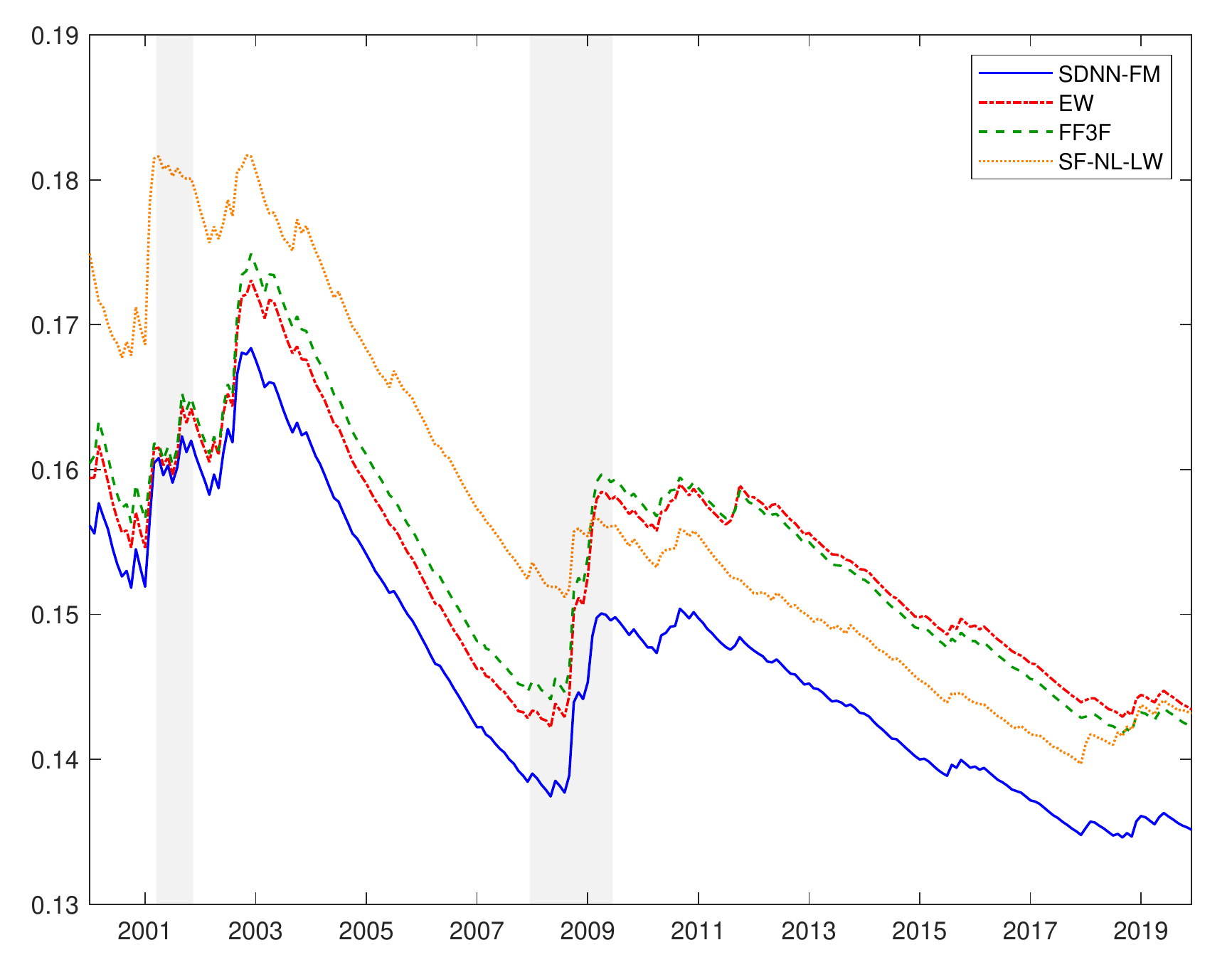} \label{fig_sd_sub_a}}
	\end{minipage}\hspace*{0.75cm}
	\begin{minipage}{.45\textwidth}
		\centering
		\captionsetup{width=1\linewidth}
		\subfloat[First sample and $J = 100$]{\includegraphics[width=1\linewidth]{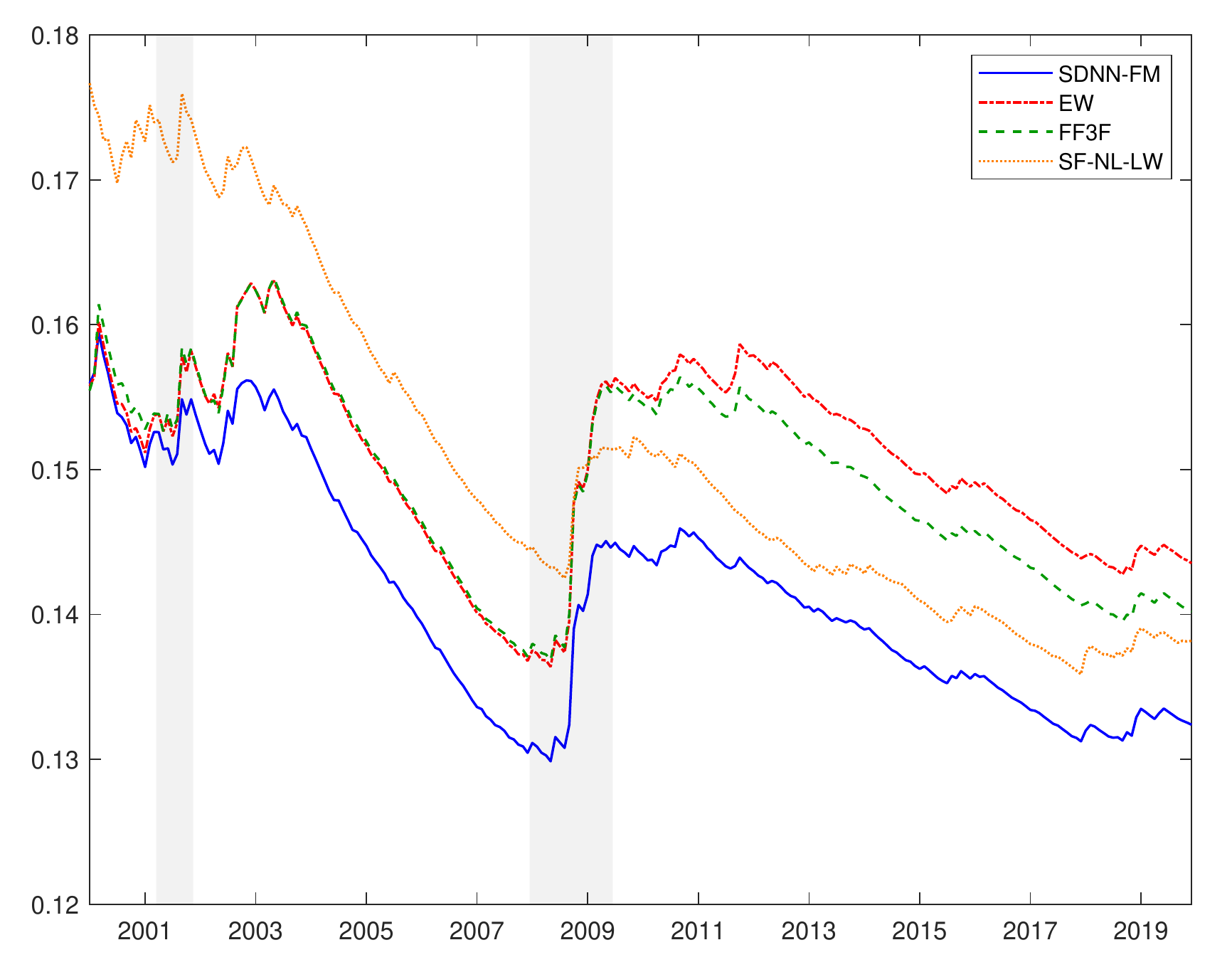}\label{fig_sd_sub_b}}
	\end{minipage}\\\vspace{0.25cm}
	
	\begin{minipage}{.45\textwidth}
		\centering
		\captionsetup{width=1\linewidth}
		\subfloat[Second sample and $J = 50$]{\includegraphics[width=1\linewidth]{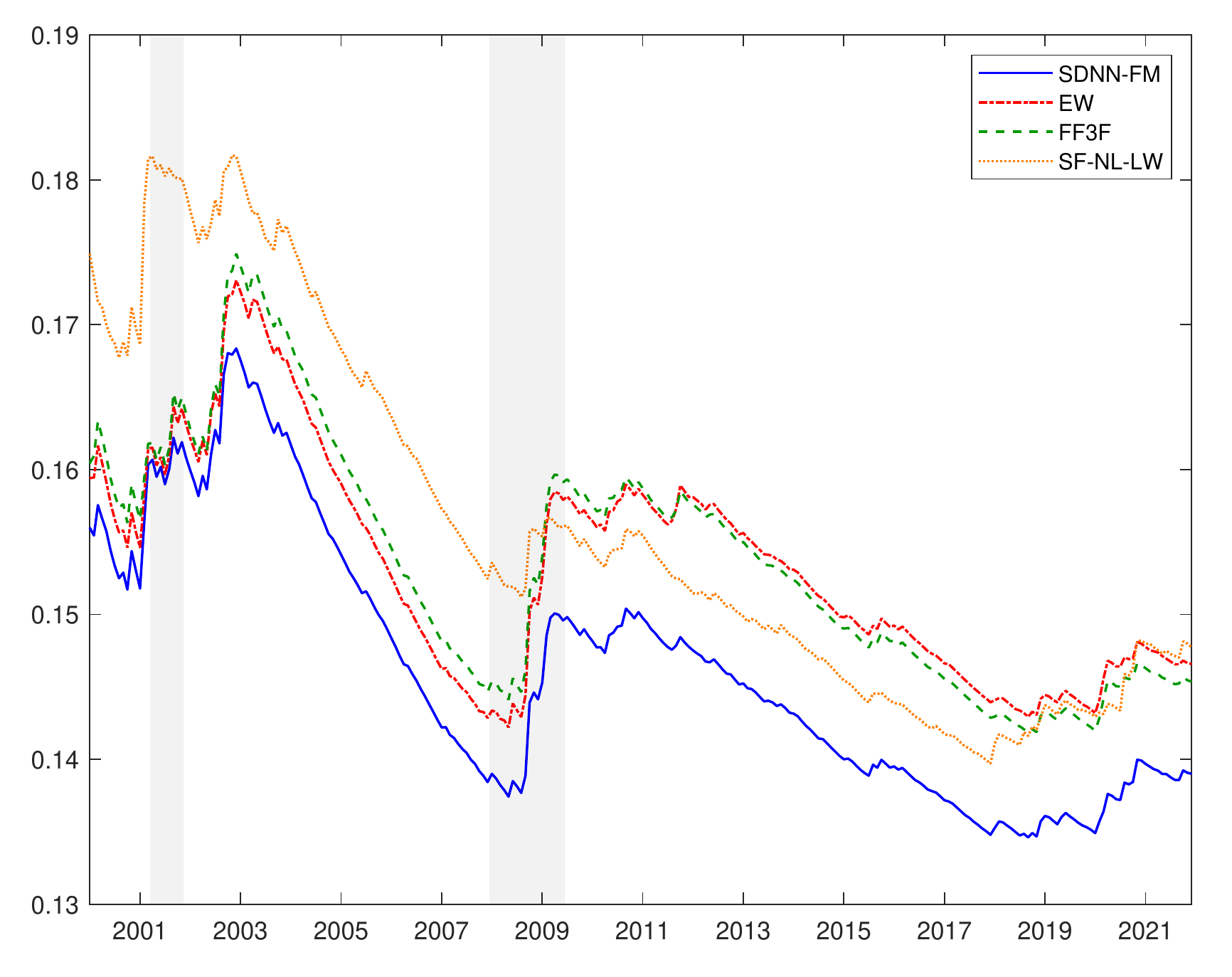}\label{fig_sd_sub_c}}
	\end{minipage}\hspace*{0.75cm}
	\begin{minipage}{.45\textwidth}
		\centering
		\captionsetup{width=1\linewidth}
		\subfloat[Second sample and $J = 100$]{\includegraphics[width=1\linewidth]{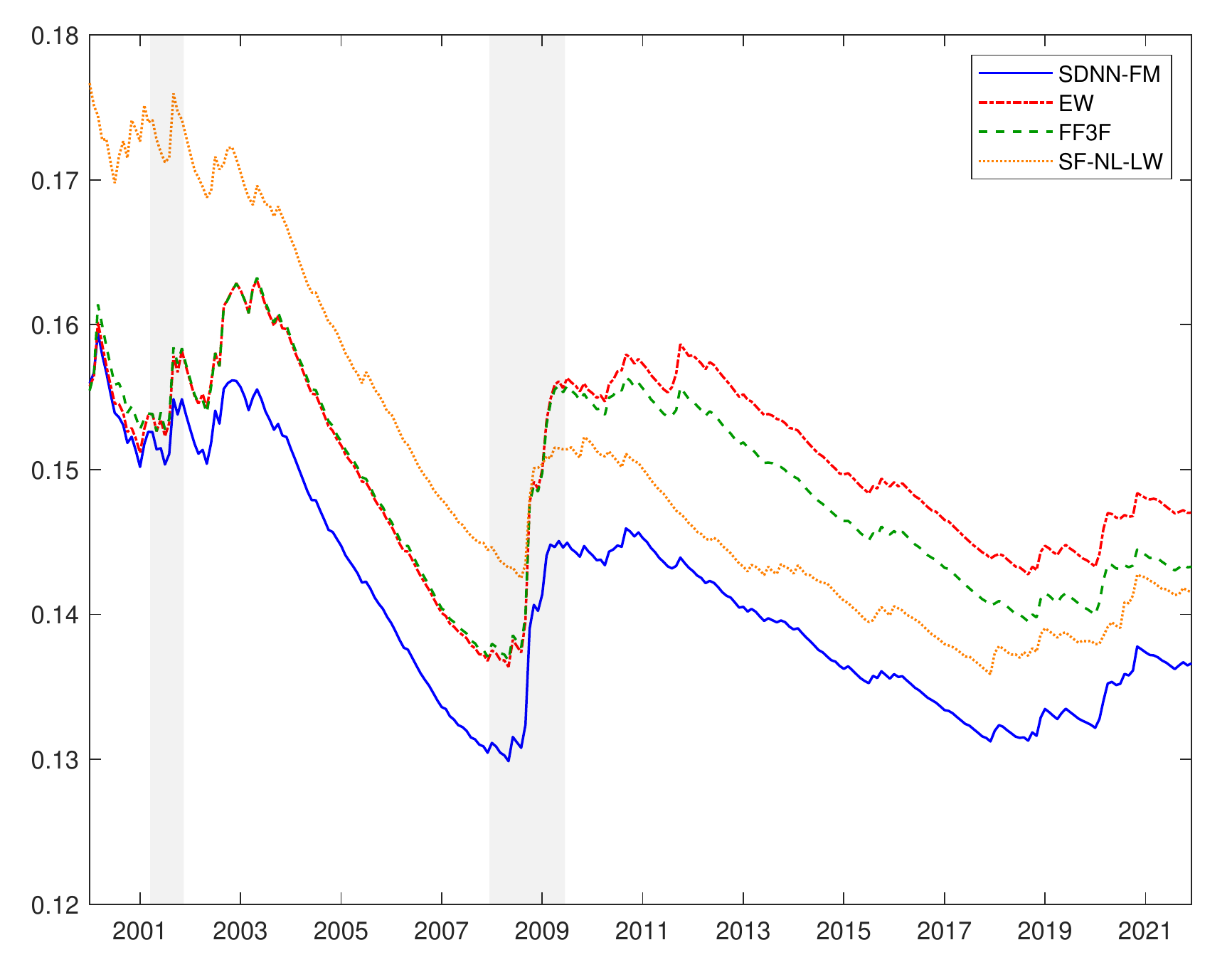}\label{fig_sd_sub_d}}
	\end{minipage}
	\caption{SD for different subperiods}\label{fig_sd_sub}
\end{figure}


In an out-of-sample portfolio forecasting application based on assets constituents of the S\&P 500 stock index and concentrating on a global minimum variance portfolio setting, we show that the DNN-FM and SDNN-FM are superior in most cases compared to competing portfolio estimators that are commonly used in the literature. More precisely, they often lead to the lowest out-of-sample portfolio standard deviation (SD) and avoid large changes in the portfolio constellation. Consequently, both models provide low portfolio turnover rates and prevent high transaction costs. This generally leads to the highest out-of-sample Sharpe rations across different asset spaces compared to the competing methods when transaction costs are taken into account. The superiority in forecasting precision is particularly pronounced during turbulent times, such as during the Great Financial Crisis of 2008/09 and the COVID-19 Crisis. Hence, the flexibility of the deep learning estimators allows for capturing strong changes and high uncertainties in financial time series during volatile periods.

As an example, we illustrate the results of a comparison among our SDNN-FM, the equally weighted portfolio (EW), the Fama-French three factor (FF3F) model, and the single factor non-linear shrinkage method of \cite{lw2017} (SF-NL-LW) in Figure \ref{fig_sd_sub}. The y-axis shows the out-of-sample standard deviation (SD) of a large portfolio, and the x-axis denotes the out-of-sample periods that we analyze. We consider two sample periods, where the first sample excludes the COVID-19 Crisis and the second one includes the COVID-19 Crisis. $J$ represents the number of assets in the portfolio, and the NBER recession years for the dot-com Crisis 2001 and the Great Financial Crisis of 2008/09 are highlighted as vertical gray bars.
The results are depicted in Figure \ref{fig_sd_sub}, where panels \subref{fig_sd_sub_a} and \subref{fig_sd_sub_b} refer to the first sample without COVID-19 Crisis and panels \subref{fig_sd_sub_c} and \subref{fig_sd_sub_d} correspond to the second sample including the COVID-19 Crisis. The outcome at a specific period $i$ incorporates the out-of-sample returns until $i$ (e.g.,\ the SD in December 2011 incorporates the out-of-sample returns from January 1996 until December 2011). The graphs illustrate that the advantage of the SDNN-FM in lowest out-of-sample portfolio standard deviation compared to the competing approaches are especially pronounced during and after the Great Financial Crisis 2008/09 and the COVID-19 Crisis 2020/21. These results confirm that our deep learning method is well suited for capturing high volatilities during turbulent episodes. Extensive results and more comparisons are given in Section \ref{sec_pf}.

Recent related papers merging the deep learning literature with asset pricing applications offer substantial advances in the empirical literature. \cite{fhpx22} analyze latent factors based on a structural model that benefits from deep learning estimation. Their well-designed model establishes connections between factor loadings, factors, and the deep learning literature. The analysis encompasses the asset returns and pricing errors. Notably, they adopt a sparse approach for selecting weights in deep neural networks, akin to our SDNN-FM model. Their asset return model is linear in factors and is empirically estimated, incorporating deep learning generated factors. Remarkably, their empirical findings demonstrate that  factors generated based on deep learning techniques lead to enhanced Sharpe Ratios in portfolios. \cite{cpz2021} utilize the no-arbitrage condition to estimate stochastic discount factor (SDF) weights and the optimal instrument function. They employ recurrent neural networks, including Long-Short-Term-Memory estimates, to obtain hidden macro variable estimates. Subsequently, feed-forward neural networks are employed to estimate SDF and the optimal function of hidden macro variables and firm characteristics. Their study employs a generative adversarial network structure, analyzing the problem from both structural and empirical perspectives. In contrast, our research focuses on the econometric theory, specifically exploring the integration of non-linear factors with deep neural networks. Moreover, one objective is to consistently estimate the precision matrix of asset returns and analyze the rate of convergence. 


The remainder of the paper is organized as follows. In Section \ref{sec_model}, we introduce the  deep neural network framework used for estimating nonparametric regressions and provide our main theoretical findings. In Section \ref{sec_cov_e_dnn_f}, we introduce a novel estimator for the covariance and precision matrix for the idiosyncratic errors of the deep neural network factor model by means of a robust adaptive threshold estimator and prove its consistency. Moreover, we elaborate on the consistency of the corresponding covariance matrix estimator of the returns in Section \ref{cvmret}. Section \ref{pmret} introduces the precision matrix estimator of the returns based on the DNN-FM and shows the consistency of the estimator. Section 6 provides the non-linear additive function case in a sparse deep neural network.
Details on the implementation are discussed in Section \ref{sec:impl}. In Section \ref{sec_sim}, we present Monte-Carlo evidence on the finite sample properties of our DNN-FM and SDNN-FM approaches in estimating the unknown function, which connects the factors with the observable variables, as well as the corresponding covariance and precision matrix. In Section \ref{sec_pf} we analyze the empirical performance of our methods in an out-of-sample portfolio forecasting exercise. Section \ref{sec_conclusions} summarizes the main findings. The proofs are provided in the Appendix.


\section{Deep Learning}\label{sec_model}

In this section, we combine the work of \cite{Farrell2021} with non-linear factor models. We extend their results to  multivariate output variables.
Assume the following nonparametric regression model for $j=1,\cdots, J$
\begin{equation}
Y_{j,i} = f_{0,j} (X_i) + u_{j,i},\label{mo}
 \end{equation}
where $Y_{j,i}$ denotes the $j$-th response variable for the $i$-th observation. The regressors $X_i$ are iid across $i=1,\cdots,n$, and incorporate observed variables of dimension $d \times 1$, where each $X_i \in [-1,1]^d$.
\footnote{Currently, we are working on a paper, which extends the modeling framework by relaxing the iid assumption on the regressors $X_i$ and allowing for a time series structure.}
Also $d$ is not growing with $n$, it is a positive integer-constant and $ d \ge 1$.
Let $u_{j,i}$ be a zero mean noise component, which is iid across $i=1,\cdots,n$, for all $j=1,\cdots,J$. Define the $J \times J$-dimensional covariance matrix of the errors as $\Sigma_u:= E u_i u_i'$, where $u_i:=(u_{1,i},\cdots,u_{j,i},\cdots, u_{J,i})'$ is a vector of dimension $J \times 1$.
$f_{0,j}(.)$ is an unknown function, and for each $j=1, \cdots, J$, it may be of a different functional form, but $f_{0,j} (.): [-1,1]^d \to R$. Moreover, we assume that $u_{j,i}$ is independent of $X_i$ for a given $j$. $J$ can be a nondecreasing function of $n$.To save from notation, we do not subscript $J$ with $n$. Also, the number of factors $d$ is constant, and does not vary with $n$.

 
                                               
\subsection{Multilayer Neural Networks}

In order to estimate the unknown function $f_{0,j}(.)$, we rely on multilayer (deep) feedforward neural networks.
\footnote{The "Deepness" in the neural network will arise from multiple hidden layers to estimate $f_{0,j}(.)$, for $j=1,\cdots, J$. "Learning" comes from estimating and correcting errors in the network parameters via an algorithm (stochastic gradient descent), which will be relegated to the literature, and will not be discussed here. For further information see e.g.,\ \cite{gbc2016}.} We are concerned about the statistical properties of deep learning, specifically multilayer feedforward neural networks. In the following, we briefly describe the model architecture. As it is common in nonparametric problems, we start with the regressors, $X_i$ and aim to determine the response $Y_{j,i}$. In order to achieve that we have to estimate the unknown function $f_{0,j}(.)$ in \eqref{mo} connecting $X_i$ with $Y_{j,i}$. In the deep learning framework, the estimation of $f_{0,j}(.)$ is carried out within $L$ hidden layers, which are incorporated between the input layer containing the regressors and the output layer that includes the response. Let $L\ge 1$, and represents the number of hidden layers. A given hidden layer is composed of hidden units $p_{l,j}$, which are defined for each layer $l$, and each function $j$, with $l=1, \cdots, L$ and $j=1,\cdots, J$. These units are transformed by an activation function: $\sigma: R \to R$ at each layer. We impose a rectifier linear unit (ReLu) activation function for $\sigma$
\[ \sigma (x)= \max (x,0),\]
which censors all negative real numbers to zero, and keeps all positive. The choice for this activation function is motivated by several factors. First, the partial derivative of the ReLu function is either zero or one, which facilitates computations. Moreover, the ReLu activation function can pass a signal without any change through all layers, i.e.\ it has the projection property. For each $j=1,\cdots, J$, we assume the same number of layers $L$. 


We formally define the network architecture by the parameters $(L,p_j)$, where the number of hidden layers $L$ represents the depth of the network, and $p_j$ corresponds to the number of units at each layer $j=1, \cdots, J$, which constitutes the width of the network. Note that we allow for a different width vector for the estimator of each function $j$.
Specifically, $p_j=(p_0,p_{1j}, \cdots, p_{Lj}, p_{L+1})'$, which is a $L+2$ vector, where $p_0$ is the number of inputs and $p_{L+1}$ defines the number of units in the output layer. The remaining quantities $p_{1j}, \cdots, p_{Lj}$, represent the number of units in hidden layers $1,\cdots, L$, respectively. In this paper, we impose one output variable, hence $p_{L+1}=1$, and the number of inputs is $p_0=d$ for each $f_j(.)$. 

\begin{figure}[!t]
    \begin{center}
		\begin{tikzpicture}
			\node[font=\scriptsize,circle,fill=green,draw=green,inner
			sep=0pt,minimum size=4mm] (1a) at (0,0) {} ;
			\node[font=\scriptsize,circle,fill=green,draw=green,inner
			sep=0pt,minimum size=4mm, label={[label distance=18pt]above:Input layer}] (1b) at (0,1) {} ;
			
			\node[font=\scriptsize,circle,fill=blue,draw=blue,inner
			sep=0pt,minimum size=4mm] (2a) at (2,-0.5) {} ;
			\node[font=\scriptsize,circle,fill=blue,draw=blue,inner
			sep=0pt,minimum size=4mm] (2b) at (2,0.5) {} ;
			\node[font=\scriptsize,circle,fill=blue,draw=blue,inner
			sep=0pt,minimum size=4mm] (2c) at (2,1.5) {} ;
			
			\node[font=\scriptsize,circle,fill=blue,draw=blue,inner
			sep=0pt,minimum size=4mm] (3a) at (4,-0.5) {} ;
			\node[font=\scriptsize,circle,fill=blue,draw=blue,inner
			sep=0pt,minimum size=4mm] (3b) at (4,0.5) {} ;
			\node[font=\scriptsize,circle,fill=blue,draw=blue,inner
			sep=0pt,minimum size=4mm] (3c) at (4,1.5) {} ;
			
			\node[font=\scriptsize,circle,fill=red,draw=red,inner
			sep=0pt,minimum size=4mm, label={[label distance=32pt]above:Output layer}] (4a) at (6,0.5) {} ;
			
			\node[inner sep=0pt] (5a) at (-1,0) {$x_2$} ;
			\node[inner sep=0pt] (5b) at (-1,1) {$x_1$} ;
			
			\node[inner sep=0pt] (6a) at (7,0.5) {$y$} ;
			
			\draw [-latex, thick, gray] (1a) -- (2a) {};
			\draw [-latex, thick, gray] (1a) -- (2b) {};
			\draw [-latex, thick, gray] (1a) -- (2c) {};
			\draw [-latex, thick, gray] (1b) -- (2a) {};
			\draw [-latex, thick, gray] (1b) -- (2b) {};
			\draw [-latex, thick, gray] (1b) -- (2c) {};
			
			\draw [-latex, thick, gray] (2a) -- (3a) {};
			\draw [-latex, thick, gray] (2a) -- (3b) {};
			\draw [-latex, thick, gray] (2a) -- (3c) {};
			\draw [-latex, thick, gray] (2b) -- (3a) {};
			\draw [-latex, thick, gray] (2b) -- (3b) {};
			\draw [-latex, thick, gray] (2b) -- (3c) {};
			\draw [-latex, thick, gray] (2c) -- (3a) {};
			\draw [-latex, thick, gray] (2c) -- (3b) {};
			\draw [-latex, thick, gray] (2c) -- (3c) {} node[midway, label={[label distance=5pt]above:\color{black}Hidden layers}] {};
			
			\draw [-latex, thick, gray] (3a) -- (4a) {};
			\draw [-latex, thick, gray] (3b) -- (4a) {};
			\draw [-latex, thick, gray] (3c) -- (4a) {};
			
			\draw [-latex, thick, gray] (5a) -- (1a) {};
			\draw [-latex, thick, gray] (5b) -- (1b) {};
			\draw [-latex, thick, gray] (4a) -- (6a) {};

		\end{tikzpicture}
	\end{center}
	\caption{Graphical representation of a multilayer feedforward neural network with two hidden layers, two regressors, one response and three units at each hidden layer.}
	\label{fig_ffnnet}
\end{figure}
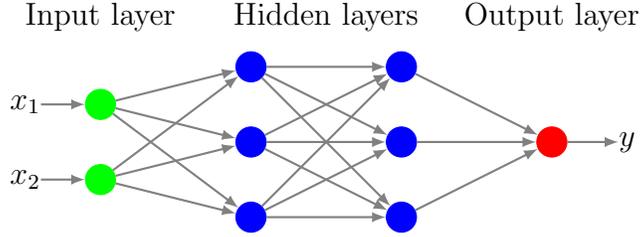


In the following, we provide a simple example of a multilayer feedforward neural network illustrated by the directed acyclic graph in Figure \ref{fig_ffnnet}. The neural network incorporates two regressors $d=2$ and one response $J=1$. Moreover, it contains two hidden layers ($L=2$) with three units at each hidden layer, i.e.,\ $p_{1,1}=p_{2,1}=3$.

The first connection arises from the regressors, which are collected in the input layer to the first hidden layer. Each regressor is connected forward with each unit in the first hidden layer. Moreover, at each hidden unit the regressors are multiplied with weights and an intercept is added. Hence, for the first unit in the first hidden layer three parameters need to be estimated. The remaining units follow an identical structure. In all this, the ReLU activation function is used to estimate the parameters at each unit. In our example, nine parameters have to be estimated in total, and three ReLU units are in the first hidden layer.

In a second step, each unit in the first hidden layer is connected with each unit in the second hidden layer. More specifically, the outputs from the first hidden layer are used as inputs to each unit at the second hidden layer. Each of those inputs is multiplied with the corresponding weights, an intercept is added and the the ReLU activation function is applied. For our example, we need to estimate 12 parameters, formed by three weights and an intercept at each unit in second hidden layer.

Finally, in the output layer, we use the least squares loss and regress $Y_i$ on all three responses from the second hidden layer and an intercept. This adds four additional parameters that have to be estimated. Hence, in total we need to estimate 25 parameters, if the regressor vector $X_i$ is $2 \times 1$-dimensional and if we incorporate two hidden layers with three units each.
For a general $(L, p_j)$ multilayer deep neural network architecture, let $W_j$ denote the total number of parameters to be estimated for each $j=1, \cdots, J$. A straightforward calculation shows that $W_j = \sum_{l=0}^L (p_{l,j} + 1) p_{l+1,j}$. 

\subsection{Estimator}

Let ${\cal F}_j (L, p_j)$ represent the multilayer perceptron architecture. To save from notation let ${\cal F}_j:= {\cal F}_j (L, p_j)$.
We define the estimator as the empirical minimizer, $\hat{f}_j $ representing the feedforward deep neural network estimator described above with unbounded weights and without any sparsity constraint on the weights, as follows 

\begin{equation}
	\hat{f}_j:= \argmin_{f_j  \in {\cal F}_j, \\ \| f_j \|_{\infty} \le 2 M_j} \sum_{i=1}^n (Y_{j,i} - f_j (X_i))^2, \label{dnn_est}
\end{equation}
where $M_j >0$ is a positive constant for each $j=1,\cdots, J$. This type of estimator is a subcase of estimators with general loss functions, see \cite{Farrell2021}. In the following, we formalize an assumption about the data. 
\begin{assum}\label{as1}
Assume that for each $ j =1,\cdots, J$, $(Y_{j,i}, X_i')'$, $ 1\le i \le n$ are iid across $i=1,\cdots, n$. $X_i$ is continuously distributed. We set 
$\| f_{0,j} \|_{\infty} \le M_j$, for $j=1,\cdots, J$, ${\cal Y}_j \subset [-M_j, M_j]$, where $M_j>0$ is a positive constant for each $j$.
\end{assum}

This assumption is a minor extension of \textit{Assumption 1} in \cite{Farrell2021} to a case with many functions (outcomes), and hence we allow for bounded outcomes but with differing bounds for each $j=1,\cdots, J$. In Section \ref{sec_factor_model}, we relax this restriction. The restriction on bounded outcomes is necessary to have Lipschitz continuity of the least squares loss and to be able to apply \textit{Lemma 2} and \textit{Lemma 5} of \cite{Farrell2021} in the proofs.
For the second assumption, we need the following definition.
The formal definition of the 
Sobolev ball
 \[{\cal W}^{\beta,\infty} ([-1,1]^d):= \{ f: \max_{\alpha, | \alpha | \le \beta} ess sup_{x \in [-1,1]^d} | D^{\alpha} f(x) | \le 1\},\] where \[\alpha=(\alpha_1, \cdots, \alpha_d)',\] and $D^{\alpha} f$ is the weak derivative, and $ess sup$ is the essential supremum, and $| \alpha| = \alpha_1 + \cdots, \alpha_d$.

\begin{assum}\label{as2}
For each $j=1,\cdots J$, let $f_{0,j} (X_i)   \in {\cal W}^{\beta} ( [-1,1]^d)$, where each function has the same smoothness parameter $\beta \in {\cal N_{+}}$.
\end{assum}

Assumptions \ref{as1} and \ref{as2} are standard and used in \cite{Farrell2021}. The width of all layers will grow at the same rate for a given $j$. Hence, the number of parameters to be estimated in the deep neural network only changes up to constants when we estimate $j=1,\cdots, J$ functions. Theorem \ref{t1} provides risk bounds for a specification with multiple outcome variables in the deep learning literature.

\begin{thm}\label{t1}
Under Assumptions \ref{as1} and \ref{as2}, where the number of hidden units at each layer grow at the same rate $p_{l, j} \asymp n^{d/2(\beta+d)}(\log n)^2$, for $l=1,\cdots, L, j=1,\cdots J$, and the depth of the neural network is $L \asymp \log n$, we obtain

\begin{enumerate}[label=(\roman*)]
\item For $n$ large enough
\[ P\left\{  \max_{1 \le j \le J} \left[ \frac{1}{n} \sum_{i=1}^n [ \hat{f}_j (X_i) - f_{0,j} (X_i)]^2 \right] \le r_{n1} + r_{n2}
\right\} \ge 1 - \exp (- n^{d/(\beta +d)} (\log n)^8),\]
with $r_{n1} := 2 C_1[n^{-\beta/(\beta+d)} (\log n)^8 + \frac{\log (\log n)}{n}]$ and $r_{n2}:= C_1[\frac{\log J }{n}]$, where $C_1 >0$ is a positive constant. \label{t1_i}

\item With $n \to \infty$, \ref{t1_i} holds with probability approaching one.

\end{enumerate}
\end{thm}

\noindent Remarks on Theorem \ref{t1}. 
\begin{enumerate}
	\item In \textit{Theorem 1} of \cite{Farrell2021} the rate is solely $r_{n1}$ since they have a specification with a single outcome. Our theorem offers an extension of \cite{Farrell2021} to multiple outcomes, which leads to the additional rate $r_{n2}$.
	
	\item Define the quantity
	\begin{equation}
		l_n:=\max(r_{n1}, r_{n2}).\label{rln}
	\end{equation}

	In the following, we analyze the rates in Theorem \ref{t1} \ref{t1_i}. We start with considering a exponentially growing number of outcomes, i.e.,\ $J = \exp (n^a)$ and set $r_{n2} = O (n^{a-1})$. If $1> a> \frac{d}{\beta+d}>0$, then we have $l_n=\max(r_{n1}, r_{n2}) =r_{n2}= O (n^{a-1})$.
	If we still consider $J = \exp(n^a)$, however $ 1>  \frac{d}{\beta + d} \ge a>0$, we get $l_n= r_{n1} = O (n^{-\beta/(\beta+d)} (\log n)^8)$.
	Hence, we obtain two different cases when the number of outcomes is growing exponentially in $n$.
	The first case leads to the rate $r_{n2}$, when the number of regressors $d$ is small. The second case yields with a larger (but constant) number of factors the rate $r_{n1}$. When $J= n^b$, $b>0$, then $l_n = r_{n1}$. We obtain the same rate $r_{n1}$, if $J= b n$. Moreover, when $J$ is a fixed number and not growing with $n$, the  rate is $r_{n1}$.


\end{enumerate}

In order to simplify the interpretation in the subsequent technical elaboration and to conform with our empirical application in Section \ref{sec_pf} considering a out-of-sample portfolio forecasting experiment, we denote the response variables $Y_{j,i}$ as asset returns and the regressors $X_i$ as observed factors, for $j = 1, \cdots, J$ and $i = 1, \cdots, n$, where $J$ represents the number of assets and $n$ denotes the number of periods. However, it is important to note that the results are still valid for general types of responses and regressors. 

\section{Covariance and Precision Matrix Estimate for Errors}\label{sec_cov_e_dnn_f}


In this section, we analyze the large sample properties of the error covariance and precision matrix estimates based on the deep neural network. We start with denoting $\sigma_{j,k}:= (\Sigma_u)_{j,k}$ as the $(j,k)$-th element of the error covariance matrix $\Sigma_u$, with $j=1,\cdots, J, k=1,\cdots,J$. 



Subsequently, we provide two maximal inequalities: the first one for the estimation of the error covariance matrix with infeasible sample average via Bernstein's inequality, and the second one provides a centered absolute version of the first maximal inequality.

\begin{lemma}\label{mtl1}

Under Assumption \ref{as1}, with a positive constant $C_2>0$, a positive constant $c_1$, and a sufficiently large $n$,\footnote {More specifically, a sufficiently large $n$ means that $n \ge \frac{C_2^2}{K^2} \log  J$ for Lemma \ref{mtl1}, given $\max_{1 \le j \le J} \max_{1 \le i \le n } |u_{j,i}| \le K^{1/2} $, where $K$ is a positive constant.} we obtain
\begin{enumerate}[label=(\roman*)]
	\item \[ P\left[  \max_{1 \le j \le J} \max_{1 \le k \le J} 
	| \frac{1}{n} \sum_{i=1}^n u_{j,i} u_{k,i} - E u_{j,i} u_{k,i} | \ge  C_2 \frac{\sqrt{\log J}}{\sqrt{n}}
	\right] \le \frac{2}{J^{c_1}}.\] \label{mtl1_i}
	\item \[
	P \left[ \max_{1 \le j \le J} \max_{1 \le k \le J} \left| \frac{1}{n} \sum_{i=1}^n |u_{j,i} u_{k,i} - \sigma_{j,k}| - E |u_{j,i} u_{k,i} - \sigma_{j,k}|
	\right|  >  C_2 \frac{\sqrt{\log J}}{\sqrt{n}} \right]\\
	\le  \frac{2}{J^{c_1}}
	\] \label{mtl1_ii}
\end{enumerate}

\end{lemma}

We use the following assumption linking the number of assets $J$ to the sample size $n$.

\begin{assum}\label{asa3}
\[ \log J/n \to 0,\]
as $n \to \infty$.

\end{assum}

\begin{lemma}\label{nl3}
Under Assumptions \ref{as1}-\ref{asa3}, we have
\begin{enumerate}[label=(\roman*)]
	\item  \[\max_{1 \le j \le J}  \frac{1}{n} \sum_{i=1}^n (u_{j,i}- \hat{u}_{j,i})^2  = O_p (l_n) =o_p (1).\] \label{l3_i}
	\item \[  \max_{1 \le j \le J } \max_{1 \le k \le J} | \frac{1}{n} \sum_{i=1}^n (\hat{u}_{j,i} \hat{u}_{k,i} - u_{j,i} u_{k,i}) | = O_p (l_n^{1/2})=o_p (1).\] \label{l3_ii}
	\item \[  \max_{1 \le j \le J }\max_{1 \le k \le J} | \frac{1}{n} \sum_{i=1}^n \hat{u}_{j,i} \hat{u}_{k,i} - E u_{j,i} u_{k,i}| = O_p (l_n^{1/2})=o_p(1).\] \label{l3_iii}
\end{enumerate}
 
\end{lemma}

Note that the results in Lemma \ref{nl3} are used to obtain the consistency of the covariance and precision matrix estimators of the returns. 
We have the following assumption that sets a lower bound on a demeaned second moment of the errors.

\begin{assum}\label{asa4}
\[ \min_{1 \le j \le J } \min_{1 \le k \le J} E | u_{j,i} u_{k,i} - \sigma_{j,k} |\ge c > 0.\]
\end{assum}

In the following, we specify the covariance matrix of the errors and its thresholding counterpart. 
Set $Y_i:= (Y_{1,i},\cdots, Y_{j,i}, \cdots, Y_{J,i})': J \times 1$.
The sample estimator is given by
\[ \hat{\Sigma}_u:= \frac{1}{n} \sum_{i=1}^n \hat{u}_i \hat{u}_i',\]
where $\hat{u}_i:= Y_i - \hat{f} (X_i)$ is the $J \times 1$ vector of residuals based on the deep neural network estimator. Also define the $(j,k)$-th element of $\hat{\Sigma}_u$ as 
$\hat{\sigma}_{j,k}$, for $j=1,\cdots, J$ and $k=1,\cdots, J$. 
We provide a new robust-adaptive thresholding estimator for the error covariance matrix estimation in deep learning, which is robust to outliers in the data, i.e., it is robust to large residuals. To that end, we define, for $j=1,\cdots, J$ and $k=1,\cdots J$ 
\begin{equation}
\hat{\theta}_{j,k}:= \frac{1}{n} \sum_{i=1}^n | \hat{u}_{j,i} \hat{u}_{k,i} - \hat{\sigma}_{j,k}|.\label{theta}
\end{equation}
Our adaptive thresholding estimator is represented as the $J \times J$ matrix $\hat{\Sigma}_u^{Th}$, where the $(j,k)$-th element of $\hat{\Sigma}_u^{Th}$ is computed as
\[ \hat{\sigma}_{j,k}^{Th}= \hat{\sigma}_{j,k} \1_{ \{ | \hat{\sigma}_{j,k}| \ge \hat{\theta}_{j,k} \omega_n \}}.\]
The rate $\omega_n$ is determined by the expression
\begin{equation}
\omega_n:= C_{**}  l_{n}^{1/2},\label{rwn}
\end{equation}
where $C_{**} > 0$ is a positive constant. Note that the above thresholding estimator is different from the one used in \cite{fan2011}, where they use a $l_2$ norm based definition for $\hat{\theta}_{j,k}$, unlike our $l_1$ norm based definition. The thresholding estimator in \cite{fan2011} is not suitable for merging theory of deep learning with non-linear factor based residuals for precision matrix estimation. By using a new $l_1$ based robust data dependent threshold in (\ref{theta}) we solve this problem, as can be seen in our proofs. 

Moreover, we define a sparsity pattern in the covariance matrix of the errors as
\[ s_n:= \max_{1 \le j \le J} \sum_{k=1}^J \1_{ \{ \sigma_{j,k} \neq 0 \}},\]
where $s_n$ represents the maximum number of nonzero elements across the rows of the error covariance matrix.

The following theorem establishes the consistency of the adaptive thresholding estimator for the covariance matrix of errors, by using the residuals resulting from the deep learning estimator in \eqref{dnn_est}.

\begin{thm}\label{thm3}
Under Assumptions \ref{as1}-\ref{asa4}, and let $C_3>0$ be a positive constant. Then,
\begin{enumerate}[label=(\roman*)]
	\item \[ P \left( \| \hat{\Sigma}_u^{Th} - \Sigma_u \|_{l_2} \le C_3 \omega_n s_n
	\right) \ge 1 - o(1)  .\] \label{thm3_i}
	\item Furthermore, if $\omega_n s_n = o(1)$ as an additional assumption,
	then with probability at least $1 - o(1)  $
	\[ Eigmin (\hat{\Sigma}_u^{Th} ) \ge Eigmin (\Sigma_u)/2,\]
	and 
	with probability at least $1 - o(1)$
	\[ \| [\hat{\Sigma}_u^{Th}]^{-1} - \Sigma_u^{-1} \|_{l_2} \le C_3 \omega_n s_n.\] \label{thm3_ii}
\end{enumerate}
\end{thm}


	\noindent Remarks. 
	1. Note that the rate of $\omega_n$ is given by $\omega_n = O (l_n^{1/2})= O (\max (r_{n1}^{1/2}, r_{n2}^{1/2}))$. Under Assumption \ref{asa3}, two possibilities arise that determine the dominant rate between $r_{n1}$ and $r_{n2}$. The first possibility is the case of an very high number of assets, where $J = \exp (n^a)$, with $1> a > \frac{d}{\beta+d}>0$. In this scenario, we have $\omega_n = O (r_{n2}^{1/2}) = O (n^{(a-1)/2})$. For all other possibilities of $J$ under Assumption \ref{asa3}, the rate is given by $\omega_n = O( r_{n1}^{1/2}) = O \left( n^{-\beta/2(\beta +d)} (\log (n))^4 \right)$. It is worth noting that this rate is not affected by the number of assets $J$, but rather affected by the number of factors $d$, which is constant.
	
	2. Negligible tail probabilities are written explicitly in our proofs, and they depend on number of assets, sample size and the number of factors.
	

\section{Covariance Matrix Estimator for the Returns}\label{cvmret}

In this section, we show that the covariance matrix of the asset returns can be estimated consistently based on the deep neural network estimator in \eqref{dnn_est}. We start with specifying the true covariance matrix of the returns and the corresponding estimator. In that respect, we consider the following nonparametric model for each period $i=1,\cdots, n$
\[ Y_i = f_0 (X_i) + u_i,\]
where $Y_i:=(Y_{1,i}, \cdots, Y_{j,i}, \cdots, Y_{J,i})'$ is a $J \times 1$-dimensional vector of asset returns, 
\[ f_0 (X_i):= (f_{0,1} (X_i), \cdots, f_{0,j} (X_i), \cdots, f_{0, J} (X_i))': \quad J \times 1,\] denotes the unknown function relating the observed factors $X_i$ of dimension $d \times 1$ with the asset returns and the idiosyncratic innovations are given by the $J \times 1$-dimensional vector $u_i:= (u_{1,i}, \cdots, u_{j,i}, \cdots, u_{J,i})'$. Define the covariance matrix of the returns as 
\[ \Sigma_y: = \Sigma^f + \Sigma_u.
\]
Subsequently, we define each component of $\Sigma_y$ and start with the covariance matrix for the functions of the factors $\Sigma^f$ given by 
\[ \Sigma^f:= E [(f_0 (X_i) - E f_0 (X_i))(f_0 (X_i) - E f_0 (X_i))'],\]
which is a $J \times J$ matrix. The $(j,k)$-th element of $\Sigma^f$ is
\[ \Sigma_{j,k}^f:= E [(f_{0,j} (X_i) - E f_{0,j} (X_i))(f_{0,k} (X_i) - E f_{0,k} (X_i))]
= E [f_{0,j} (X_i) f_{0,k} (X_i)] - E [f_{0,j} (X_i)] E [f_{0,k} (X_i)].\]

Moreover, we define the corresponding estimator for $\Sigma^f$ as follows
\[ \hat{\Sigma}^f:= \frac{1}{n} \sum_{i=1}^n (\hat{f} (X_i) - \bar{f} (X_i))(\hat{f} (X_i) - \bar{f} (X_i))',\]
with $\bar{f} (X_i):= \frac{1}{n} \sum_{i=1}^n \hat{f} (X_i)$, which is a $J \times 1$ vector, and 
\[ \hat{f} (X_i):= (\hat{f}_1 (X_i), \cdots, \hat{f}_j (X_i), \cdots, \hat{f}_J (X_i))': \quad J \times 1.\]

The following lemma proves the consistency of the estimate for the covariance matrix function of factors in a non-linear nonparametric model based on the deep neural network estimator \eqref{dnn_est}. As far as we know, this is a new result in the deep learning literature. It is important to note that we allow for the large dimensional setting $J >n$.

\begin{lemma}\label{l3a}
Under Assumptions \ref{as1}-\ref{asa3}, we have
\[ \| \hat{\Sigma}^f - \Sigma^f \|_{\infty} = O_p (l_{n}^{1/2}).\]

\end{lemma}

The estimator for the covariance matrix of the returns based on the deep neural network in \eqref{dnn_est} is given by
\[ \hat{\Sigma}_y:= \hat{\Sigma}^f + \hat{\Sigma}_u^{Th},\]
where we estimate the covariance matrix of the innovations using our adaptive thresholding estimator $\hat{\Sigma}_u^{Th}$ introduced in Section \ref{sec_cov_e_dnn_f}.
We establish the following theorem for the consistency of the covariance matrix of returns based on the deep neural network. To the best of our knowledge this is a novel result in the deep learning literature. 

\begin{thm}\label{covret}
Under Assumptions \ref{as1}-\ref{asa4} 
\[ \| \hat{\Sigma}_y - \Sigma_y \|_{\infty} = O_p (\omega_n).\]

\end{thm}

\noindent Remarks on Theorem \ref{covret}. 

\begin{enumerate}
	
	\item The sparsity of the error covariance matrix does not play any role in the estimation error, i.e.,\ the estimation error is not affected by $s_n$.
	
	\item The rate of convergence corresponds to
	\[ \omega_n = O (l_{n}^{1/2}) = O \left( \max(n^{-\beta/2(\beta+d)} [\log (n)]^4 , \sqrt{\frac{\log J}{n}}) \right).\]
	Hence, the rate is affected by the number of factors, $d$, jointly with smoothness coefficient $\beta$, or by the number of assets $J$. In the case of a very large number of assets, e.g.,\ $J = \exp (n^a)$, with $1> a > \frac{d}{\beta+d} >0$, the rate of convergence is given by $l_n^{1/2}=r_{n2}^{1/2}:= n^{(a-1)/2}$. For all remaining specifications of $J$ that satisfy Assumption \ref{asa3}, we obtain a rate $l_n^{1/2}= r_{n1}^{1/2}= O (n^{-\beta/2(\beta+d)} (\log n)^4)$.  
	
\end{enumerate}

\section{Precision Matrix Estimator for the Returns}\label{pmret}

In this section, we provide an explicit expression for the precision matrix for the returns. We start with introducing a formula for the inverse of the sum of two square matrices $A$ and $B$ both of dimension $J \times J$, where $A$ is a nonsingular matrix, and $B$ can be any square matrix
\begin{equation}
 (A + B)^{-1}= A^{-1} - A^{-1}  B ( I + A^{-1} B )^{-1} A^{-1}, \label{minv}
 \end{equation}
 which is from p. 348-349, Fact 3.20.8 of \cite{bern2018}. Set $A = \Sigma_u$ and $B = \Sigma^f$. Since $\Sigma_y = \Sigma^f + \Sigma_u$, where $\Sigma_u$ is a nonsingular matrix, we obtain the following expression for the precision matrix

\begin{equation}
\Sigma_y^{-1} = \Sigma_u^{-1} - \Sigma_u^{-1} \Sigma^f [ I_J + \Sigma_u^{-1} \Sigma^f]^{-1} \Sigma_u^{-1}.\label{s4-1}
\end{equation}
It is important to note that even with $\Sigma^f$ being singular, the inverse, $\Sigma_y^{-1}$ exists. In the same way, since $\hat{\Sigma}_u^{Th}$ is nonsingular, with probability approaching one as shown in Lemma \ref{la7}(ii), the estimate for the precision matrix of the returns based on the deep neural network in \eqref{dnn_est} is 
\begin{equation}
\hat{\Sigma}_y^{-1}= (\hat{\Sigma}_u^{Th})^{-1} - (\hat{\Sigma}_u^{Th})^{-1} \hat{\Sigma}^f [ I_J + (\hat{\Sigma}_u^{Th})^{-1}  \hat{\Sigma}^f ]^{-1} 
(\hat{\Sigma}_u^{Th})^{-1}.\label{s4-2} 
\end{equation}


In the following, we provide two assumptions that will be essential for proving the consistency of the precision matrix estimator based on the deep neural network.

\begin{assum}\label{as8}
\leavevmode
\begin{enumerate}[label=(\roman*)]
	\item \[ Eigmax (\Sigma_u ) \le C \delta_n,\] \label{as8_i}
	with $C > 0$ is a positive constant, and $\delta_n \to \infty$ with $n \to \infty$ and $\delta_n/J \to 0$ with $n \to \infty, J \to \infty$.
	\item \[ 0 \le e_n \le Eigmin (\Sigma^f) \le Eigmax (\Sigma^f) \le c_2 J,\] \label{as8_ii}
	with $  c_2 >0$ a positive constant, and $e_n \to 0$, with $n \to \infty$, or $e_n=0$.
	\item $\Sigma_u^{-1} \Sigma^f$ is symmetric.\label{as8_iii}
\end{enumerate} 
\end{assum}

The following assumption replaces Assumption \ref{asa3} and introduces a sparsity assumption.

\begin{assum}\label{as9}

$J^2  \omega_n s_n \to 0$, and $J = n^{b_1}, 0 < b_1 < \frac{\beta}{4(\beta+d)}$.

\end{assum}

Assumption \ref{as8}\ref{as8_i} is standard in large $J>n$ asymptotic framework, and commonly used in the literature, for the covariance matrices $\Sigma_u$ and $\Sigma^f$, which are both of dimension $J \times J$. Hence, the maximum eigenvalue of the covariance matrix of the errors is growing with $J$ at the rate $\delta_n$ but slower than $J$ itself. In that sense we control the rate of the noise. Moreover, in Assumption \ref{as8}\ref{as8_ii} the maximum eigenvalue of the covariance matrix of the factors is upper bounded by a multiple of $J$. This is also an extension of the linear-additive factor models in \cite{fan2011} to flexible non-linear factor models considered here. The minimum eigenvalue of the covariance matrix of the factors is either zero, or local to zero, reflecting the large dimensional problems resulting in singularity-near-singularity. Assumption \ref{as8}\ref{as8_iii} is technical in nature and helps us to derive a lower bound on minimum eigenvalue of a certain matrix in Lemma \ref{la6}(i). For example if 
$\Sigma_u^{-1}$ and $\Sigma^f$ are commuting, this assumption is satisfied, see p.604, Fact 7.6.16(i) of \cite{bern2018}.
Assumption \ref{as9} restricts $J <<n,$ unlike all the previous results. The main reason for this restriction is: to get a positive lower bound on the minimum eigenvalue of covariance matrix of the functions of factors. There is a tradeoff between number of factors $d$, and the number of assets $J$ as seen in Assumption \ref{as9}.

\subsection{Signal-to-Noise Ratio  and Weak Factor Issues} \label{sec_sn_wf}

In the following, we show a key aspect of our Assumptions \ref{as8}\ref{as8_i} and \ref{as8_ii}. Specifically, we will demonstrate that our assumptions allow for low signal-to-noise ratio, which is one of the main characteristics of the stock market. In order to analyze the implications, we consider a $J  \times 1$ vector, $\tau$, with $\| \tau \|_2^2=1$, which simplifies the theoretical elaboration below. We propose the following measure to quantify the signal-to-noise ratio in the deep neural network for a large portfolio of assets
\begin{equation}
 \frac{\tau' \Sigma^f \tau}{\tau' \Sigma_u \tau}.\label{sn1}
 \end{equation}
Note that in (\ref{sn1}) we can have the lower bound
\begin{equation}
\frac{\tau' \Sigma^f \tau}{\tau' \Sigma_u \tau} \ge \frac{ \| \tau\|_2^2 Eigmin (\Sigma^f)}{\| \tau \|_2^2 Eigmax (\Sigma_u)}
\ge \frac{e_n}{C \delta_n},\label{sn2}
\end{equation} 
where we use the Rayleigh inequality-quotient on p.343 of \cite{abamag2005} for both the numerator and denominator. Moreover, we use Assumptions \ref{as8}\ref{as8_i} and \ref{as8_ii} for the last inequality. Based on Assumption \ref{as8}\ref{as8_ii}, we know that the ratio $e_n/\delta_n$ is either zero or converging to zero. Hence, this result clearly shows that we allow for a low signal-to-noise ratio (as well as none). The lowest signal-to-noise ratio according to our definition in \eqref{sn1} is 0, which corresponds to the case, where there is only pure noise. If $e_n/\delta_n$ is zero or converging to zero (low signal-to-noise setup), the signal-to-noise ratio can be either 0 or converging to 0 from above. 
It is important to note that in a setting with low signal-to-noise ratio, which is implied by Assumptions \ref{as8}\ref{as8_i} and \ref{as8_ii}, the widely used Sherman-Morrison-Woodbury formula cannot be used, as it requires computing the inverse of $\Sigma^f$, which is singular-near singular in this setting. Hence, by introducing the new formula for the precision matrix of the returns in \eqref{s4-1}, we provide an expression that is compatible with a setting of low signal (i.e.,\ the minimum eigenvalue of the covariance matrix of non-linear factors is zero or local to zero).

Now we provide an upper bound for signal-to-noise ratio and tie this upper bound to our discussion of weak factors below.
In (\ref{sn1}), an upper bound on signal-to-noise ratio is
\begin{equation}
\frac{\tau' \Sigma^f \tau}{\tau' \Sigma_u \tau} \le \frac{ \| \tau\|_2^2 Eigmax (\Sigma^f)}{\| \tau \|_2^2 Eigmin (\Sigma_u)}
\le \frac{J}{C },\label{sn3}
\end{equation}

Note that the signal-to-noise ratio can be between the lower and upper bounds, and we show below that the upper bound can be smaller and even collapse to zero under a modification of Assumption \ref{as8}. However, note that our standard set of assumptions in Assumption \ref{as8} allow for a signal-to-noise ratio of zero or converging to zero by the minimum eigenvalue condition.

In the following, we analyze two cases that conform with the weak factor framework.

\noindent{\bf Case 1}. 
Based on Assumptions \ref{as8}\ref{as8_i} and \ref{as8_ii}, our deep learning framework allows for measuring weak factors. Compared to the pervasiveness assumption conventionally imposed in the standard linear factor models (see e.g., \cite{Bai2002} and \cite{fan2011}), which implies that the largest $d$ eigenvalues in $\Sigma^f$ diverge with the fast rate $J$, the weak factor assumption incorporates factors that considerably less influential.\footnote{See, e.g.\ \cite{Chudik2011}, \cite{Onatski2012}, \cite{Uematsu2022}, \cite{bai2023approximate}, and the references therein, who analyze the implications of weak factors in the linear factor model framework.} Specifically, in the weak factor setting, the $d$ largest eigenvalues of $\Sigma^f$ diverge with a rate, which is substantially slower than $J$. We analyze this case in the proofs after the proof of Theorem \ref{thm4}. To get some intuition, let us replace Assumption \ref{as8}\ref{as8_ii} by the following weak factor assumption

\renewcommand{\theassumwfactors}{W.5(ii)}
\begin{assumwfactors}\label{as8_alt1}
	\begin{equation}
		0 \le e_n \le Eigmin (\Sigma^f) \le Eigmax (\Sigma^f) \le c_2 \kappa_n,\label{sn4}
	\end{equation}
	with $  c_2 >0$ a positive constant, and $\kappa_n \to \infty$, but $\kappa_n/J \to 0$, and 
	$e_n \to 0$, with $n \to \infty$, or $e_n=0$.
\end{assumwfactors}
	
Hence, this new assumption offers an alternative to Assumption \ref{as8}\ref{as8_ii} and assumes a divergent largest eigenvalue for the non-linear factor covariance matrix, however it grows at a slower rate than the number of assets $J$. Hence, Assumption \ref{as8_alt1} is more amenable for the weak factor scenario. Moreover, we obtain the following lower and upper bounds on signal-to-noise ratio
\begin{equation}
\frac{e_n}{C \delta_n} \le  \frac{\tau' \Sigma^f \tau}{\tau' \Sigma_u \tau} \le \frac{c_2 \kappa_n}{C}. \label{sn5}
 \end{equation}
Hence, by taking $n \to \infty$, the lower bound will collapse to zero, and upper bound still diverges, however at a slower rate compared to the setting with Assumption \ref{as8}\ref{as8_ii}, as shown by equation \eqref{sn3}. We cover a large number of possibilities regarding the signal-to-noise and factor models with (\ref{sn5}).

\noindent{\bf Case 2.} 
In the second case, we consider a setting with very weak factors, which implies that the largest eigenvalue of the covariance matrix of the non-linear factors converges to zero. Hence, similar as the lower bound, the upper bound on the signal-to-noise ratio collapses to zero as well. In order to see this, we change Assumption \ref{as8}\ref{as8_ii} to the following specification, which incorporates a environment with very weak factors (this replaces Assumption \ref{as8_alt1})

\renewcommand{\theassumwfactors}{VW.5(ii)}
\begin{assumwfactors}\label{as8_alt2}
	\begin{equation}
		0 \le e_n \le Eigmin (\Sigma^f) \le Eigmax (\Sigma^f) \le c_2 \kappa_n,\label{sn6}
	\end{equation}
	with $  c_2 >0$ a positive constant, and $\kappa_n \to 0$, $\kappa_n > e_n$ and 
	$e_n \to 0$, with $n \to \infty$, or $e_n=0$.
\end{assumwfactors}
	
In this setting, the largest eigenvalue of the covariance matrix of the non-linear factor model can converge to zero, which implies that all factors can be very weak. This gives rise to a collapsing signal-to-noise ratio to zero. In \eqref{sn5}, with $\kappa_n \to 0, e_n \to 0$, $\delta_n \to \infty$, the signal-to-noise ratio collapses to zero. For the case that the rate of the largest  eigenvalue is $\kappa_n = O(1)$ , we obtain the same results for precision matrix estimation of  the returns as in the case of very weak factors. Hence, we only briefly cover that scenario after the proof of Theorem \ref{thm4} in the Appendix \ref{sec_A_proofs}.
	
The proof of consistency and rate of convergence of the estimate of the precision matrix of returns this very-weak factor case will be presented after the proof of Theorem \ref{thm4}.
Hence, the weak factor setting conforms with the case of a low or very low signal-to-noise ratio as elaborated in the previous paragraph. Moreover, empirical evidence testifies that the weak factor assumption better explains the spectral properties of financial and macroeconomic datasets (see, e.g.,\ \cite{Trzcinka1986} and \cite{ludvigson2009macro}). Hence, our model setting is better suited to measure the spectral structure of datasets incorporating assets returns compared to standard linear factor models that rely on the pervasiveness assumption.

\subsection{Main Theorem}

The following theorem shows the consistency of the precision matrix of the returns under our standard set of assumptions. We proceed with a new proof due to the non-linear nature of the factors.

\begin{thm}\label{thm4}
Under Assumptions \ref{as1}-\ref{as2},\ref{asa4}-\ref{as9} 
\[ \| \hat{\Sigma}_y^{-1} - \Sigma_y^{-1} \|_{l_2} = O_p ( J^2 \omega_n s_n) = o_p (1).\]

\end{thm}

\noindent Remarks on Theorem \ref{thm4}.

\begin{enumerate}
	\item Our rate both depends on both the number of assets $J$ and the number of factors $d$, where the rate deteriorates with larger portfolios and an increasing number of factors. We can only allow for $J<<n$ as shown by Assumption \ref{as9}. This is due to non-linear nature of the factor model and Assumption \ref{as8}. Note that our rate does not depend on $e_n, \delta_n$ which control the minimum signal to noise ratio as in (\ref{sn2}).
	
	\item It is also difficult to compare our rate with any known linear model in terms of rates of convergence, without putting any structure on the non-linearity. We cover a general non-linear unknown function with observed factors, and estimate it with deep neural networks.																																																	  
	
	\item Since $\omega_n=O( l_n^{1/2})$, based on Assumption \ref{as9} we have that $l_n = O (r_{n1})$. Hence, $J^2 \omega_n s_n$ simplifies to
	\[ J^2 \omega_n s_n = O (s_n n^{\frac{4 b_1 (\beta+ d) - \beta}{2(\beta+d)}} (\log n)^4).\]
	
	\item We analyze the implications of a weak factor or a very weak factor assumption on the results for Theorem \ref{thm4} in a separate paragraph after the proof of Theorem \ref{thm4} in Appendix \ref{sec_A_proofs}. Specifically, we establish the Corollaries \ref{coll_wf1} and \ref{coll_wf2}, which show that the rate of convergence of the precision matrix estimator improves compared to the rate derived in Theorem \ref{thm4} and depends linearly in $J$. This is due to the fact that the precision matrix estimator is no longer affected by the fast diverging eigenvalues of the covariance matrix of the non-linear factors, which increase with the rate $J$ implied by the standard strong factor assumption. However, for the consistency of the estimator we still require $J <n$.
\end{enumerate}


\section{Additive Non-linear Factor Models}\label{sec_factor_model}

In this part of the paper, we make some simplifying assumptions and show that the curse of dimensionality due to the number of factors $d$ may disappear in an additive but still non-linear-unknown function model. 

\subsection{The Additive Model}

In the finance literature, asset returns are governed by common factors. The following model is used in \cite{fan2011}:
\begin{equation}
	Y_{j,i} = \sum_{m=1}^d b_{j,m} X_{m,i} + u_{j,i},\label{lfm}
\end{equation}
for all $j=1,\cdots, J$, where $J$ represents the number of assets in the portfolio, and $i=1,\cdots, n$ denotes the time span of the portfolio. The number of factors is $d$, and $d \ge 1$, and the number of factors do not vary with $n$, and  $d$ is constant. The factors are observed. $X_{m,i}$ represents the $m$-th factor at time period $i$, and $b_{j,m}$ depicts the factor loading corresponding to the $j$-th asset and $m$-th factor. The model described above defines a linear relation between the factors and returns through the factor loadings. 
However, a more flexible relationship between asset returns and factors can be put forward.  Instead of the linear model specification in \eqref{lfm}, we assume that the returns evolve through the following model
\begin{equation}
Y_{j,i} = f_{0,j} (X_i) + u_{j,i},\label{m1}
\end{equation}
with 
\begin{equation}
 f_{0,j} (X_i):= \sum_{m=1}^d f_{j,m} (X_{m,i}),\label{m1a}
 \end{equation}
where $f_{0,j}(.)$ represents the true but unknown function relating the factors $X_i:=(X_{1i},\cdots, X_{m,i},\cdots, X_{d,i})': d \times 1$ to the returns for each asset. $f_{j,m}(.)$ corresponds to the unknown function of factor $X_{m,i}$ for the $m$-th factor at time period $i$. The model specification \eqref{m1} enhances the flexibility in the relationship between the factors and assets to a large extent, compared to the rigid all linear additive formulae \eqref{lfm}. We want to estimate $f_{0,j}(.)$ with sparse deep learning, for each $j=1,\cdots, J$. The additive-flexible model (\ref{m1}) is related to Section 4 and equation (12) of \cite{sh2020}. We can specify the true function $f_{0,j}(.)$ as the composite of two functions as follows
\begin{equation}
f_{0,j}(.)   = g_{1,j}(.)  \circ g_{0,j}(.) ,\label{5a}
\end{equation}
where
\[ g_{0,j}(.) := [g_{0,j,1}(.) ,\cdots, g_{0,j,m}(.) ,\cdots, g_{0,j,d}(.) ]'
:=[ f_{j,1}(.),\cdots, f_{j,m} (.),\cdots, f_{j,d}(.) ]'
,\]
which is a $d \times 1$ vector. Then
\[ g_{1,j} (X_i):= \sum_{m=1}^d g_{0,j,m} (X_{m,i}) := \sum_{m=1}^d f_{j,m} (X_{m,i}).\]

For each $j=1, \cdots, J$ and $m=1,\cdots d$ 
\[ f_{j,m} \in {\cal C}_1^{\beta} ([0,1], K),\]
where ${\cal C}_1^{\beta} ([0,1], K)$ represents ball of $\beta$ Hölder functions
with radius $K$, see p.1880, 1885 of \cite{sh2020}. These $\beta$ Hölder functions are functions with partial derivative up to order $\beta_f$ exist and are bounded, and the partial derivative of order 
$\beta_f$ are $\beta-\beta_f$ Hölder, and $\beta_f$ represents the largest integer strictly smaller than $\beta$.
Since $\beta>0$, and Hölder ball definition-domain is slightly different there is a different notation in Assumption \ref{as2}. The technical definition of $\beta$ Hölder functions is relegated to Appendix \ref{sec_b1_gcf}.
Furthermore we impose that true composite function belongs to the following function space
\begin{equation}
 f_{0,j} \in {\cal G} (1, \tilde{d}, \tilde{t}, \tilde{\beta}, (K+1) d),\label{fg}
 \end{equation}
where we define  the function space ${\cal G}(.), \tilde{d}, \tilde{t}, \tilde{\beta}$ in Appendix \ref{sec_b1_gcf}. This function space, ${\cal G}$,  consists of composite of $\beta$ Hölder functions.

\subsection{S-sparse Deep Neural Networks}

Our empirical minimizer satisfies the following sparsity and bounded parameter requirements in the deep neural network.
\cite{sh2020} introduces them in equation (4) of his paper and requires two important restrictions. \cite{sh2020} argues that non-sparse deep neural networks can result in overfitting hence large out-of-sampling errors. We analyze this last statement both theoretically, but also with simulation, and most importantly in out-of-sample asset pricing example from the US Stock Market. There are two key aspects of deep-sparse neural networks. 
First, the parameters will be bounded by one, and second the network will be sparse. Note that also by imposing a sparse deep neural network estimator with bounded weights we try to understand how sparsity especially play a role in the theory in deep networks. Sparsity in our proof is used to control the covering numbers for the function space as shown in Step 1b of the proof of Theorem \ref{thm1}.
In Remark 8 of Theorem \ref{thm1} we show relaxing of these restrictions does not change risk upper bound up to a slowly varying function in $n$.

The first restriction can be formally depicted as follows: For each $j=1,\cdots, J$

\begin{equation}
\max_{0 \le l \le L } [\| W_l \|_{\infty} \vee  \| v_l \|_{\infty}] \le 1.\label{c1}
\end{equation}

This restriction is necessary to ensure the practical applicability of the neural networks. Statistical results using large parameters/weights are usually not observed, see \cite{gbc2016}. To be specific, \cite{sh2020} argues that computational algorithms use random, nearly orthogonal matrices as initial weights. Hence, the initial weights are bounded by one, and the model training leads to trained weights, which are close to the initial values. Therefore, it seems reasonable to set a bound of one for the weights. 

In the next step, we introduce our sparsity assumption, which is crucial to prevent overfit caused by a probable large number of hidden units in each layer. For each function $j=1,\cdots, J$, we impose the same number of hidden units $p=(p_0, p_1, \cdots, p_L, p_{L+1})'$ where $p_0=d, p_{L+1}=1$, where we have $d$ inputs for each function $f_{0j}$ and the output is scalar.
Let $\| W_l\|_0, \| v_l \|_0$ be the number of nonzero entries of $W_l, v_l$, respectively. The s-sparse deep neural networks are given in \cite{sh2020}, subject to \eqref{c1} and 
\begin{equation}
	{\cal F}_j (L, p, s_j):= \{ f_j (.): \sum_{l=0}^L  \| W_l \|_0 + \sum_{l=1}^L  \| v_ l \|_0 \le s_j, \quad  \max_{1 \le j \le J} |f_j (.)| \le F \},\label{c2}
\end{equation}
where the output dimension is one and the functions are uniformly bounded by $F$, which is a positive constant. Clearly, the sparsity is different for each ${\cal F}_j (L, p, s_j)$. In order to simplify the notation, we represent ${\cal F}_j (L, p, s_j)$ as ${\cal F}_j$.
The sparsity structure imposed on the neural networks weights are additionally valuable for the consistency of the of deep learning estimators compared to model specifications that only rely on the implicit regularization introduced by the stochastic gradient decent algorithm during the computational stage. This point is conjectured explicitly, and an example is provided on p. 1916 of \cite{sh2020b}.

\subsection{Theory}

We start with Assumptions that are necessary for the sparse deep neural network. In contrast to the large sample elaboration on the dense deep neural network in Section \ref{sec_model}, we will not assume bounded outcomes in the sparse DNN framework. However, we strengthen other assumptions such as sparsity of the neural network parameters. The  Sparse Deep Neural Network (SDNN-FM) factor model estimator   $\hat{f}_j $
 is defined as
\[ \hat{f}_j:= \argmin_{f_j \in {\cal F}_j} \sum_{i=1}^n [ Y_{j,i} - \sum_{m=1}^d f_{j,m} (X_{m,i})]^2. 
\]

Let $\phi_n^*:= n^{-2\beta/(2 \beta +1)}$ be the rate of the approximation error for the true function $f_{0,j}$ by the sparse deep neural network estimator $\hat{f}_j$ for each $j=1,\cdots, J$, described in Appendix \ref{sec_b1_gcf} and (\ref{aer}).

\begin{assum}\label{asa1}
Assume that $u_{j,i}$ are iid zero mean with unit variance across $i=1,\cdots, n$ with subgaussian distribution and Orlicz norm, $\max_{1 \le j \le J} \| u_{j,i} \|_{{\psi}_2}:= C_{\psi}$,
which is a positive constant. Let the minimum eigenvalue of the covariance matrix of errors be $Eigmin (\Sigma_u) \ge c > 0$, where $c$ is a positive constant. $X_i$ are iid across
 $i=1,\cdots,n$ and independent from $u_{j,i}$ for each $j$.
\end{assum}



 Assumption \ref{asa1} is subgaussian noise extension of the gaussian noise assumption imposed in \cite{sh2020}.

\begin{assum}\label{as4}
Assume $f_{0,j}(.)$ is a composite function generated by equation (\ref{fg}) for each $j=1,\cdots, J$. Let $ {\cal F}_j$ satisfy \eqref{c1} and \eqref{c2}.\end{assum}

\begin{assum}\label{as5}

\begin{enumerate}[label=(\roman*)]
	\item The envelope $F$  of the network in (\ref{c2}) is an upper bound on the range of the true function as in (\ref{fg}): $F \ge  (K+1)d$, \label{as5_i}
	\item  The optimal number of hidden layers $L$ is growing with sample size:
	$ L \asymp \log_2 n, $ \label{as5_ii}
	\item  The number of hidden units-minimum across all hidden layers-  should exceed the following lower bound
	$n \phi_n^* \le C \min_{1 \le l \le L } p_l$, \label{as5_iii}
	\item The sparsity of the network parameters can grow with n, for each $j=1,\cdots, J$
	$ s_j \asymp n \phi_n^* \log n$ for each $j=1,\cdots, J$.\label{as5_iv}
\end{enumerate}
\end{assum}

Assumptions \ref{as4} and \ref{as5} are used to approximate the true function, $f_{0,j}(.)$ by the sparse deep learning estimator $\hat{f}_j$. Assumption \ref{as4} specifies the true composite function. Assumption \ref{as5}\ref{as5_i}-\ref{as5_ii} put an upper bound on the functions, and specifies the number of layers $L$. The number of layers is an increasing function of $n$. The number of units at each layer is specified in Assumption \ref{as5}\ref{as5_iii} and provides a lower bound of this quantity. Hence, the number of units increase with $n$, and form a wide layer. The sparsity restriction is specified in Assumption \ref{as5}\ref{as5_iv}. Given the specification of the rate $\phi_n^*$, the sparsity assumption shows that $\bar{s}/n \to 0$, with $\bar{s}:=\max_{1 \le j \le J} s_j $
. Assumptions \ref{as4}-\ref{as5} are directly taken from \cite{sh2020}, and see p.1885 of \cite{sh2020} for non-linear-additive functions. In the following, we provide the upper bound on the risk for our functions that are estimated by deep learning.

\begin{thm}\label{thm1}
Under Assumptions \ref{asa1}-\ref{as5}, for sufficiently large $n$,
\begin{enumerate}[label=(\roman*)]
  
	\item \[ \max_{1 \le j \le J} E \left[ \frac{1}{n} \sum_{i=1}^n (\hat{f}_j (X_i)  - f_j (X_i))^2 \right]\le C \phi_n^* L \log^2 n.\] \label{thm1_i}

	\item \[ P \left[  \max_{1 \le j \le J} \frac{1}{n} \sum_{i=1}^n [ \hat{f}_j (X_i) - f_{0,j} (X_i)]^2 \ge  r_{n3} + r_{n4}
	\right] \le \frac{1}{J^{c_1^2}},\]
	with rates $r_{n3}= O ( n^{\frac{-2\beta}{2 \beta +1}} \log^3 n), r_{n4}= O (\sqrt{\log J/n}).$ \label{thm1_ii}
\end{enumerate}
\end{thm}

\noindent Remarks. 
\begin{enumerate}
	\item By  $L = O (\log_2 n)$, in which case the right side in \ref{thm1_i} can be written as 
	\[ \max_{1 \le j \le J} E \left[ \frac{1}{n} \sum_{i=1}^n (\hat{f}_j (X_i)  - f_j (X_i))^2 \right] \le C \phi_n^* \log^3 n .\] 
	
	\item The results in Theorem \ref{thm1} indicate that we require a large number of layers, and the number of units at each hidden layer cannot be small, in order to obtain a good function approximation. These restrictions are specified in Assumption \ref{as5}. However, for a better prediction we need fewer layers as observed by our result in Theorem \ref{thm1}. Hence, there is a tradeoff in the selection of the optimal number of layers. While the depth of the neural network is essential for achieving a better function approximation, the question arises: how deep should the network be? \textit{Theorem 5} in \cite{sh2020} emphasizes the importance of deep layers in obtaining an improved function approximation. 
	
	\item The results of Theorem \ref{thm1} further show that the upper bound is uniform over $j=1,\cdots, J$. This is due to Assumption \ref{as5}\ref{as5_iv}. Moreover, we also allow that the true functional forms are different. However, they have the same Hölder smoothness. Concerning the deep learning estimators, we have the same number of layers and units for each estimate but their sparsity patterns can vary.
	
	\item The sparsity in the neural network can be achieved by active regularization at each layer through an elastic net penalization, or an iterative pruning approach as suggested on p. 1882 of \cite{sh2020}, and p. 1917 of \cite{sh2020b}, respectively. \textit{Lemma 5} of \cite{sh2020} combined with our Lemma \ref{l2} and Step 1b in the proof of Theorem \ref{thm1}\ref{thm1_i} shows that the sparsity in the network parameters at each layer is essential for a better prediction. The implicit regularization induced by the stochastic gradient descent algorithm will not be sufficient, and the consistency of the deep neural network estimator may be not achieved as shown in pp. 1916-1917 of \cite{sh2020b}. We also analyze these issues of sparse versus non-sparse networks in an empirical out-of-sample portfolio exercise for the US Stock Market. Moreover, we consider different simulation studies, which aim at comparing the two approaches in an in-sample estimation analysis.

	\item An important aspect concerns the consistency of the deep neural network estimator, which is not affected by number of factors, $d$. This is due to additive structure of the model, the novel proofs in \cite{sh2020} and our proof for the subgaussian noise in Theorem \ref{thm1}. The quantities $r_{n3}$ and $r_{n4}$ correspond to the rates of convergence associated with the risk of the deep learning estimator and the fact of using $J$ assets in the portfolio, respectively.
	
	\item Theorem \ref{thm1}\ref{thm1_ii} is a novel result in the literature and shows the implications of the size of the portfolio and the number of factors on the convergence rate of the deep learning estimate for the true underlying function that relates the factors to the returns. Our result extends the contributions of \cite{sh2020} to the estimation of multiple functions. Hence, we obtain the additional rate $r_{n,4}$ due to estimation of $J$ different functions.
	
	\item In the subsequent analysis, we examine which rate may be the slowest among $r_{n3}$ and $r_{n4}$. Suppose that $J = \exp (n^{a_1})$ and $0 < a_1 < 1$. We obtain $r_{n,4}$ as the slower rate as long as $(a_1-1)/2 > -2 \beta/(2 \beta +1)$, which is true if $\beta> \frac{1-a_1}{2 (1+ a_1)}$. Hence, regardless of $0 < a_1 < 1$, when $\beta > 1/2$, we obtain the rate $ r_{n,4}$ with $J = \exp (n^{a_1})$.	If we consider $J = a_2 n$, with $0 < a_2 \le C < \infty$ and $\beta > 1/2$, a similar reasoning applies, leading to the rate $r_{n4}$. The same logic applies to the case of $J = n^{a_3}$, with $a_3 > 0$. However, it is important to note that if $\beta \le 1/2$, it is not clear which rate will be slowest. The rate depends on the tradeoff between the smoothness coefficient $\beta$, and the number of assets $J$.
	
	
	\item A non-sparse approach, allowing for unbounded weights in deep neural network estimation, has been explored in \cite{kl2021} using \textit{Theorem 1a}. Their approach yields similar rates of convergence to those presented in \cite{sh2020} for the additive non-linear case, up to a slowly varying function in $n$. However, it remains unclear how this approach can be extended to handle multiple outcomes, as established in our paper. In contrast, our paper provides a general framework through Theorem \ref{t1}, which does not rely on a sparsity assumption, boundedness of the weights, or the assumption of a composite true function. Notably, \cite{kl2021} still rely on the assumption of a composite true function.

\end{enumerate}

Subsequently, we set up rates that will be used in the following theorem. Let

\begin{equation}
a_n^2:= \max(r_{n3}, r_{n4}).\label{an}
\end{equation}
 and 
\[ \bar{\omega}_n:= C \left( \frac{\log J}{n} + a_n \right) = O (a_n),\]
where the last equality is due to $r_{n4}= O (\sqrt{\frac{\log J}{n}})$.

Moreover, let us denote the estimate of the precision matrix of errors by using the sparse deep neural network estimator in a non-linear additive model by 
$(\hat{\Sigma}_u^s)^{-1}$. The main difference from $(\hat{\Sigma}_u^{Th})^{-1}$, which corresponds to the error precision matrix of the dense deep neural network in \eqref{dnn_est}, concerns the implied residuals that are obtained from the sparse deep neural network estimator. Moreover, the covariance matrix estimator is subject to the same thresholding formula, however $\omega_n$ in the threshold is replaced with the quantity $\bar{\omega}_n$. Denote $\hat{\Sigma}_y^s$ and $(\hat{\Sigma}_y^s)^{-1}$ as the covariance and precision matrix of the returns in a non-linear additive model estimated by the sparse deep neural network.
The following theorem establishes the consistency of the estimate of precision matrix of the errors and returns for a non-linear additive model setting estimated by the sparse deep neural network.

\begin{thm}\label{ds}
\leavevmode
\begin{enumerate}[label=(\roman*)]
	\item Under Assumptions \ref{asa3}-\ref{asa4} and \ref{asa1}-\ref{as5}, with $\bar{\omega}_n s_n \to 0$ \label{ds_i}
	\[ \| (\hat{\Sigma}_u^s)^{-1} - \Sigma_u^{-1} \|_{l_2} = O_p (\bar{\omega}_n s_n ) = o_p(1).\]

	\item Under Assumptions \ref{asa3}-\ref{asa4} and \ref{asa1}-\ref{as5},
	\[ \| \hat{\Sigma}_y^s - \Sigma_y \|_{\infty} = O_p (\bar{\omega}_n) = O_p ( a_n) = o_p (1).\] \label{ds_ii}

	\item Under Assumptions \ref{asa4}-\ref{as5}
	\[ \| (\hat{\Sigma}_y^s)^{-1} - \Sigma_y^{-1} \|_{l_2} = O_p ( J^2 \bar{\omega}_n s_n) = o_p (1).\]\label{ds_iii}
\end{enumerate}

\end{thm}

\noindent Remarks on Theorem \ref{ds}. 

\begin{enumerate}


	\item  The proof of Theorem \ref{ds} follows directly from the proof of Theorems \ref{thm3} and \ref{thm4}, by replacing $l_{n}^{1/2}$ with $a_n$. The details and a step by step proof are given in \cite{cd2022} which is the working paper that contains the proofs of the sparse-network that is analyzed in this section.

	\item Theorem \ref{ds}\ref{ds_iii} is a new result that provides a consistent deep learning based non-linear factor model estimate of the precision matrix of the returns. It is important to note that the estimation error does not depend on the number of factors. However, it is affected by the sparsity ($s_n$) in the covariance matrix of errors, the number of assets $J$, and the estimation error rate $\omega_n$, which is used for the estimator of the error covariance matrix. Hence, we can only allow for $J <<n$, however we still have $J \to \infty$ when $n \to \infty$.

\item Issues that are related to signal-to-noise ratio and weak factors are the same as in Section \ref{pmret}.

\end{enumerate}

\section{Implementation}\label{sec:impl}
In the following, we discuss the implementation details of our deep neural network factor model with fully connected units (DNN-FM) in Section 2 and our sparse deep neural network factor model (SDNN-FM) with sparsely connected units introduced in Section 6, which are crucial for the performance of the models. 

In order to train and validate our models on different sets of data, we split our in-sample data into two distinct subsets, which correspond to the training and validation dataset, respectively. This data division is essential to reduce the overfitting. More precisely, only the data in the training set is used to estimate the DNN-FM and SDNN-FM for a specific set of hyperparameters. The validation set serves as proxy for an out-of-sample test of the model and is used to optimize the hyperparameters. In our implementation, we use the first 80\% of the in-sample data for training the model, whereas the remaining 20\% are used for the model validation.

In order to explicitly incorporate sparsity in the weights of our SDNN-FM we use three regularization methods. These methods reduce the effective number of parameters to be estimated in the neural network. Specifically, we augment the objective function by a $l_1$-norm penalty on the weights of the neural network, which sets elements of the weight matrices to zero. Hence, the $l_1$-norm penalty induces sparsity in the parameters of the neural network and allows for disregarding uninformative weights. The strength of the penalty is controlled by a tuning parameter, which we select based on the validation set. Furthermore, we use Dropout, introduced by \cite{Srivastava2014} as a second technique to reduce overfitting in the neural network. The Dropout technique randomly disables a prespecified number of units and their corresponding connections in each layer of the neural network during the training process. This reduces the occurrence of complex co-adaptations between the units on the training data. These co-adaptations generally lead to a close adjustment of the neural network to the training data, which compromises its ability to generalize to new data that has not been seen during training. Hence, Dropout mitigates this problem and improves the out-of-sample performance. In our implementation, we randomly disable 20\% of the units in each layer during the training process. As a third regularization technique, we adopt Early Stopping. During each step of an iterative optimization algorithm (e.g.,\ stochastic gradient decent (SGD)) the neural network parameters are adjusted such that the fitting error based on the training data is minimized. While the training error decreases in each iteration, this is not true for the validation error, which is used as a proxy for the out-of-sample error. Early Stopping keeps track of the validation error and terminates the optimization as soon as the validation error starts to increase. For both SDNN-FM and DNN-FM we use the Early Stopping rule. Hence, the difference between these two methods arise from incorporating $l_1$-norm regularization and Dropout in the SDNN-FM. Clearly, we need Early Stopping for both non-sparse and sparse deep neural network estimation, to avoid an increase of the out-of-sample validation error.

In order to optimize the DNN-FM and SDNN-FM, we refer to the adaptive moment estimation algorithm (Adam) introduced by \cite{Kingma2014}, which offers an efficient adaptation of the stochastic gradient decent (SGD) algorithm. More precisely, compared to SGD it provides an adaptive learning rate by using the information of the first and second moments of the gradient.

We estimate the covariance matrix of the residuals of the DNN-FM based on our novel thresholding procedure introduced in Section \ref{sec_cov_e_dnn_f}. Specifically, we use $\hat{\theta}_{j,k}\omega_n$ as threshold level, where $\hat{\theta}_{j,k}$ is specified in \eqref{theta} and $\omega_n$ is set to $3\sqrt{\frac{\log J}{n}}$. To estimate the covariance matrix of the innovations for the SDNN-FM, we also rely on our thresholding procedure, however use the SDNN-FM residuals as elaborated in Section \ref{sec_factor_model} to build $\hat{\theta}_{j,k}$ and $\bar{\omega}_n$ is set to the same level as $\omega_n$ so that DNN-FM and SDNN-FM differences can be only due to sparse number of parameters estimated in hidden layers, and possible data generating processes.


\section{Simulation Evidence}\label{sec_sim}

The aim of this section is threefold. First, we verify our theorems. Second, we compare-contrast our DNN-FM, SDNN-FM methods with the ones that are widely used in the literature tied to machine learning. Third, we analyze the performance of our methods under low signal-to-noise ratio and weak factor setup. 

\subsection{Monte Carlo designs and Models}
To evaluate our theoretical findings, we analyze three Monte Carlo designs. For the first design, we consider the following data generating process (DGP) 

\begin{align}
	Y_{j,i} = \sum_{m = 1}^{d} \1_{\left\{m \text{ is even} \right\}} \beta_{m,j} X_{m,i} + \1_{\left\{m \text{ is odd} \right\}} \beta_{m,j} X_{m,i}^2 + u_{j,i},\label{sim_dgp1}
\end{align}
where $\1_{\{\cdot\}}$ defines an indicator function that is equal to one if the boolean argument in braces is true. Moreover, the coefficients $\beta_{m,j}$ and the explanatory variables $X_{m,i}$ are drawn from the standard normal distribution, for $m = 1, \cdots d$, $j = 1, \cdots, J$ and $i = 1, \cdots, n$.\footnote{We run the same study with fixed coefficients $\beta_{m,j}$ and obtain similar results as in the random coefficients design. Therefore, we omitted these simulation results, however they can be obtained from the authors upon request.} This DGP is more aligned with SDNN-FM since it comprises a non-linear additive structure. The innovations $u_{j,i}$ are generated according to the following process, which is similar to the DGP used in \cite{Bai2016}. 
\begin{align}
	\begin{split}
		& u_{1,i} = e_{1,i}, \quad u_{2,i} = e_{2,i} + a_1e_{1,i}, \quad  u_{3,i} = e_{3,i} + a_2 e_{2,i} + b_1 e_{1,i}, \\
		& u_{j, i} = e_{j, i} + a_{j-1} e_{j-1,i} + b_{j-2} e_{j-2, i} + c_{j-3} e_{j-3, i}, \quad \text{ for } j = 4, \cdots, J, \label{sim_dgp_u}
	\end{split}
\end{align}
where $e_{j,i}$ are iid $\mathcal{N}(0,1)$, for $j = 1, \cdots, J$, $i = 1, \cdots, n$ and $a_j, b_j, c_j$ are independently drawn from $ \mathcal{N}(0,0.25)$, for $j = 1, \cdots, J$. Hence, the process \eqref{sim_dgp_u} implies some cross-sectional correlations between the innovations and yields a banded covariance matrix $\Sigma_u$.

Compared to the standard static factor model specification, the process in \eqref{sim_dgp1} additionally incorporates non-linearities through the term $\beta_{m,j} X_{m,i}^2$. Hence, we expect that the traditional static factor model estimated with principal component analysis may have greater difficulties in capturing these non-linearities compared to our deep neural network factor models.

In order to verify the robustness of our simulation results, we consider two additional Monte Carlo designs. The second Monte Carlo design relies on a similar DGP as in \cite{Farrell2021}. Specifically, we simulate the data according to the following process
\begin{equation}
	Y_{j,i} = \alpha_j' X_i + \beta_j' \psi(X_i) + u_{j,i},\label{sim_dgp2}
\end{equation}
where $X_i$ incorporates $d$ independent standard normally distributed random variables and the idiosyncratic innovations follow the process in \eqref{sim_dgp_u}. Moreover, $\psi(.)$ denotes a non-linear multivariate transformation function, which incorporates second-degree polynomials and pairwise interactions and more in line with DNN-FM and what we observe in financial data. Hence, this simulation design enhances the non-linear influence of $X_i$ on the response compared to the first design in \eqref{sim_dgp1}. The coefficient vectors $\alpha_j$ and $\beta_j$ are both of dimension $d \times 1$ and drawn from $\mathcal{U}(-1, 1)$ and $\mathcal{U}(-0.5, 0.5)$, respectively, and they represent the columns in $\alpha, \beta$ matrices which are of dimension $d \times J$.

\begin{table}[!t]
	\footnotesize
	\centering
	\caption{First Monte Carlo design - Signal-to-noise ratio and spectral properties}
	\begin{threeparttable}
		\begin{tabular}{c|ccccccc}
			\hline
			\hline
			$d$     & \multicolumn{1}{l}{Signal} & \multicolumn{1}{l}{Noise} & \multicolumn{1}{l}{Signal-to-Noise Ratio} & \multicolumn{1}{l}{Eigmin $\Sigma_f$} & \multicolumn{1}{l}{Eigmax $\Sigma_f$} & \multicolumn{1}{l}{Eigmin $\Sigma_u$} & \multicolumn{1}{l}{Eigmax $\Sigma_u$} \bigstrut\\
			\hline
			& \multicolumn{7}{c}{$J = 50$} \bigstrut\\
			\hline
			1     & 2.08  & 1.73  & 1.21  & -2.16E-14 & 99.72 & 0.03  & 5.82 \bigstrut[t]\\
			3     & 5.36  & 1.75  & 3.06  & -2.81E-14 & 119.91 & 0.03  & 5.78 \\
			5     & 8.37  & 1.74  & 4.81  & -3.26E-14 & 133.74 & 0.03  & 5.81 \\
			7     & 10.67 & 1.73  & 6.17  & -3.59E-14 & 144.75 & 0.03  & 5.78 \bigstrut[b]\\
			\hline
			& \multicolumn{7}{c}{$J = 100$} \bigstrut\\
			\hline
			1     & 1.90  & 1.74  & 1.09  & -3.59E-14 & 199.68 & 0.02  & 6.35 \bigstrut[t]\\
			3     & 4.84  & 1.72  & 2.80  & -4.39E-14 & 227.13 & 0.01  & 6.22 \\
			5     & 8.08  & 1.73  & 4.67  & -5.08E-14 & 245.50 & 0.02  & 6.31 \\
			7     & 10.97 & 1.74  & 6.31  & -5.64E-14 & 260.92 & 0.02  & 6.26 \bigstrut[b]\\
			\hline
			& \multicolumn{7}{c}{$J = 200$} \bigstrut\\
			\hline
			1     & 1.84  & 1.75  & 1.05  & -5.66E-14 & 399.81 & 0.01  & 6.67 \bigstrut[t]\\
			3     & 4.85  & 1.73  & 2.80  & -6.81E-14 & 438.71 & 0.01  & 6.72 \\
			5     & 8.24  & 1.75  & 4.72  & -7.71E-14 & 462.72 & 0.01  & 6.75 \\
			7     & 11.35 & 1.73  & 6.55  & -8.57E-14 & 484.49 & 0.01  & 6.74 \bigstrut[b]\\
			\hline
			\hline
		\end{tabular}%
		\vspace*{-0.3cm}
		\begin{tablenotes}
			\footnotesize
			\singlespacing
			\item \leavevmode\kern-\scriptspace\kern-\labelsep 
			Note: The first column (signal) corresponds to the informational content explained by the covariance matrix of the factors measured by $\tau' \Sigma_f \tau$, where each element in the $J \times 1$ vector $\tau$ is $1/J^{1/2}$ and the second column (noise) measured by $\tau' \Sigma_u \tau$ is the variation explained by the innovations. Eigmin $\Sigma_f$, Eigmax $\Sigma_f$ are the smallest and largest eigenvalues of $\Sigma_f$ and Eigmin $\Sigma_u$, Eigmax $\Sigma_u$ are the smallest and largest eigenvalues of $\Sigma_u$, respectively. All the quantities correspond to averages across the 500 simulation repetitions.
		\end{tablenotes}
	\end{threeparttable}
	\label{tab_sn_dgp1}%
\end{table}%

Based on the third simulation design, we analyze the finite sample properties of our deep learning approaches for the case that the DGP incorporates weak factors. More precisely, we generate the data according to process \eqref{sim_dgp2}, however introduce sparsity in the $d \times J$-dimensional coefficient matrices $\alpha$ and $\beta$. Given that for any matrix $A$, $\Vert A \Vert_{l_0}$ denotes the $l_0$-norm of $A$, counting the non-zero elements in $A$, we set the cardinalities of each row  of $\alpha, \beta$ are as follows:\footnote{All the cardinalities are rounded to the smaller integer.} $\Vert \alpha_m \Vert_{l_0} = \Vert \beta_m \Vert_{l_0} = J^{0.5}$, for $m = 1$, $\Vert \alpha_m \Vert_{l_0} = \Vert \beta_m \Vert_{l_0} = J^{0.4}$, for $m = 2,3$, $\Vert \alpha_m \Vert_{l_0} = \Vert \beta_m \Vert_{l_0} = J^{0.3}$, for $m = 4,5$ and $\Vert \alpha_m \Vert_{l_0} = \Vert \beta_m \Vert_{l_0} = J^{0.2}$, for $m = 6, \cdots, d$, where we randomly select the positions of the non-zero elements in $\alpha$ and $\beta$. Moreover, similar as for the second design, we draw the non-zero coefficients in $\alpha$ and $\beta$ from $\mathcal{U}(-1, 1)$ and $\mathcal{U}(-0.5, 0.5)$, respectively.
The time dimension $n$ is set to 60, 120 and 240 for all simulations. Moreover, we consider several dimensions for $J$ and $d$. Specifically, $J \in \{50, 100, 200\}$ and $d \in \{1, 3, 5, 7\}$. The number of replications is 500.

In the following, we analyze the signal-to-noise ratio characteristics and spectral properties of the true covariance matrix of the functions of the factors $\Sigma_f$ and the covariance matrix of the innovations $\Sigma_u$ for each of the three simulation designs. Table \ref{tab_sn_dgp1} shows that the first DGP incorporates settings with a low signal-to-noise ratio (for $d = 1$) and high ratios for $d > 1$. In addition, it corresponds to a strong factor setting as the largest eigenvalue of $\Sigma_f$ increases with the rate $J$. The characteristics for the second DGP are reported in Table \ref{tab:sn_dgp2}. The results illustrate that the signal-to-noise ratio is substantially smaller compared to the first simulation design. Hence, the second DGP is well suited for analyzing environments with a very low signal-to-noise ratio ($d=1$), low signal-to-noise ratio ($d=3$) and higher signal-to-noise ratios ($d = 5,7$). Moreover, the process also refers to a strong factor framework, since the largest eigenvalue of $\Sigma_f$ diverges with the rate $J$. Finally, Table \ref{tab_sn_dgp3} reports the properties of the third DGP. Based on the evolution of the largest eigenvalue of $\Sigma_f$, we can see that this process perfectly represents the weak factor framework. In fact, compared to strong factor setting, the largest eigenvalue diverges with a considerable slower rate than $J$. Moreover, the signal-to-noise ratio is very low and stays on a stable level even for increasing number of factors $d$.

Hence, our analysis shows that all three Monte Carlo designs are well suited for measuring the performance of our deep learning models and the competing approaches for settings with a very low signal-to-noise ratio. It is important to note that in all three setting, we have that $\Sigma_f$ is rank deficient, where the $m$-th eigenvalues, for $m > d$, are zero or close to zero. This is due to the fact that all DGPs follow a (non-linear) factor structure.

\begin{table}[!t]
	\footnotesize
	\centering
\begin{threeparttable}
	\caption{Second Monte Carlo design - Signal-to-noise ratio and spectral properties}
	\begin{tabular}{c|ccccccc}
		\hline
		\hline
		$d$     & \multicolumn{1}{l}{Signal} & \multicolumn{1}{l}{Noise} & \multicolumn{1}{l}{Signal-to-Noise Ratio} & \multicolumn{1}{l}{Eigmin $\Sigma_f$} & \multicolumn{1}{l}{Eigmax $\Sigma_f$} & \multicolumn{1}{l}{Eigmin $\Sigma_u$} & \multicolumn{1}{l}{Eigmax $\Sigma_u$} \bigstrut\\
		\hline
		& \multicolumn{7}{c}{$J = 50$} \bigstrut\\
		\hline
		1     & 0.50  & 1.70  & 0.29  & -3.10E-15 & 17.03 & 0.03  & 5.80 \bigstrut[t]\\
		3     & 1.77  & 1.72  & 1.03  & -4.38E-15 & 22.34 & 0.03  & 5.85 \\
		5     & 3.33  & 1.73  & 1.93  & -4.70E-15 & 26.61 & 0.02  & 5.79 \\
		7     & 5.27  & 1.72  & 3.07  & -3.81E-15 & 31.04 & 0.03  & 5.82 \bigstrut[b]\\
		\hline
		& \multicolumn{7}{c}{$J = 100$} \bigstrut\\
		\hline
		1     & 0.50  & 1.76  & 0.29  & -4.72E-15 & 33.68 & 0.01  & 6.32 \bigstrut[t]\\
		3     & 1.74  & 1.72  & 1.01  & -6.87E-15 & 40.73 & 0.02  & 6.29 \\
		5     & 3.36  & 1.73  & 1.94  & -8.01E-15 & 46.20 & 0.01  & 6.28 \\
		7     & 5.23  & 1.73  & 3.03  & -8.64E-15 & 51.32 & 0.01  & 6.26 \bigstrut[b]\\
		\hline
		& \multicolumn{7}{c}{$J =200$} \bigstrut\\
		\hline
		1     & 0.53  & 1.74  & 0.30  & -7.15E-15 & 66.77 & 0.01  & 6.80 \bigstrut[t]\\
		3     & 1.78  & 1.74  & 1.02  & -1.06E-14 & 76.71 & 0.01  & 6.72 \\
		5     & 3.40  & 1.74  & 1.96  & -1.29E-14 & 83.53 & 0.01  & 6.71 \\
		7     & 5.26  & 1.74  & 3.02  & -1.43E-14 & 90.03 & 0.01  & 6.76 \bigstrut[b]\\
		\hline
		\hline
	\end{tabular}%
	\vspace*{-0.3cm}
	\begin{tablenotes}
		\footnotesize
		\singlespacing
		\item \leavevmode\kern-\scriptspace\kern-\labelsep 
		Note: The first column (signal) corresponds to the informational content explained by the covariance matrix of the factors measured by $\tau' \Sigma_f \tau$, where each element in the $J \times 1$ vector $\tau$ is $1/J^{1/2}$ and the second column (noise) measured by $\tau' \Sigma_u \tau$ is the variation explained by the innovations. Eigmin $\Sigma_f$, Eigmax $\Sigma_f$ are the smallest and largest eigenvalues of $\Sigma_f$ and Eigmin $\Sigma_u$, Eigmax $\Sigma_u$ are the smallest and largest eigenvalues of $\Sigma_u$, respectively. All the quantities correspond to averages across the 500 simulation repetitions.
	\end{tablenotes}
	\label{tab:sn_dgp2}
\end{threeparttable}
\end{table}%

\begin{table}[!t]
	\footnotesize
	\centering
	\caption{Third Monte Carlo design - Signal-to-noise ratio and spectral properties}
	\begin{threeparttable}
	\begin{tabular}{c|ccccccc}
		\hline
		\hline
		$d$     & \multicolumn{1}{l}{Signal} & \multicolumn{1}{l}{Noise} & \multicolumn{1}{l}{Signal-to-Noise Ratio} & \multicolumn{1}{l}{Eigmin $\Sigma_f$} & \multicolumn{1}{l}{Eigmax $\Sigma_f$} & \multicolumn{1}{l}{Eigmin $\Sigma_u$} & \multicolumn{1}{l}{Eigmax $\Sigma_u$} \bigstrut\\
		\hline
		& \multicolumn{7}{c}{$J = 50$} \bigstrut\\
		\hline
		1     & 0.07  & 1.72  & 0.04  & -5.84E-16 & 2.44  & 0.02  & 5.85 \bigstrut[t]\\
		3     & 0.13  & 1.73  & 0.08  & -6.44E-16 & 2.69  & 0.02  & 5.75 \\
		5     & 0.19  & 1.71  & 0.11  & -6.15E-16 & 2.76  & 0.03  & 5.77 \\
		7     & 0.22  & 1.73  & 0.13  & -5.57E-16 & 2.79  & 0.03  & 5.83 \bigstrut[b]\\
		\hline
		& \multicolumn{7}{c}{$J = 100$} \bigstrut\\
		\hline
		1     & 0.05  & 1.74  & 0.03  & -8.98E-16 & 3.41  & 0.02  & 6.27 \bigstrut[t]\\
		3     & 0.10  & 1.74  & 0.06  & -9.92E-16 & 3.62  & 0.02  & 6.28 \\
		5     & 0.12  & 1.75  & 0.07  & -9.47E-16 & 3.66  & 0.01  & 6.31 \\
		7     & 0.14  & 1.73  & 0.08  & -9.22E-16 & 3.72  & 0.02  & 6.25 \bigstrut[b]\\
		\hline
		& \multicolumn{7}{c}{$J =200$} \bigstrut\\
		\hline
		1     & 0.04  & 1.75  & 0.02  & -1.23E-15 & 4.65  & 0.01  & 6.71 \bigstrut[t]\\
		3     & 0.07  & 1.73  & 0.04  & -1.30E-15 & 4.80  & 0.01  & 6.72 \\
		5     & 0.08  & 1.75  & 0.05  & -1.31E-15 & 4.92  & 0.01  & 6.72 \\
		7     & 0.09  & 1.75  & 0.05  & -1.27E-15 & 4.83  & 0.01  & 6.76 \bigstrut[b]\\
		\hline
		\hline
	\end{tabular}%
	\vspace*{-0.3cm}
\begin{tablenotes}
	\footnotesize
	\singlespacing
	\item \leavevmode\kern-\scriptspace\kern-\labelsep 
	Note: The first column (signal) corresponds to the informational content explained by the covariance matrix of the factors measured by $\tau' \Sigma_f \tau$, where each element in the $J \times 1$ vector $\tau$ is $1/J^{1/2}$ and the second column (noise) measured by $\tau' \Sigma_u \tau$ is the variation explained by the innovations. Eigmin $\Sigma_f$, Eigmax $\Sigma_f$ are the smallest and largest eigenvalues of $\Sigma_f$ and Eigmin $\Sigma_u$, Eigmax $\Sigma_u$ are the smallest and largest eigenvalues of $\Sigma_u$, respectively. All the quantities correspond to averages across the 500 simulation repetitions.
\end{tablenotes}
\end{threeparttable}
	\label{tab_sn_dgp3}%
\end{table}%

We compare the precision of our deep neural network factor models (DNN-FM and SDNN-FM) with methods that are commonly used in the literature. The following models are included in the simulation study:

\begin{itemize}
	\item SFM-POET: The static factor model with observed factors, where the covariance matrix of the idiosyncratic errors is estimated based on the principal orthogonal complement thresholding method (POET) from \cite{fan2013}.
	\item DNN-FM: Our non-linear nonparametric factor model estimated based on deep neural network in Section \ref{sec_model}.
	\item SDNN-FM: Our non-linear nonparametric factor model estimated based on the s-sparse deep neural network in Section \ref{sec_factor_model}.
	\item NL-LW: The non-linear shrinkage estimator of \cite{lw2017}.
	\item SF-NL-LW: The single factor non-linear shrinkage estimator of \cite{lw2017}.
	\item RN-FM: Residual nodewise regression estimation of factor models by \cite{caner2022}.
\end{itemize} 

\subsection{Simulation results}

The simulation results for the first Monte Carlo design in \eqref{sim_dgp1} are illustrated in Tables \ref{tab_sim_dgp1_err_v2} to \ref{tab_sim_dgp1_inv_cov_v2}. Table \ref{tab_sim_dgp1_err_v2} provides the results of the static linear factor model with observed factors (SFM-POET) and our deep learning factor models (DNN-FM, SDNN-FM) in estimating the unknown true function  in \eqref{sim_dgp1}, which connects the factors $X$ with the response variables $Y$. The non-linear (NL-LW, SF-NL-LW) shrinkage estimators of \citeauthor{lw2003} only provide estimates for the covariance and precision matrix of the returns. For this reason, these methods are not included in Table \ref{tab_sim_dgp1_err_v2}. We also use the linear shrinkage estimator of \cite{lw2003} in an earlier version of this paper, however as other methods lead to much better results, hence we omitted the linear shrinkage method from Tables \ref{tab_sim_dgp1_cov_v2} and \ref{tab_sim_dgp1_inv_cov_v2}.

\begin{table}[!t]
	\footnotesize
	\centering
	\caption{Simulation results - First Monte Carlo design \\ Function estimation}
	\begin{threeparttable}
		\begin{tabular}{ccc|ccc|ccc|ccc}
			\hline
			\hline
			$d$     & $n$     & $J$     & SFM-POET & \textbf{DNN-FM} & \textbf{SDNN-FM} & $d$     & $n$     & $J$     & SFM-POET & \textbf{DNN-FM} & \textbf{SDNN-FM} \bigstrut\\
			\hline
			\multirow{9}[2]{*}{1} & 60    & 50    & 7.05  & 3.67  & 3.87  & \multirow{9}[2]{*}{5} & 60    & 50    & 6.17  & 2.46  & 2.88 \bigstrut[t]\\
			& 60    & 100   & 8.62  & 4.68  & 5.08  &       & 60    & 100   & 7.27  & 3.03  & 3.90 \\
			& 60    & 200   & 9.95  & 5.12  & 6.36  &       & 60    & 200   & 8.66  & 3.70  & 5.62 \\
			& 120   & 50    & 6.84  & 3.49  & 3.68  &       & 120   & 50    & 6.16  & 1.88  & 2.64 \\
			& 120   & 100   & 7.97  & 4.15  & 4.41  &       & 120   & 100   & 6.84  & 2.15  & 3.31 \\
			& 120   & 200   & 9.41  & 5.24  & 6.06  &       & 120   & 200   & 9.00  & 2.69  & 5.52 \\
			& 240   & 50    & 5.90  & 3.16  & 3.27  &       & 240   & 50    & 5.24  & 1.52  & 2.12 \\
			& 240   & 100   & 9.11  & 4.42  & 4.88  &       & 240   & 100   & 6.66  & 1.87  & 3.02 \\
			& 240   & 200   & 9.49  & 4.71  & 5.33  &       & 240   & 200   & 8.17  & 2.38  & 4.17 \bigstrut[b]\\
			\hline
			\multirow{9}[2]{*}{3} & 60    & 50    & 6.84  & 2.74  & 3.30  & \multirow{9}[2]{*}{7} & 60    & 50    & 5.74  & 2.67  & 2.80 \bigstrut[t]\\
			& 60    & 100   & 7.02  & 2.88  & 3.81  &       & 60    & 100   & 8.23  & 3.74  & 4.65 \\
			& 60    & 200   & 9.20  & 3.79  & 5.66  &       & 60    & 200   & 9.46  & 4.69  & 6.25 \\
			& 120   & 50    & 5.68  & 2.09  & 2.54  &       & 120   & 50    & 5.80  & 1.76  & 2.62 \\
			& 120   & 100   & 6.27  & 2.43  & 3.15  &       & 120   & 100   & 6.77  & 2.11  & 3.47 \\
			& 120   & 200   & 8.30  & 3.07  & 4.69  &       & 120   & 200   & 7.61  & 2.36  & 5.05 \\
			& 240   & 50    & 6.72  & 2.04  & 2.77  &       & 240   & 50    & 4.90  & 1.24  & 1.95 \\
			& 240   & 100   & 8.13  & 2.71  & 3.79  &       & 240   & 100   & 6.99  & 1.66  & 3.25 \\
			& 240   & 200   & 8.29  & 2.79  & 4.25  &       & 240   & 200   & 8.06  & 1.94  & 4.14 \bigstrut[b]\\
			\hline
			\hline
		\end{tabular}%
		\vspace*{-0.3cm}
		\begin{tablenotes}
			\footnotesize
			\singlespacing
			\item \leavevmode\kern-\scriptspace\kern-\labelsep 
			Note: The quantities in table relate to the error metric used in Theorem \ref{t1} and Theorem  \ref{thm1} and correspond to the maximum difference between the estimated function and true function ($\sum_{m = 1}^{d} \1_{\left\{m \text{ is even} \right\}} \beta_{m,j} X_{m,i} + \1_{\left\{m \text{ is odd} \right\}} \beta_{m,j} X_{m,i}^2$) in \eqref{sim_dgp1}, for $j = 1, \cdots, J$.
		\end{tablenotes}
	\end{threeparttable}
	\label{tab_sim_dgp1_err_v2}%
\end{table}%

The quantities in Table \ref{tab_sim_dgp1_err_v2} relate to the error metric used in Theorem \ref{t1} and Theorem \ref{thm1}, which corresponds to the maximum difference between the estimated and true function. The results indicate that our DNN-FM and SDNN-FM uniformly provide more precise function estimates compared to the SFM-POET. Consequently, these methods are better suited for measuring non-linear transformations of the observed factors and complex transferring mechanisms between the factors and the observed variables. As predicted by the theory, the estimation error of DNN-FM and SDNN-FM converges to zero as $n$ increases, e.g.,\ for the DNN-FM with $d = 7$ and $J = 200$, the error rate decreases from 4.69 ($n = 60$) to 1.94 ($n = 240$). This result is valid for different dimensions of the number of factors $d$ and number of variables $J$. Moreover, the error rates of SFM-POET are sensitive to an increase in the number of factors and occasionally rise as $d$ gets large. In contrast to that the error rates of DNN-FM and SDNN-FM are in most of the cases unaffected by an increase of $d$, e.g.,\ for the SDNN-FM with $n = 240, J= 50$, the error rate gradually decreases from 3.27 ($d = 1$) to 1.95 ($d = 7$). This result is in line with the theory in Theorem \ref{thm1}. Generally, the error rates of the DNN-FM are smaller compared to the SDNN-FM. This result is anticipated as our simulation study focuses on an in-sample analysis and the DNN-FM with fully connected units offers a more precise function approximation. However, this is not necessarily valid for an out-of-sample analysis, as the DNN-FM might be overfitting compared to a deep neural network with sparsely connected units.


\begin{table}[!t]
	\footnotesize
	
	\caption{Simulation results - First Monte Carlo design \\ Covariance matrix estimation}
	
	\centering
	\begin{threeparttable}
		
		\begin{tabular}{ccc|cccccc}
			\hline
			\hline
			$d$     & $n$     & $J$     & SFM-POET & \textbf{DNN-FM} & \textbf{SDNN-FM}  & NL-LW & SF-NL-LW & RN-FM \bigstrut\\
			\hline
			\multirow{9}[2]{*}{1} & 60    & 50    & 2.13  & 1.37  & 1.68  & 1.32  & 1.38  & 2.00 \bigstrut[t]\\
			& 60    & 100   & 2.62  & 1.71  & 2.11  & 1.60  & 1.74  & 2.24 \\
			& 60    & 200   & 3.10  & 2.01  & 2.61  & 1.90  & 2.05  & 2.66 \\
			& 120   & 50    & 2.20  & 1.31  & 1.68  & 1.30  & 1.33  & 1.76 \\
			& 120   & 100   & 2.72  & 1.64  & 2.13  & 1.57  & 1.59  & 2.27 \\
			& 120   & 200   & 3.18  & 1.90  & 2.57  & 1.87  & 2.08  & 2.61 \\
			& 240   & 50    & 2.25  & 1.18  & 1.53  & 1.32  & 1.35  & 1.44 \\
			& 240   & 100   & 2.77  & 1.47  & 1.96  & 1.57  & 1.59  & 2.14 \\
			& 240   & 200   & 3.24  & 1.73  & 2.38  & 1.86  & 1.89  & 2.56 \bigstrut[b]\\
			\hline
			\multirow{9}[2]{*}{3} & 60    & 50    & 1.46  & 0.91  & 1.28  & 1.60  & 1.64  & 1.83 \bigstrut[t]\\
			& 60    & 100   & 1.73  & 1.10  & 1.53  & 1.89  & 2.00  & 1.77 \\
			& 60    & 200   & 2.02  & 1.29  & 1.89  & 2.17  & 2.35  & 2.04 \\
			& 120   & 50    & 1.53  & 0.78  & 1.26  & 1.54  & 1.56  & 1.48 \\
			& 120   & 100   & 1.80  & 0.95  & 1.52  & 1.80  & 1.80  & 1.70 \\
			& 120   & 200   & 2.09  & 1.15  & 1.87  & 2.11  & 2.23  & 1.97 \\
			& 240   & 50    & 1.57  & 0.68  & 1.16  & 1.51  & 1.53  & 1.17 \\
			& 240   & 100   & 1.85  & 0.83  & 1.38  & 1.79  & 1.78  & 1.66 \\
			& 240   & 200   & 2.15  & 1.01  & 1.67  & 2.12  & 2.11  & 1.95 \bigstrut[b]\\
			\hline
			\multirow{9}[2]{*}{5} & 60    & 50    & 1.15  & 0.77  & 1.04  & 1.60  & 1.63  & 1.58 \bigstrut[t]\\
			& 60    & 100   & 1.36  & 0.90  & 1.23  & 1.94  & 1.98  & 1.59 \\
			& 60    & 200   & 1.58  & 1.07  & 1.50  & 2.27  & 2.27  & 1.79 \\
			& 120   & 50    & 1.18  & 0.61  & 1.01  & 1.57  & 1.58  & 1.30 \\
			& 120   & 100   & 1.39  & 0.76  & 1.22  & 1.83  & 1.85  & 1.54 \\
			& 120   & 200   & 1.61  & 0.90  & 1.48  & 2.19  & 2.20  & 1.72 \\
			& 240   & 50    & 1.22  & 0.52  & 0.95  & 1.55  & 1.55  & 1.07 \\
			& 240   & 100   & 1.43  & 0.64  & 1.11  & 1.82  & 1.81  & 1.51 \\
			& 240   & 200   & 1.66  & 0.76  & 1.29  & 2.14  & 2.11  & 1.75 \bigstrut[b]\\
			\hline
			\multirow{9}[2]{*}{7} & 60    & 50    & 1.01  & 0.71  & 0.89  & 1.58  & 1.61  & 1.48 \bigstrut[t]\\
			& 60    & 100   & 1.16  & 0.84  & 1.06  & 1.88  & 1.91  & 1.51 \\
			& 60    & 200   & 1.33  & 0.95  & 1.27  & 2.24  & 2.27  & 1.66 \\
			& 120   & 50    & 0.99  & 0.53  & 0.87  & 1.53  & 1.54  & 1.22 \\
			& 120   & 100   & 1.15  & 0.64  & 1.04  & 1.85  & 1.86  & 1.45 \\
			& 120   & 200   & 1.33  & 0.74  & 1.26  & 2.21  & 2.22  & 1.56 \\
			& 240   & 50    & 1.03  & 0.44  & 0.83  & 1.50  & 1.51  & 1.00 \\
			& 240   & 100   & 1.20  & 0.54  & 0.97  & 1.83  & 1.83  & 1.40 \\
			& 240   & 200   & 1.37  & 0.64  & 1.07  & 2.20  & 2.21  & 1.57 \bigstrut[b]\\
			\hline
			\hline
		\end{tabular}%
		\vspace*{-0.3cm}
		\begin{tablenotes}
			\footnotesize
			\singlespacing
			\item \leavevmode\kern-\scriptspace\kern-\labelsep 
			Note: The quantities in the table refer to the error metric used in Theorem \ref{covret}  and Theorem \ref{ds}\ref{ds_ii}, which corresponds to the maximum matrix norm of the difference between the estimated covariance matrices $\hat{\Sigma}_y$, $\hat{\Sigma}_y^s$ and the true covariance matrix $\Sigma_y$. The deep neural network factor model (DNN-FM) and sparse deep neural network factor model (SDNN-FM) are compared to the static factor model with observed factors and POET estimator from \cite{fan2013} (SFM-POET), the non-linear shrinkage estimator (NL-LW), the single factor non-linear shrinkage estimator (SF-NL-LW) of \cite{lw2017} and the residual nodewise regression estimation of factor models by \cite{caner2022}.
		\end{tablenotes}
	\end{threeparttable}
	\label{tab_sim_dgp1_cov_v2}%
\end{table}%


Table \ref{tab_sim_dgp1_cov_v2} illustrates the simulation results for estimating the true covariance matrix of the data $\Sigma_y$ corresponding to the first Monte Carlo design. The quantities in the table refer to the error metric used in Theorem \ref{covret} and Theorem \ref{ds}\ref{ds_ii}, which corresponds to the maximum matrix norm of the difference between the covariance matrix estimated based on the DNN-FM $\hat{\Sigma}_y$ and $\Sigma_y$ and the covariance matrix based on the SDNN-FM $\hat{\Sigma}_y^s$ and $\Sigma_y$, respectively. The results indicate that both our DNN-FM and SDNN-FM offer the most precise covariance matrix estimates compared to the competing approaches when $d=3,5,7$. For example at $n=240$, with $J=100$, DNN-FM has estimation error of 0.54, whereas the next best method (non neural network based) SFM-POET has an error of 1.20. Also, as in line with Theorem 3, for the DNN-FM, with $d = 7, J = 200$, the error rates gradually decrease from 0.95 ($n = 60$) to 0.64 ($n = 240$). Generally, the DNN-FM offers slightly more precise results compared to the SDNN-FM, which indicates that the richer parametrization of the DNN-FM is more efficient in an in-sample experiment. 
These results reinforce the conclusion that the rigid linear relationship of the SFM-POET and RN-FM is too restrictive for measuring more complex relations as in \eqref{sim_dgp1}.

\begin{table}[!t]
	\footnotesize
	
	\caption{Simulation results - First Monte Carlo design \\ Precision matrix estimation}
	
	\centering
	\begin{threeparttable}
		
		\begin{tabular}{ccc|cccccc}
			\hline
			\hline
			$d$     & $n$     & $J$     & SFM-POET & \textbf{DNN-FM} & \textbf{SDNN-FM} & NL-LW & SF-NL-LW & RN-FM \bigstrut\\
			\hline
			\multirow{9}[2]{*}{1} & 60    & 50    & 1.75  & 2.16  & 1.94  & 20.49 & 25.88 & 235.66 \bigstrut[t]\\
			& 60    & 100   & 1.90  & 2.30  & 2.01  & 2.53  & 4.37  & 8.65 \\
			& 60    & 200   & 2.20  & 2.64  & 2.19  & 2.14  & 4.46  & 4.69 \\
			& 120   & 50    & 1.43  & 1.69  & 1.53  & 27.20 & 20.93 & 12.65 \\
			& 120   & 100   & 1.56  & 1.85  & 1.66  & 43.54 & 33.55 & 6.66 \\
			& 120   & 200   & 1.77  & 2.11  & 1.81  & 2.76  & 4.45  & 2.63 \\
			& 240   & 50    & 1.30  & 1.34  & 1.23  & 31.36 & 26.51 & 8.69 \\
			& 240   & 100   & 1.38  & 1.47  & 1.32  & 68.43 & 42.16 & 6.07 \\
			& 240   & 200   & 1.55  & 1.69  & 1.47  & 68.98 & 151.23 & 2.85 \bigstrut[b]\\
			\hline
			\multirow{9}[2]{*}{3} & 60    & 50    & 2.03  & 3.86  & 2.19  & 10.51 & 32.65 & 966.47 \bigstrut[t]\\
			& 60    & 100   & 2.40  & 4.65  & 2.22  & 4.26  & 9.14  & 12.88 \\
			& 60    & 200   & 2.70  & 5.21  & 2.12  & 3.62  & 9.65  & 7.99 \\
			& 120   & 50    & 1.65  & 2.90  & 1.70  & 23.67 & 28.07 & 183.20 \\
			& 120   & 100   & 1.87  & 3.37  & 1.78  & 33.67 & 76.04 & 125.16 \\
			& 120   & 200   & 2.07  & 3.75  & 1.70  & 4.79  & 9.63  & 4.15 \\
			& 240   & 50    & 1.53  & 2.19  & 1.39  & 29.46 & 38.66 & 24.91 \\
			& 240   & 100   & 1.66  & 2.54  & 1.46  & 67.05 & 78.68 & 63.66 \\
			& 240   & 200   & 1.77  & 2.84  & 1.53  & 131.20 & 154.48 & 4.12 \bigstrut[b]\\
			\hline
			\multirow{9}[2]{*}{5} & 60    & 50    & 2.06  & 5.49  & 2.08  & 14.50 & 20.20 & 2662.03 \bigstrut[t]\\
			& 60    & 100   & 2.36  & 7.42  & 1.97  & 5.44  & 12.25 & 17.53 \\
			& 60    & 200   & 2.78  & 9.18  & 1.84  & 5.02  & 14.27 & 10.45 \\
			& 120   & 50    & 1.77  & 4.26  & 1.55  & 24.20 & 29.27 & 254.63 \\
			& 120   & 100   & 1.84  & 5.08  & 1.42  & 19.79 & 41.55 & 16.30 \\
			& 120   & 200   & 2.03  & 6.07  & 1.36  & 6.43  & 14.07 & 5.81 \\
			& 240   & 50    & 1.67  & 3.10  & 1.24  & 31.66 & 41.35 & 46.48 \\
			& 240   & 100   & 1.68  & 3.70  & 1.15  & 72.07 & 96.80 & 415.53 \\
			& 240   & 200   & 1.74  & 4.37  & 1.06  & 50.98 & 94.74 & 5.58 \bigstrut[b]\\
			\hline
			\multirow{9}[2]{*}{7} & 60    & 50    & 2.18  & 5.18  & 2.31  & 16.29 & 18.67 & 3423.15 \bigstrut[t]\\
			& 60    & 100   & 2.26  & 7.27  & 2.06  & 6.08  & 13.14 & 26.85 \\
			& 60    & 200   & 2.53  & 10.11 & 1.96  & 5.80  & 16.03 & 17.61 \\
			& 120   & 50    & 1.85  & 4.95  & 1.73  & 27.44 & 35.33 & 221.15 \\
			& 120   & 100   & 1.87  & 6.49  & 1.50  & 34.42 & 59.73 & 53.50 \\
			& 120   & 200   & 1.91  & 8.01  & 1.54  & 7.70  & 16.68 & 7.09 \\
			& 240   & 50    & 1.75  & 3.84  & 1.27  & 35.37 & 42.70 & 184.32 \\
			& 240   & 100   & 1.76  & 4.71  & 0.97  & 77.23 & 86.94 & 325.20 \\
			& 240   & 200   & 1.77  & 5.73  & 0.77  & 29.22 & 66.89 & 7.92 \bigstrut[b]\\
			\hline
			\hline
		\end{tabular}%
		\vspace*{-0.3cm}
		\begin{tablenotes}
			\footnotesize
			\singlespacing
			\item \leavevmode\kern-\scriptspace\kern-\labelsep 
			Note: The quantities in the table represent the error metric of Theorem \ref{thm4} and Theorem \ref{ds}\ref{ds_iii}, which measures the spectral norm of the difference between the estimated and true precision matrix. The deep neural network factor model (DNN-FM) and sparse deep neural network factor model (SDNN-FM) are compared to the static factor model with observed factors and POET estimator from \cite{fan2013} (SFM-POET), the non-linear shrinkage estimator (NL-LW), the single factor non-linear shrinkage estimator (SF-NL-LW) of \cite{lw2017} and the residual nodewise regression estimation of factor models by \cite{caner2022}.
		\end{tablenotes}
	\end{threeparttable}
	\label{tab_sim_dgp1_inv_cov_v2}%
\end{table}%


Table \ref{tab_sim_dgp1_inv_cov_v2} provides the results of estimating the true data precision matrix $\Sigma_y^{-1}$. The quantities in the table represent the error metric of Theorem \ref{thm4} and Theorem \ref{ds}\ref{ds_iii}, which measures the spectral norm of the difference between the estimated and true precision matrix. The results indicate that both the DNN-FM and SDNN-FM consistently estimate the true precision matrix as $n$ increases, e.g.,for the SDNN-FM with $d = 7, J = 200$, the error rates rapidly decrease from 1.96 ($n=60$) to 0.77 ($n =240$), which is in line with the theory in Theorem \ref{ds}\ref{ds_iii}. Compared to the previous results related to the function and covariance matrix estimation, the SDNN-FM is generally more precise compared to the DNN-FM in estimating the precision matrix. This result shows that the sparsity in the deep neural network weights stabilizes the spectral distribution of the estimated precision matrix. Moreover, the SDNN-FM generally outperforms SFM-POET across different sample size combinations and number of factors, except $d=1$, and few cases when   $d=3, 5, 7$. To be specific, SDNN-FM has the lowest estimation error in 8 out of 9 setups (n,J combinations) in $d=5, 7$, and 7 setups out of 9 when $d=3$. In addition, the results show that NL-LW and SF-NL-LW and RN-FM lead to high error rates in terms of the spectral norm. Note that precision matrix estimation in RN-FM assumes a linear factor structure but estimates the precision matrix errors by a novel residual nodewise regression technique
and then uses Sherman-Morrison-Woodbury formula.
Moreover, both NL-LW and SF-NL-LW, RN-FM are unstable for different specifications of $d$.

\begin{table}[!t]
	\footnotesize
	\centering
	\caption{Simulation results - Second Monte Carlo design \\ Function estimation}
	\begin{threeparttable}
		\begin{tabular}{ccc|ccc|ccc|ccc}
			\hline
			\hline
			$d$     & $n$     & $J$     & SFM-POET & \textbf{DNN-FM} & \textbf{SDNN-FM} & $d$     & $n$     & $J$     & SFM-POET & \textbf{DNN-FM} & \textbf{SDNN-FM} \bigstrut\\
			\hline
			\multirow{9}[2]{*}{1} & 60    & 50    & 6.77  & 6.55  & 6.47  & \multirow{9}[2]{*}{5} & 60    & 50    & 6.46  & 5.48  & 5.44 \bigstrut[t]\\
			& 60    & 100   & 8.38  & 8.20  & 8.11  &       & 60    & 100   & 7.51  & 6.86  & 6.62 \\
			& 60    & 200   & 9.89  & 9.71  & 9.02  &       & 60    & 200   & 9.27  & 8.27  & 7.61 \\
			& 120   & 50    & 6.68  & 6.37  & 6.42  &       & 120   & 50    & 6.53  & 4.94  & 5.32 \\
			& 120   & 100   & 8.16  & 7.88  & 8.00  &       & 120   & 100   & 7.18  & 5.69  & 6.20 \\
			& 120   & 200   & 10.06 & 8.81  & 9.29  &       & 120   & 200   & 8.36  & 6.81  & 7.53 \\
			& 240   & 50    & 6.81  & 6.56  & 6.59  &       & 240   & 50    & 5.87  & 4.10  & 4.86 \\
			& 240   & 100   & 8.39  & 8.10  & 8.26  &       & 240   & 100   & 7.16  & 5.24  & 6.28 \\
			& 240   & 200   & 9.47  & 9.14  & 8.81  &       & 240   & 200   & 8.25  & 6.13  & 7.76 \bigstrut[b]\\
			\hline
			\multirow{9}[2]{*}{3} & 60    & 50    & 6.59  & 5.94  & 5.77  & \multirow{9}[2]{*}{7} & 60    & 50    & 6.45  & 5.43  & 5.18 \bigstrut[t]\\
			& 60    & 100   & 7.92  & 7.52  & 7.26  &       & 60    & 100   & 8.11  & 6.75  & 6.55 \\
			& 60    & 200   & 9.51  & 8.95  & 9.26  &       & 60    & 200   & 9.56  & 8.22  & 8.61 \\
			& 120   & 50    & 6.11  & 5.29  & 5.31  &       & 120   & 50    & 5.95  & 4.30  & 4.85 \\
			& 120   & 100   & 7.43  & 6.39  & 6.76  &       & 120   & 100   & 7.40  & 5.48  & 6.27 \\
			& 120   & 200   & 9.00  & 8.11  & 9.38  &       & 120   & 200   & 8.41  & 6.55  & 8.50 \\
			& 240   & 50    & 6.03  & 5.15  & 5.29  &       & 240   & 50    & 6.40  & 3.79  & 5.05 \\
			& 240   & 100   & 7.16  & 6.11  & 6.61  &       & 240   & 100   & 7.01  & 4.39  & 6.18 \\
			& 240   & 200   & 8.69  & 7.52  & 8.11  &       & 240   & 200   & 7.97  & 5.37  & 7.42 \bigstrut[b]\\
			\hline
			\hline
		\end{tabular}%
		\vspace*{-0.3cm}
		\begin{tablenotes}
			\footnotesize
			\singlespacing
			\item \leavevmode\kern-\scriptspace\kern-\labelsep 
			Note: The quantities in table relate to the error metric used in Theorem \ref{t1} and Theorem \ref{thm1} and correspond to the maximum difference between the estimated function $\hat{\alpha}' X + \hat{\beta}'\hat{\psi}(X)$ and true function $\alpha' X + \beta'\psi(X)$ in \eqref{sim_dgp2}.
		\end{tablenotes}
	\end{threeparttable}
	\label{tab_sim_dgp2_err}%
\end{table}%


It is important to note that our SDNN-FM is not affected by an increase of the number of included factors, as predicted by our theory. Specifically, the error rates  either hardly change or decrease as $d$ increases. E.g., for $n=240$, $J=200$, the error rate decreases from 1.47 ($d = 1$) to 0.77 ($d=7$). DNN-FM is affected by the increasing number of factors.

In order to verify the robustness of the results from the first simulation design, we analyze in the following the Monte Carlo outcomes for the second simulation design in \eqref{sim_dgp2}, which incorporates linear, as well as non-linear effects and pairwise interactions in $X$ on the dependent variable $Y$ in accordance with Section 2. The results are provided in the Tables \ref{tab_sim_dgp2_err} to \ref{tab_sim_dgp2_inv_cov}. 
The simulation results are qualitatively similar to the ones that we obtain for the first Monte Carlo design. Specifically, Table \ref{tab_sim_dgp2_err} shows that both DNN-FM and SDNN-FM consistently estimate the true functional form in \eqref{sim_dgp2} as $n$ increases. Moreover, they  provide more precise estimates than the SFM-POET, showing that DNN-FM and SDNN-FM are better suited to capture non-linear transformations in the observed factors, except one setup at $n=120, J=200, d=7$.


\begin{table}[!t]
	\footnotesize
	\centering
	\caption{Simulation results - Second Monte Carlo design \\ Covariance matrix estimation}
	\begin{threeparttable}
		\begin{tabular}{ccc|cccccc}
			\hline
			\hline
			$d$     & $n$     & $J$     & SFM-POET & \textbf{DNN-FM} & \textbf{SDNN-FM} & NL-LW & SF-NL-LW & RN-FM \bigstrut\\
			\hline
			\multirow{9}[2]{*}{1} & 60    & 50    & 0.87  & 0.86  & 0.86  & 0.60  & 0.62  & 1.17 \bigstrut[t]\\
			& 60    & 100   & 0.95  & 0.94  & 0.95  & 0.83  & 0.79  & 1.03 \\
			& 60    & 200   & 1.05  & 1.04  & 1.06  & 1.11  & 0.95  & 0.97 \\
			& 120   & 50    & 0.87  & 0.72  & 0.73  & 0.46  & 0.44  & 0.77 \\
			& 120   & 100   & 0.94  & 0.82  & 0.84  & 0.67  & 0.56  & 0.83 \\
			& 120   & 200   & 1.04  & 0.92  & 0.96  & 0.97  & 0.75  & 0.79 \\
			& 240   & 50    & 0.87  & 0.55  & 0.56  & 0.33  & 0.32  & 0.51 \\
			& 240   & 100   & 0.95  & 0.62  & 0.64  & 0.51  & 0.41  & 0.76 \\
			& 240   & 200   & 1.04  & 0.71  & 0.74  & 0.79  & 0.56  & 0.77 \bigstrut[b]\\
			\hline
			\multirow{9}[2]{*}{3} & 60    & 50    & 0.71  & 0.68  & 0.68  & 0.53  & 0.67  & 1.06 \bigstrut[t]\\
			& 60    & 100   & 0.76  & 0.74  & 0.74  & 0.64  & 0.81  & 1.17 \\
			& 60    & 200   & 0.83  & 0.83  & 0.83  & 0.79  & 0.93  & 1.14 \\
			& 120   & 50    & 0.69  & 0.56  & 0.59  & 0.45  & 0.51  & 0.71 \\
			& 120   & 100   & 0.74  & 0.63  & 0.67  & 0.53  & 0.59  & 0.67 \\
			& 120   & 200   & 0.80  & 0.70  & 0.79  & 0.67  & 0.71  & 0.66 \\
			& 240   & 50    & 0.69  & 0.43  & 0.46  & 0.39  & 0.42  & 0.51 \\
			& 240   & 100   & 0.74  & 0.49  & 0.53  & 0.46  & 0.50  & 0.70 \\
			& 240   & 200   & 0.80  & 0.55  & 0.63  & 0.57  & 0.58  & 0.65 \bigstrut[b]\\
			\hline
			\multirow{9}[2]{*}{5} & 60    & 50    & 0.61  & 0.60  & 0.55  & 0.56  & 0.73  & 1.01 \bigstrut[t]\\
			& 60    & 100   & 0.67  & 0.68  & 0.61  & 0.64  & 0.85  & 1.25 \\
			& 60    & 200   & 0.74  & 0.76  & 0.69  & 0.73  & 0.96  & 1.25 \\
			& 120   & 50    & 0.54  & 0.46  & 0.49  & 0.52  & 0.58  & 0.58 \\
			& 120   & 100   & 0.59  & 0.51  & 0.56  & 0.59  & 0.67  & 0.54 \\
			& 120   & 200   & 0.64  & 0.57  & 0.64  & 0.68  & 0.78  & 0.54 \\
			& 240   & 50    & 0.54  & 0.35  & 0.40  & 0.49  & 0.51  & 0.48 \\
			& 240   & 100   & 0.59  & 0.39  & 0.46  & 0.57  & 0.60  & 0.59 \\
			& 240   & 200   & 0.63  & 0.44  & 0.55  & 0.64  & 0.68  & 0.54 \bigstrut[b]\\
			\hline
			\multirow{9}[2]{*}{7} & 60    & 50    & 0.56  & 0.57  & 0.51  & 0.58  & 0.74  & 1.00 \bigstrut[t]\\
			& 60    & 100   & 0.64  & 0.64  & 0.54  & 0.67  & 0.87  & 1.25 \\
			& 60    & 200   & 0.72  & 0.74  & 0.60  & 0.77  & 0.99  & 1.29 \\
			& 120   & 50    & 0.46  & 0.44  & 0.44  & 0.55  & 0.62  & 0.47 \\
			& 120   & 100   & 0.50  & 0.49  & 0.47  & 0.65  & 0.72  & 0.47 \\
			& 120   & 200   & 0.54  & 0.54  & 0.53  & 0.73  & 0.82  & 0.48 \\
			& 240   & 50    & 0.44  & 0.32  & 0.37  & 0.53  & 0.56  & 0.44 \\
			& 240   & 100   & 0.47  & 0.35  & 0.41  & 0.63  & 0.66  & 0.49 \\
			& 240   & 200   & 0.51  & 0.38  & 0.48  & 0.70  & 0.73  & 0.55 \bigstrut[b]\\
			\hline
			\hline
		\end{tabular}%
		\vspace*{-0.3cm}
		\begin{tablenotes}
			\footnotesize
			\singlespacing
			\item \leavevmode\kern-\scriptspace\kern-\labelsep 
			Note: The quantities in the table refer to the error metric used in Theorem \ref{covret}  and Theorem \ref{ds}\ref{ds_ii}, which corresponds to the maximum matrix norm of the difference between the estimated covariance matrix $\hat{\Sigma}_y$ and $\Sigma_y$, and $\hat{\Sigma}_y^s$ and $\Sigma_y$ respectively. The deep neural network factor model (DNN-FM) and sparse deep neural network factor model (SDNN-FM) are compared to the static factor model with observed factors and POET estimator from \cite{fan2013} (SFM-POET), the non-linear shrinkage estimator (NL-LW), the single factor non-linear shrinkage estimator (SF-NL-LW) of \cite{lw2017} and the residual nodewise regression estimation of factor models by \cite{caner2022}.
		\end{tablenotes}
	\end{threeparttable}
	\label{tab_sim_dgp2_cov}%
\end{table}%


Table \ref{tab_sim_dgp2_cov} illustrates the error rates in terms of the maximum matrix norm of the different approaches in estimating the true covariance matrix. The results are similar to the first simulation design. In fact, DNN-FM and SDNN-FM offer consistent estimates of the covariance matrix and SDNN-FM uniformly outperforms the SFM-POET for different combinations of $d, n$ and $J$. Moreover, it leads to more precise estimates compared to the non-linear shrinkage estimators of \cite{lw2017}, for $d > 3$. For $d = 1, 3$, NL-LW achieves a similar precision as our deep neural network factor models and in most cases offers better performance compared to the remaining competing methods. Furthermore, we observe that the error rates of the our approaches are not affected by an increase in the number of factors, which is in accordance with our theory. 
The simulation results for estimating the precision matrix are outlined in Table \ref{tab_sim_dgp2_inv_cov}.  Table \ref{tab_sim_dgp2_inv_cov} shows when we increase $n$ our estimation errors decrease. To give an example, at $d=3$ DNN-FM has the error 1.59 at $n=60, J=50$, and when $n=240$, $J=50$ the error decreases to 1.10. We see that SDNN-FM and SFM-POET do generally well among all methods.



\begin{table}[!t]
	\footnotesize
	\centering
	\caption{Simulation results - Second Monte Carlo design \\ Precision matrix estimation}
	\begin{threeparttable}
		\begin{tabular}{ccc|cccccc}
			\hline
			\hline
			$d$     & $n$     & $J$     & SFM-POET & \textbf{DNN-FM} & \textbf{SDNN-FM} & NL-LW & SF-NL-LW & RN-FM \bigstrut\\
			\hline
			\multirow{9}[2]{*}{1} & 60    & 50    & 1.10  & 1.08  & 1.06  & 5.76  & 8.82  & 119.70 \bigstrut[t]\\
			& 60    & 100   & 1.20  & 1.16  & 1.13  & 1.05  & 1.35  & 23.64 \\
			& 60    & 200   & 1.29  & 1.26  & 1.20  & 0.89  & 1.29  & 20.47 \\
			& 120   & 50    & 0.97  & 0.86  & 0.85  & 8.20  & 7.79  & 6.66 \\
			& 120   & 100   & 1.08  & 0.93  & 0.97  & 8.82  & 14.20 & 13.93 \\
			& 120   & 200   & 1.16  & 1.01  & 1.11  & 1.12  & 1.28  & 1.38 \\
			& 240   & 50    & 0.92  & 0.67  & 0.68  & 7.10  & 7.07  & 4.22 \\
			& 240   & 100   & 1.04  & 0.74  & 0.87  & 17.33 & 14.59 & 4.00 \\
			& 240   & 200   & 1.13  & 0.81  & 1.06  & 22.74 & 34.43 & 1.41 \bigstrut[b]\\
			\hline
			\multirow{9}[2]{*}{3} & 60    & 50    & 1.29  & 1.59  & 1.37  & 2.08  & 3.31  & 235.53 \bigstrut[t]\\
			& 60    & 100   & 1.42  & 1.75  & 1.42  & 1.21  & 1.51  & 53.61 \\
			& 60    & 200   & 1.51  & 1.90  & 1.28  & 1.13  & 1.53  & 55.45 \\
			& 120   & 50    & 1.10  & 1.21  & 1.12  & 4.46  & 4.71  & 4.24 \\
			& 120   & 100   & 1.23  & 1.33  & 1.23  & 3.67  & 6.15  & 1.86 \\
			& 120   & 200   & 1.32  & 1.44  & 1.16  & 1.29  & 1.44  & 1.51 \\
			& 240   & 50    & 1.03  & 0.90  & 0.97  & 4.71  & 4.50  & 2.74 \\
			& 240   & 100   & 1.16  & 1.01  & 1.12  & 10.16 & 9.24  & 2.06 \\
			& 240   & 200   & 1.25  & 1.10  & 1.10  & 4.46  & 3.01  & 1.52 \bigstrut[b]\\
			\hline
			\multirow{9}[2]{*}{5} & 60    & 50    & 1.40  & 2.26  & 1.61  & 1.77  & 2.40  & 404.42 \bigstrut[t]\\
			& 60    & 100   & 1.56  & 2.60  & 1.60  & 1.36  & 1.68  & 137.05 \\
			& 60    & 200   & 1.64  & 2.75  & 1.33  & 1.36  & 1.74  & 85.91 \\
			& 120   & 50    & 1.14  & 1.78  & 1.28  & 2.86  & 3.18  & 1.98 \\
			& 120   & 100   & 1.29  & 2.06  & 1.34  & 3.63  & 3.95  & 1.51 \\
			& 120   & 200   & 1.39  & 2.19  & 1.18  & 1.55  & 1.70  & 1.52 \\
			& 240   & 50    & 1.04  & 1.27  & 1.11  & 3.04  & 2.91  & 2.02 \\
			& 240   & 100   & 1.19  & 1.48  & 1.21  & 6.91  & 6.92  & 1.58 \\
			& 240   & 200   & 1.29  & 1.61  & 1.10  & 4.67  & 4.11  & 1.46 \bigstrut[b]\\
			\hline
			\multirow{9}[2]{*}{7} & 60    & 50    & 1.46  & 2.86  & 1.75  & 1.44  & 1.77  & 64182.15 \bigstrut[t]\\
			& 60    & 100   & 1.63  & 3.28  & 1.66  & 1.43  & 1.75  & 116.88 \\
			& 60    & 200   & 1.74  & 3.60  & 1.35  & 1.51  & 1.92  & 114.56 \\
			& 120   & 50    & 1.14  & 2.44  & 1.34  & 2.15  & 2.32  & 1.55 \\
			& 120   & 100   & 1.30  & 2.93  & 1.34  & 2.67  & 2.55  & 1.46 \\
			& 120   & 200   & 1.41  & 3.23  & 1.17  & 1.78  & 1.93  & 1.51 \\
			& 240   & 50    & 1.01  & 1.77  & 1.14  & 2.11  & 2.16  & 1.47 \\
			& 240   & 100   & 1.17  & 2.16  & 1.19  & 5.40  & 5.41  & 1.39 \\
			& 240   & 200   & 1.29  & 2.41  & 1.08  & 4.70  & 5.16  & 1.40 \bigstrut[b]\\
			\hline
			\hline
		\end{tabular}%
		\vspace*{-0.3cm}
		\begin{tablenotes}
			\footnotesize
			\singlespacing
			\item \leavevmode\kern-\scriptspace\kern-\labelsep 
			Note: The quantities in the table represent the error metric of Theorem \ref{thm4} and  Theorem \ref{ds}\ref{ds_iii}, which measures the spectral norm of the difference between the estimated and true precision matrix. The deep neural network factor model (DNN-FM) and sparse deep neural network factor model (SDNN-FM) are compared to the static factor model with observed factors and POET estimator from \cite{fan2013} (SFM-POET), the non-linear shrinkage estimator (NL-LW), the single factor non-linear shrinkage estimator (SF-NL-LW) of \cite{lw2017} and the residual nodewise regression estimation of factor models by \cite{caner2022}.
		\end{tablenotes}
	\end{threeparttable}
	\label{tab_sim_dgp2_inv_cov}%
\end{table}%


The results for our third Monte Carlo design, which incorporates weak factors in the data generating process are reported in Tables \ref{tab_sim_dgp3_err} to \ref{tab_sim_dgp3_inv_cov}. Table \ref{tab_sim_dgp3_err} results are very similar to other tables considering function estimation. Table \ref{tab_sim_dgp3_cov}
 provides consistent estimation  of covariance matrix by our methods. In Table \ref{tab_sim_dgp3_inv_cov} our methods dominate all others within $n=120$, $n=240$ setups with differing $J$, differing $d$. To give an example, at $n=240, J=200$, with $d=7$ SDNN-FM has 0.69 error, DNN-FM has 0.81, and third best SFM-POET has 0.90 as an estimation error.

 \subsection{Brief Analysis of Low Signal-To-Noise Ratio and Weak Factors}

 Here in this subsection we briefly look at precision matrix estimation errors with an eye on low(very low) signal-to-noise ratio and weak factor setups. These specifically correspond to Table \ref{tab_sim_dgp1_inv_cov_v2},  $d=1$ (low signal to noise, strong factor), Table \ref{tab_sim_dgp2_inv_cov} d=1 (very low signal-to-noise ratio, strong factor), d=3 (low signal-to-noise ratio, strong factor), and Table \ref{tab_sim_dgp3_inv_cov}  (very low signal-to-noise ratio and weak factors). These setups correspondence to signal-to-noise ratio and strength of factors can be seen from Tables \ref{tab_sn_dgp1}-\ref{tab_sn_dgp3}. We characterize low signal to noise ratio as a number slightly above  one (1.03-1.21) and very low signal to noise ratio as number below one. Weaker factors are designated as $Eigmax \Sigma^f$ not increasing rapidly with $J$, hence only Table   \ref{tab_sim_dgp3_inv_cov}. To summarize SDNN-FM has the best performance overall.
 DNN-FM and SFM-POET and NL-LW  also has some good performance as well. To give an example at Table \ref{tab_sim_dgp3_inv_cov}, SDNN-FM with $d=5$, has the best record 6 out of 9 setups (n, J combinations) and NL-LW has 2 out of 9, and SFM-POET has 1 out of 9. DNN-FM does also well coming second in 6 out of 9 setups. 
   


\begin{table}[!t]
	\footnotesize
	\centering
	\caption{Simulation results - Third Monte Carlo design \\ Function estimation}
	\begin{threeparttable}
		\begin{tabular}{ccc|ccc|ccc|ccc}
			\hline
			\hline
			$d$     & $n$     & $J$     & SFM-POET & \textbf{DNN-FM} & \textbf{SDNN-FM} & $d$     & $n$     & $J$     & SFM-POET & \textbf{DNN-FM} & \textbf{SDNN-FM} \bigstrut\\
			\hline
			\multirow{9}[2]{*}{1} & 60    & 50    & 7.14  & 6.84  & 6.71  & \multirow{9}[2]{*}{5} & 60    & 50    & 8.28  & 7.45  & 7.12 \bigstrut[t]\\
			& 60    & 100   & 9.34  & 8.87  & 8.64  &       & 60    & 100   & 9.41  & 9.16  & 8.74 \\
			& 60    & 200   & 11.34 & 10.74 & 10.57 &       & 60    & 200   & 11.51 & 11.23 & 10.76 \\
			& 120   & 50    & 7.36  & 6.71  & 6.67  &       & 120   & 50    & 6.95  & 6.52  & 6.31 \\
			& 120   & 100   & 9.24  & 8.49  & 8.43  &       & 120   & 100   & 8.83  & 8.71  & 8.33 \\
			& 120   & 200   & 10.67 & 10.36 & 10.36 &       & 120   & 200   & 11.01 & 10.63 & 10.33 \\
			& 240   & 50    & 6.72  & 6.11  & 6.15  &       & 240   & 50    & 6.94  & 6.48  & 6.25 \\
			& 240   & 100   & 8.79  & 8.11  & 8.09  &       & 240   & 100   & 9.57  & 8.71  & 8.46 \\
			& 240   & 200   & 10.86 & 10.09 & 10.14 &       & 240   & 200   & 11.37 & 10.80 & 10.50 \bigstrut[b]\\
			\hline
			\multirow{9}[2]{*}{3} & 60    & 50    & 8.48  & 7.68  & 7.33  & \multirow{9}[2]{*}{7} & 60    & 50    & 8.00  & 7.64  & 7.33 \bigstrut[t]\\
			& 60    & 100   & 9.56  & 9.04  & 8.72  &       & 60    & 100   & 9.86  & 9.17  & 8.73 \\
			& 60    & 200   & 11.51 & 11.34 & 10.74 &       & 60    & 200   & 12.64 & 11.59 & 10.92 \\
			& 120   & 50    & 7.64  & 6.90  & 6.76  &       & 120   & 50    & 7.16  & 6.87  & 6.54 \\
			& 120   & 100   & 9.55  & 9.11  & 8.83  &       & 120   & 100   & 9.95  & 9.38  & 8.84 \\
			& 120   & 200   & 11.26 & 10.89 & 10.64 &       & 120   & 200   & 11.10 & 11.03 & 10.61 \\
			& 240   & 50    & 7.14  & 6.62  & 6.46  &       & 240   & 50    & 6.98  & 6.46  & 6.17 \\
			& 240   & 100   & 8.88  & 8.21  & 8.10  &       & 240   & 100   & 9.42  & 8.84  & 8.51 \\
			& 240   & 200   & 11.09 & 10.51 & 10.44 &       & 240   & 200   & 11.06 & 10.68 & 10.18 \bigstrut[b]\\
			\hline
			\hline
		\end{tabular}%
		\vspace*{-0.3cm}
		\begin{tablenotes}
			\footnotesize
			\singlespacing
			\item \leavevmode\kern-\scriptspace\kern-\labelsep 
			Note: The quantities in table relate to the error metric used in Theorem \ref{t1} and Theorem \ref{thm1} and correspond to the maximum difference between the estimated and true function.
		\end{tablenotes}
	\end{threeparttable}
	\label{tab_sim_dgp3_err}%
\end{table}%


\begin{table}[!t]
	\footnotesize
	\centering
	\caption{Simulation results - Third Monte Carlo design \\ Covariance matrix estimation}
	\begin{threeparttable}
		\begin{tabular}{ccc|cccccc}
			\hline
			\hline
			$d$     & $n$     & $J$     & SFM-POET & \textbf{DNN-FM} & \textbf{SDNN-FM} & NL-LW & SF-NL-LW & RN-FM \bigstrut\\
			\hline
			\multirow{9}[2]{*}{1} & 60    & 50    & 0.96  & 0.79  & 0.80  & 0.73  & 0.77  & 1.27 \bigstrut[t]\\
			& 60    & 100   & 1.04  & 0.91  & 0.91  & 0.95  & 0.88  & 1.06 \\
			& 60    & 200   & 1.15  & 1.03  & 1.04  & 1.27  & 1.11  & 1.07 \\
			& 120   & 50    & 0.93  & 0.61  & 0.65  & 0.55  & 0.55  & 0.83 \\
			& 120   & 100   & 1.03  & 0.70  & 0.74  & 0.77  & 0.64  & 0.89 \\
			& 120   & 200   & 1.12  & 0.81  & 0.84  & 1.09  & 0.91  & 0.90 \\
			& 240   & 50    & 0.93  & 0.49  & 0.56  & 0.45  & 0.45  & 0.57 \\
			& 240   & 100   & 1.04  & 0.56  & 0.64  & 0.59  & 0.50  & 0.74 \\
			& 240   & 200   & 1.12  & 0.62  & 0.71  & 0.89  & 0.72  & 0.81 \bigstrut[b]\\
			\hline
			\multirow{9}[2]{*}{3} & 60    & 50    & 0.91  & 0.77  & 0.77  & 0.71  & 0.73  & 1.31 \bigstrut[t]\\
			& 60    & 100   & 1.02  & 0.88  & 0.89  & 0.96  & 0.93  & 1.30 \\
			& 60    & 200   & 1.10  & 1.00  & 1.02  & 1.25  & 1.08  & 1.26 \\
			& 120   & 50    & 0.89  & 0.63  & 0.64  & 0.56  & 0.56  & 0.77 \\
			& 120   & 100   & 0.99  & 0.72  & 0.73  & 0.76  & 0.67  & 0.81 \\
			& 120   & 200   & 1.09  & 0.82  & 0.83  & 1.05  & 0.88  & 0.86 \\
			& 240   & 50    & 0.89  & 0.48  & 0.55  & 0.44  & 0.43  & 0.57 \\
			& 240   & 100   & 0.99  & 0.58  & 0.63  & 0.59  & 0.52  & 0.72 \\
			& 240   & 200   & 1.08  & 0.68  & 0.70  & 0.85  & 0.67  & 0.78 \bigstrut[b]\\
			\hline
			\multirow{9}[2]{*}{5} & 60    & 50    & 0.85  & 0.74  & 0.75  & 0.67  & 0.72  & 1.35 \bigstrut[t]\\
			& 60    & 100   & 0.96  & 0.85  & 0.86  & 0.91  & 0.91  & 1.57 \\
			& 60    & 200   & 1.07  & 0.96  & 0.99  & 1.21  & 1.06  & 1.64 \\
			& 120   & 50    & 0.82  & 0.60  & 0.62  & 0.53  & 0.51  & 0.75 \\
			& 120   & 100   & 0.93  & 0.71  & 0.71  & 0.76  & 0.65  & 0.80 \\
			& 120   & 200   & 1.05  & 0.81  & 0.81  & 1.04  & 0.89  & 0.85 \\
			& 240   & 50    & 0.83  & 0.49  & 0.53  & 0.41  & 0.40  & 0.55 \\
			& 240   & 100   & 0.94  & 0.60  & 0.62  & 0.58  & 0.52  & 0.68 \\
			& 240   & 200   & 1.06  & 0.68  & 0.69  & 0.84  & 0.66  & 0.75 \bigstrut[b]\\
			\hline
			\multirow{9}[2]{*}{7} & 60    & 50    & 0.82  & 0.74  & 0.75  & 0.69  & 0.74  & 1.46 \bigstrut[t]\\
			& 60    & 100   & 0.96  & 0.86  & 0.86  & 0.92  & 0.91  & 1.74 \\
			& 60    & 200   & 1.06  & 0.96  & 0.99  & 1.24  & 1.07  & 1.98 \\
			& 120   & 50    & 0.80  & 0.60  & 0.63  & 0.54  & 0.54  & 0.72 \\
			& 120   & 100   & 0.92  & 0.71  & 0.72  & 0.75  & 0.64  & 0.78 \\
			& 120   & 200   & 1.03  & 0.81  & 0.82  & 1.06  & 0.88  & 0.85 \\
			& 240   & 50    & 0.81  & 0.50  & 0.54  & 0.42  & 0.41  & 0.55 \\
			& 240   & 100   & 0.94  & 0.60  & 0.63  & 0.58  & 0.50  & 0.66 \\
			& 240   & 200   & 1.06  & 0.67  & 0.69  & 0.86  & 0.67  & 0.74 \bigstrut[b]\\
			\hline
			\hline
		\end{tabular}%
		\vspace*{-0.3cm}
		\begin{tablenotes}
			\footnotesize
			\singlespacing
			\item \leavevmode\kern-\scriptspace\kern-\labelsep 
			Note: The quantities in the table refer to the error metric used in Theorem \ref{covret}  and Theorem \ref{ds}\ref{ds_ii}, which corresponds to the maximum matrix norm of the difference between the estimated covariance matrix $\hat{\Sigma}_y$ and $\Sigma_y$, and $\hat{\Sigma}_y^s$ and $\Sigma_y$ respectively. The deep neural network factor model (DNN-FM) and sparse deep neural network factor model (SDNN-FM) are compared to the static factor model with observed factors and POET estimator from \cite{fan2013} (SFM-POET), the non-linear shrinkage estimator (NL-LW), the single factor non-linear shrinkage estimator (SF-NL-LW) of \cite{lw2017} and the residual nodewise regression estimation of factor models by \cite{caner2022}.
		\end{tablenotes}
	\end{threeparttable}
	\label{tab_sim_dgp3_cov}%
\end{table}%


\begin{table}[!t]
	\footnotesize
	\centering
	\caption{Simulation results - Third Monte Carlo design \\ Precision matrix estimation}
	\begin{threeparttable}
		\begin{tabular}{ccc|cccccc}
			\hline
			\hline
			$d$     & $n$     & $J$     & SFM-POET & \textbf{DNN-FM} & \textbf{SDNN-FM} & NL-LW & SF-NL-LW & RN-FM \bigstrut\\
			\hline
			\multirow{9}[2]{*}{1} & 60    & 50    & 1.04  & 1.02  & 1.01  & 4.62  & 6.37  & 86.19 \bigstrut[t]\\
			& 60    & 100   & 1.10  & 1.09  & 1.06  & 1.00  & 1.34  & 22.05 \\
			& 60    & 200   & 1.19  & 1.18  & 1.13  & 0.83  & 1.25  & 19.98 \\
			& 120   & 50    & 0.87  & 0.76  & 0.75  & 8.78  & 6.85  & 5.89 \\
			& 120   & 100   & 0.90  & 0.82  & 0.80  & 10.81 & 12.74 & 3.06 \\
			& 120   & 200   & 0.94  & 0.89  & 0.86  & 1.04  & 1.23  & 1.15 \\
			& 240   & 50    & 0.80  & 0.56  & 0.57  & 9.24  & 7.14  & 3.30 \\
			& 240   & 100   & 0.81  & 0.61  & 0.61  & 13.27 & 14.87 & 5.61 \\
			& 240   & 200   & 0.84  & 0.67  & 0.67  & 24.64 & 25.30 & 85.47 \bigstrut[b]\\
			\hline
			\multirow{9}[2]{*}{3} & 60    & 50    & 1.16  & 1.18  & 1.14  & 3.50  & 5.34  & 405.06 \bigstrut[t]\\
			& 60    & 100   & 1.22  & 1.26  & 1.14  & 1.00  & 1.35  & 278.29 \\
			& 60    & 200   & 1.30  & 1.34  & 1.16  & 0.84  & 1.27  & 60.62 \\
			& 120   & 50    & 0.93  & 0.86  & 0.83  & 5.34  & 6.18  & 6.28 \\
			& 120   & 100   & 0.97  & 0.93  & 0.87  & 4.63  & 10.12 & 4.89 \\
			& 120   & 200   & 1.00  & 0.97  & 0.87  & 1.03  & 1.23  & 1.25 \\
			& 240   & 50    & 0.83  & 0.63  & 0.62  & 5.39  & 4.89  & 2.79 \\
			& 240   & 100   & 0.85  & 0.71  & 0.67  & 11.13 & 11.38 & 2.05 \\
			& 240   & 200   & 0.87  & 0.76  & 0.69  & 11.11 & 20.95 & 1.24 \bigstrut[b]\\
			\hline
			\multirow{9}[2]{*}{5} & 60    & 50    & 1.22  & 1.31  & 1.24  & 3.16  & 3.84  & 395.95 \bigstrut[t]\\
			& 60    & 100   & 1.35  & 1.44  & 1.23  & 0.98  & 1.36  & 137.48 \\
			& 60    & 200   & 1.40  & 1.48  & 1.17  & 0.84  & 1.27  & 93.51 \\
			& 120   & 50    & 0.95  & 0.92  & 0.87  & 4.14  & 4.48  & 6.34 \\
			& 120   & 100   & 1.00  & 0.99  & 0.88  & 3.43  & 6.30  & 32.19 \\
			& 120   & 200   & 1.04  & 1.04  & 0.88  & 1.02  & 1.23  & 1.33 \\
			& 240   & 50    & 0.84  & 0.70  & 0.67  & 4.19  & 3.52  & 2.63 \\
			& 240   & 100   & 0.87  & 0.75  & 0.68  & 7.46  & 7.64  & 1.97 \\
			& 240   & 200   & 0.88  & 0.79  & 0.69  & 7.61  & 12.90 & 1.24 \bigstrut[b]\\
			\hline
			\multirow{9}[2]{*}{7} & 60    & 50    & 1.31  & 1.45  & 1.37  & 2.20  & 4.07  & 2113.83 \bigstrut[t]\\
			& 60    & 100   & 1.39  & 1.51  & 1.26  & 0.97  & 1.34  & 154.32 \\
			& 60    & 200   & 1.51  & 1.62  & 1.18  & 0.83  & 1.27  & 120.78 \\
			& 120   & 50    & 0.99  & 1.00  & 0.93  & 3.57  & 3.72  & 5.86 \\
			& 120   & 100   & 1.03  & 1.05  & 0.89  & 3.68  & 4.06  & 1.55 \\
			& 120   & 200   & 1.07  & 1.10  & 0.88  & 1.02  & 1.22  & 1.37 \\
			& 240   & 50    & 0.86  & 0.73  & 0.69  & 3.31  & 2.92  & 2.41 \\
			& 240   & 100   & 0.88  & 0.78  & 0.69  & 7.42  & 6.58  & 2.17 \\
			& 240   & 200   & 0.90  & 0.81  & 0.69  & 5.76  & 11.91 & 1.28 \bigstrut[b]\\
			\hline
			\hline
		\end{tabular}%
		\vspace*{-0.3cm}
		\begin{tablenotes}
			\footnotesize
			\singlespacing
			\item \leavevmode\kern-\scriptspace\kern-\labelsep 
			Note: The quantities in the table represent the error metric of Theorem \ref{thm4} and  Theorem \ref{ds}\ref{ds_iii}, which measures the spectral norm of the difference between the estimated and true precision matrix. The deep neural network factor model (DNN-FM) and sparse deep neural network factor model (SDNN-FM) are compared to the static factor model with observed factors and POET estimator from \cite{fan2013} (SFM-POET), the non-linear shrinkage estimator (NL-LW), the single factor non-linear shrinkage estimator (SF-NL-LW) of \cite{lw2017} and the residual nodewise regression estimation of factor models by \cite{caner2022}.
		\end{tablenotes}
	\end{threeparttable}
	\label{tab_sim_dgp3_inv_cov}%
\end{table}%


\FloatBarrier


\section{Empirical Application}\label{sec_pf}

In order to verify the practical validity of the DNN and SDNN factor models, we investigate their efficiency for estimating high-dimensional empirical portfolios.

In an out-of-sample portfolio forecasting experiment, we compare the performance of the global minimum variance portfolio (GMVP) strategy based on the covariance matrix estimated by our multilayer neural networks with fully and sparsely connected units as illustrated in Sections \ref{sec_model} and \ref{sec_factor_model}, respectively, with popular alternative portfolio strategies commonly used in the literature. As we are mainly interested in analyzing the empirical quality of the covariance matrix estimator, we concentrate on the estimation of GMVP weights, which are solely a function of the covariance matrix of the asset returns.

\subsection{Data and illustration of the forecasting experiment}

For our analysis we use the excess returns of stocks of the S\&P 500 index, which were constituents of the index on December 31, 2021. The excess returns are constructed by subtracting the corresponding one-month Treasury bill rate from the asset returns. 
We compare the forecasting results for asset returns on monthly frequency, where we use the prices available at the end of the corresponding month to calculate the returns. 

For our forecasting experiment, we consider two different time periods.
\begin{enumerate}
	\item Period 1: January 1986 - December 2019: The first sample period yields $n = 407$ monthly return observations and 227 assets are available for the entire sample.
	\item Period 2: January 1986 - December 2021: The second period yields $n = 431$ return observations for 227 assets. Compared to the first sample period, the second period incorporates the COVID-19 Crisis 2020/21 and allows us to analyze the effect of this turbulent episode on the forecasting performance of the considered methods.
\end{enumerate}

We conduct an out-of-sample forecasting experiment based on a rolling window approach with an insample size of ten years, which corresponds to $n_I = 120$ monthly observations. Specifically, at any investment date $h$, we use for the most recent 120 monthly returns for the estimation. Based on the estimated portfolio weights $\hat{\omega}_h$, we compute the out-of-sample portfolio return for period $h+1$ as $\hat{r}^{pf}_{h+1} = \hat{\omega}_h' r_{h+1}$. In the following step, we shift the insample window by one observation, estimate the portfolio weights $\hat{\omega}_{h+1}$ for investment period $h+1$ and compute the out-of-sample portfolio return $\hat{r}^{pf}_{h+1}$. This procedure is repeated until we reach the end of the sample. Hence, for the first period we obtain a series of $n - n_I = 287$ out-of-sample portfolio returns, whereas for the second sample period we retain 311 out-of-sample portfolio returns. All portfolios are updated on a monthly basis.

This result is used to calculate the average out-of-sample portfolio return and variance as follows
\begin{equation*}
	\hat{\mu} = \frac{1}{n - n_I} \sum_{h = n_I}^{n-1} \hat{r}^{pf}_{h+1} \quad \text{ and } \quad \hat{\sigma}^2 = \frac{1}{n - n_I - 1} \sum_{h = n_I}^{n-1} \left(\hat{r}^{pf}_{h+1} - \hat{\mu}\right)^2
\end{equation*}

In order to evaluate the performance of the approaches for different asset dimensions, we consider the following portfolio sizes: $J \in \{50, 100, 200\}$. Specifically, at the investment date $h$, we select the largest $J$ assets, as measured by their market value, which have a complete return history over the most recent $n_I$ periods, similar to \cite{lw2017}.

In addition, we investigate the portfolio performance in the presence of transaction costs. Following \cite{Li2015}, we calculate the out-of-sample excess returns with transaction costs according to
\begin{equation*}
	\hat{r}^{pf,tc}_{h+1} = \hat{\omega}_h' r_{h+1} - c \left(1 + \hat{\omega}_h' r_{h+1}\right) \sum_{j = 1}^{J} |\hat{\omega}_{h+1,j} - \hat{\omega}_{h,j}^+|,
\end{equation*}
where $\hat{\omega}_{h,j}^+ = \hat{\omega}_{h,j}(1+r_{h+1,j})/(1+r_{h+1}^{pf})$ is the portfolio weight of the $j$-th asset before rebalancing and $c$ are the proportional transaction costs, which we set to 50 basis points per transaction as adopted by \cite{demiguel2007optimal}.
Based on the previous definitions, we define the mean and variance of the excess portfolio returns with transaction costs as
\begin{equation*}
	\hat{\mu}^{tc} = \frac{1}{n - n_I} \sum_{h = n_I}^{n-1} \hat{r}^{pf, tc}_{h+1} \quad \text{ and } \quad \hat{\sigma}^{2, tc} = \frac{1}{n - n_I - 1} \sum_{h = n_I}^{n-1} \left(\hat{r}^{pf, tc}_{h+1} - \hat{\mu}^{tc}\right)^2.
\end{equation*}
Moreover, we determine the portfolio turnover by
\begin{equation*}
	\text{PT} =\frac{1}{n - n_I} \sum_{h = n_I}^{n-1}\sum_{j = 1}^{J} |\hat{\omega}_{h+1,j} - \hat{\omega}_{h,j}^+|.
\end{equation*}

To evaluate the portfolio performance of each method, we concentrate on the annualized out-of-sample standard deviation (SD), average return (AV) and Sharpe ratio (SR). Moreover, we analyze the average out-of-sample Portfolio Turnover (PT).

In the following, we provide an overview of the approaches, which we incorporate in the empirical study:
\begin{itemize}
	\item EW: The equally weighted portfolio.
	\item DNN-FM: Our non-linear nonparametric factor model estimated based on the deep neural network in Section \ref{sec_model}.
	\item SDNN-FM: Our non-linear nonparametric factor model estimated based on the s-sparse deep neural network in Section \ref{sec_factor_model}.
	
	In order to implement the DNN-FM and SDNN-FM approaches, we use the Fama-French three factors by \cite{Fama1993} as observed factors.
	
	\item POET: The principal orthogonal complement thresholding covariance matrix estimator from \cite{fan2013}, with latent factors and the number of factors is estimated based on the $IC_1$ criterion by \cite{Bai2002}.
	\item FF3F: The Fama-French three factor model introduced by \cite{Fama1993}.
	\item NL-LW: The non-linear shrinkage estimator of \cite{lw2017}.
	\item SF-NL-LW: The single factor non-linear shrinkage estimator of \cite{lw2017}.
	\item RN-FM: Residual based nodewise regression factor model of \cite{caner2022} incorporating the Fama-French three factors by \cite{Fama1993}.
\end{itemize}

\subsection{Out-of-sample portfolio results}
\begin{table}[!t]
	\footnotesize
	\centering
	\caption{Out-of-sample portfolio application results for the first sample}
	\begin{threeparttable}
		\begin{tabular}{c|cccccccc}
			\hline
			\hline
			\multicolumn{9}{l}{Period 1: January 1986 - December 2019} \bigstrut[t]\\
			\multicolumn{9}{l}{Without transaction costs} \bigstrut[b]\\
			\hline
			Model & EW    & \textbf{DNN-FM} & \textbf{SDNN-FM} & POET  & FF3F  & NL-LW & SF-NL-LW & RN-FM \bigstrut\\
			\hline
			\multicolumn{9}{c}{$J = 50$} \bigstrut\\
			\hline
			SD    & 0.143 & 0.136 & 0.136 & 0.143 & 0.142 & 0.146 & 0.144 & 0.139 \bigstrut[t]\\
			AV    & 0.067 & 0.077 & 0.075 & 0.067 & 0.068 & 0.082 & 0.083 & 0.072 \\
			SR    & 0.464 & 0.562 & 0.555 & 0.467 & 0.476 & 0.560 & 0.576 & 0.517 \\
			PT    & 0.071 & 0.199 & 0.175 & 0.075 & 0.083 & 0.374 & 0.375 & 0.241 \bigstrut[b]\\
			\hline
			\multicolumn{9}{c}{$J = 100$} \bigstrut\\
			\hline
			SD    & 0.144 & 0.133 & 0.133 & 0.145 & 0.140 & 0.141 & 0.139 & 0.137 \bigstrut[t]\\
			AV    & 0.069 & 0.078 & 0.077 & 0.064 & 0.072 & 0.061 & 0.059 & 0.074 \\
			SR    & 0.479 & 0.585 & 0.582 & 0.446 & 0.516 & 0.430 & 0.426 & 0.544 \\
			PT    & 0.073 & 0.284 & 0.243 & 0.121 & 0.098 & 0.516 & 0.535 & 0.220 \bigstrut[b]\\
			\hline
			\multicolumn{9}{c}{$J = 200$} \bigstrut\\
			\hline
			SD    & 0.145 & 0.137 & 0.138 & 0.142 & 0.140 & 0.134 & 0.132 & 0.139 \bigstrut[t]\\
			AV    & 0.086 & 0.113 & 0.117 & 0.082 & 0.089 & 0.089 & 0.090 & 0.094 \\
			SR    & 0.592 & 0.822 & 0.852 & 0.577 & 0.636 & 0.666 & 0.680 & 0.678 \\
			PT    & 0.066 & 0.461 & 0.437 & 0.235 & 0.107 & 0.483 & 0.470 & 0.180 \bigstrut[b]\\
			\hline
			\hline
			\multicolumn{1}{c}{} &       &       &       &       &       &       & 	& \bigstrut[t]\\
			\multicolumn{9}{l}{Period 1: January 1986 - December 2019} \\
			\multicolumn{9}{l}{With transaction costs} \bigstrut[b]\\
			\hline
			Model & EW    & \textbf{DNN-FM} & \textbf{SDNN-FM} & POET  & FF3F  & NL-LW & SF-NL-LW & RN-FM \bigstrut\\
			\hline
			\multicolumn{9}{c}{$J = 50$} \bigstrut\\
			\hline
			SD    & 0.143 & 0.136 & 0.136 & 0.143 & 0.142 & 0.146 & 0.144 & 0.139 \bigstrut[t]\\
			AV    & 0.062 & 0.065 & 0.065 & 0.062 & 0.063 & 0.059 & 0.060 & 0.058 \\
			SR    & 0.435 & 0.475 & 0.478 & 0.436 & 0.441 & 0.406 & 0.420 & 0.413 \bigstrut[b]\\
			\hline
			\multicolumn{9}{c}{$J = 100$} \bigstrut\\
			\hline
			SD    & 0.144 & 0.132 & 0.132 & 0.144 & 0.140 & 0.141 & 0.139 & 0.137 \bigstrut[t]\\
			AV    & 0.064 & 0.060 & 0.062 & 0.057 & 0.065 & 0.029 & 0.027 & 0.062 \\
			SR    & 0.449 & 0.457 & 0.472 & 0.396 & 0.467 & 0.210 & 0.194 & 0.452 \bigstrut[b]\\
			\hline
			\multicolumn{9}{c}{$J = 200$} \bigstrut\\
			\hline
			SD    & 0.145 & 0.136 & 0.137 & 0.142 & 0.140 & 0.133 & 0.132 & 0.139 \bigstrut[t]\\
			AV    & 0.082 & 0.085 & 0.091 & 0.068 & 0.082 & 0.060 & 0.061 & 0.083 \\
			SR    & 0.565 & 0.622 & 0.663 & 0.476 & 0.591 & 0.448 & 0.466 & 0.600 \bigstrut[b]\\
			\hline
			\hline
		\end{tabular}%
		\vspace*{-0.3cm}
		\begin{tablenotes}
			\footnotesize
			\singlespacing
			\item \leavevmode\kern-\scriptspace\kern-\labelsep 
			Note: Our deep neural network factor model (DNN-FM) and sparse deep neural network factor model (SDNN-FM) are compared to the equally weighted portfolio (EW), the POET estimator of \cite{fan2013} (POET), the three factor model of \cite{Fama1993} (FF3F), the non-linear shrinkage estimator (NL-LW) and the single factor non-linear shrinkage estimator (SF-NL-LW) of \cite{lw2017} and the residual based nodewise regression factor model of \cite{caner2022}.
		\end{tablenotes}
	\end{threeparttable}
	
	\label{tab_appl_p1}%
\end{table}%

The annualized out-of-sample portfolio results for the first sample period, which excludes the COVID-19 Crisis 2020/21 are illustrated in Table \ref{tab_appl_p1}. Compared to the competing methods, both DNN-FM and SDNN-FM provide the lowest out-of-sample standard deviations if the asset dimension $J$ is smaller than the in-sample size $n_I$ of 120 months without transaction costs. For $J = 200$ the non-linear shrinkage estimators of \cite{lw2017} lead to slightly lower portfolio standard deviations. Given the results of our theory and the simulation results, this outcome is anticipated, as our approaches offer the best performance when $n_I > J$. If we consider the out-of-sample SR without transaction costs, DNN-FM and SDNN-FM outperform the competing approaches for high portfolio dimensions, i.e.,\ $J \ge 100$. For example at $J=200$, our Sharpe ratios are 0.822, 0.852 for DNN-FM and SDNN-FM respectively, whereas the next highest Sharpe Ratio is with SF-NL-LW at 0.680 and RN-FM has Sharpe Ratio of 0.678. The differences are not minor between the methods.
For low asset dimensions ($J = 50$) our methods lead to a similar performance as NL-LW and SF-NL-LW in terms of SR. However, it is important to note that the non-linear shrinkage estimators of \cite{lw2017} generate the highest portfolio turnovers across the considered approaches. If we compare DNN-FM and SDNN-FM, we observe that both models lead to very similar portfolio performances, if transactions costs are not considered.  


\begin{table}[!t]
	\footnotesize
	\centering
	\caption{Out-of-sample portfolio application results for the second sample}
	\begin{threeparttable}
		\begin{tabular}{c|cccccccc}
			\hline
			\hline
			\multicolumn{9}{l}{Period 2: January 1986 - December 2021} \bigstrut[t]\\
			\multicolumn{9}{l}{Without transaction costs} \bigstrut[b]\\
			\hline
			Model & EW    & \textbf{DNN-FM} & \textbf{SDNN-FM} & POET  & FF3F  & NL-LW & SF-NL-LW & RN-FM \bigstrut\\
			\hline
			\multicolumn{9}{c}{$J = 50$} \bigstrut\\
			\hline
			SD    & 0.147 & 0.139 & 0.140 & 0.147 & 0.145 & 0.150 & 0.148 & 0.143 \bigstrut[t]\\
			AV    & 0.076 & 0.089 & 0.088 & 0.078 & 0.078 & 0.099 & 0.099 & 0.082 \\
			SR    & 0.520 & 0.637 & 0.632 & 0.528 & 0.534 & 0.659 & 0.665 & 0.573 \\
			PT    & 0.072 & 0.204 & 0.180 & 0.083 & 0.084 & 0.376 & 0.377 & 0.242 \bigstrut[b]\\
			\hline
			\multicolumn{9}{c}{$J = 100$} \bigstrut\\
			\hline
			SD    & 0.147 & 0.137 & 0.137 & 0.146 & 0.143 & 0.145 & 0.142 & 0.142 \bigstrut[t]\\
			AV    & 0.078 & 0.088 & 0.088 & 0.076 & 0.079 & 0.078 & 0.073 & 0.088 \\
			SR    & 0.528 & 0.644 & 0.638 & 0.517 & 0.553 & 0.536 & 0.517 & 0.620 \\
			PT    & 0.073 & 0.290 & 0.250 & 0.135 & 0.101 & 0.555 & 0.569 & 0.223 \bigstrut[b]\\
			\hline
			\multicolumn{9}{c}{$J = 200$} \bigstrut\\
			\hline
			SD    & 0.150 & 0.143 & 0.144 & 0.144 & 0.144 & 0.136 & 0.135 & 0.143 \bigstrut[t]\\
			AV    & 0.093 & 0.117 & 0.120 & 0.085 & 0.095 & 0.098 & 0.098 & 0.101 \\
			SR    & 0.619 & 0.816 & 0.836 & 0.589 & 0.659 & 0.720 & 0.724 & 0.706 \\
			PT    & 0.067 & 0.480 & 0.449 & 0.245 & 0.112 & 0.485 & 0.471 & 0.187 \bigstrut[b]\\
			\hline
			\hline
			\multicolumn{1}{r}{} &       &       &       &       &       &       &   & \bigstrut[t]\\
			\multicolumn{9}{l}{Period 2: January 1986 - December 2021} \\
			\multicolumn{9}{l}{With transaction costs} \bigstrut[b]\\
			\hline
			Model & EW    & \textbf{DNN-FM} & \textbf{SDNN-FM} & POET  & FF3F  & NL-LW & SF-NL-LW & RN-FM \bigstrut\\
			\hline
			\multicolumn{9}{c}{$J = 50$} \bigstrut\\
			\hline
			SD    & 0.147 & 0.139 & 0.139 & 0.147 & 0.145 & 0.150 & 0.148 & 0.143 \bigstrut[t]\\
			AV    & 0.072 & 0.076 & 0.077 & 0.073 & 0.073 & 0.076 & 0.076 & 0.067 \\
			SR    & 0.490 & 0.549 & 0.555 & 0.495 & 0.499 & 0.508 & 0.512 & 0.472 \bigstrut[b]\\
			\hline
			\multicolumn{9}{c}{$J = 100$} \bigstrut\\
			\hline
			SD    & 0.147 & 0.137 & 0.137 & 0.146 & 0.143 & 0.143 & 0.141 & 0.142 \bigstrut[t]\\
			AV    & 0.073 & 0.071 & 0.072 & 0.067 & 0.073 & 0.044 & 0.039 & 0.075 \\
			SR    & 0.498 & 0.516 & 0.529 & 0.462 & 0.511 & 0.306 & 0.275 & 0.528 \bigstrut[b]\\
			\hline
			\multicolumn{9}{c}{$J = 200$} \bigstrut\\
			\hline
			SD    & 0.150 & 0.142 & 0.143 & 0.144 & 0.143 & 0.136 & 0.135 & 0.143 \bigstrut[t]\\
			AV    & 0.089 & 0.088 & 0.093 & 0.070 & 0.088 & 0.069 & 0.069 & 0.090 \\
			SR    & 0.593 & 0.616 & 0.650 & 0.486 & 0.613 & 0.505 & 0.514 & 0.628 \bigstrut[b]\\
			\hline
			\hline
		\end{tabular}%
		\vspace*{-0.3cm}
		\begin{tablenotes}
			\footnotesize
			\singlespacing
			\item \leavevmode\kern-\scriptspace\kern-\labelsep 
			Note: Our deep neural network factor model (DNN-FM) and sparse deep neural network factor model (SDNN-FM) are compared to the equally weighted portfolio (EW), the POET estimator of \cite{fan2013} (POET), the three factor model of \cite{Fama1993} (FF3F), the non-linear shrinkage estimator (NL-LW) and the single factor non-linear shrinkage estimator (SF-NL-LW) of \cite{lw2017} and the residual based nodewise regression factor model of \cite{caner2022}.
		\end{tablenotes}
	\end{threeparttable}
	
	\label{tab_appl_p2}%
\end{table}%


When transaction cost are taken into account, the SDNN-FM offers the highest SR for all portfolio dimensions. Also the larger J, the larger the Sharpe-Ratio of SDNN-FM. At $J=50$, Sharpe Ratio of SDNN-FM is 0.478, but increases to 0.663 at $J=200$.
This is mainly due to high out-of-sample returns and a relative low portfolio turnover. Hence, the sparsity in the neural network weights of the SDNN-FM has a stabilizing effect on the corresponding portfolio weights. The general better performance of the SDNN-FM compared to FF3F and POET testifies the advantage of measuring non-linear relations in high-dimensional portfolios based on deep neural networks compared to models, which solely allow for capturing linear effects. Moreover, the empirical results confirm our theoretical findings in Section \ref{pmret}, which reinforce the effectiveness of DNN estimators settings characterized by a low signal-to-noise ratio, as commonly observed in financial data. In contrast to that, traditional approximate factor models rely on the pervasiveness assumption, which implies an environment with high signal. Hence, the superior performance of the DNN estimators compared to the POET approach can be attributed to their ability to effectively capture the nuances and complexities inherent in low signal scenarios, providing a more accurate and reliable estimation framework.

The annualized results for the second sample period, which incorporates the COVID-19 Crisis are reported in Table \ref{tab_appl_p2}. Overall the results are very similar to the ones obtained for the first sample period without COVID-19 Crisis. However, as anticipated, the recent turbulent period in 2020/21 leads to a general increase in out-of-sample portfolio volatility for all considered methods and portfolio dimensions. Nevertheless, the DNN-FM and SDNN-FM lead to the lowest SD for $n_I=120 > J$, which is even more pronounced compared to the first sample period. To see this fact, at $J=50$, DNN-FM, SDNN-FM have standard deviations of 0.139, and 0.140, respectively, whereas the next best method is RN-FM with 0.143 as the standard deviation without transaction costs.
This indicates that the both models can compensate the large volatility during the COVID-19 Crisis. Moreover, when transaction costs are taken into account the SDNN-FM leads to the highest SR for all portfolio dimensions. To give an example with $J=100$, SDNN-FM has the highest Sharpe Ratio with 0.650, compared with second best RN-FM at 0.628.



\section{Conclusions}\label{sec_conclusions}

In this paper, we contribute to the theoretical understanding of the advantages in predictive power of modern deep neural networks. Specifically, we analyze the large sample properties of feedforward multilayer neural networks (multilayer perceptron) in multivariate nonparametric regression models. Our theoretical elaboration generalizes and extends the deep learning results of \cite{sh2020} and \cite{Farrell2021}. 
We provide uniform results on the expected estimation risk that are novel to the deep learning literature. More precisely, we analyze the properties of a deep neural network that incorporates a set of multiple response variables. We derive an upper bound on the expected risk, show that these bounds are uniform over the large set of variables. The rates depend on the tradeoff between the smoothness of the underlying function, the number of regressors and the number of variables.

Moreover, we build a bridge between our deep learning framework and non-linear nonparametric factor models in finance applications. Our deep neural network factor model (DNN-FM) allows to measure flexible linear and non-linear interactions between the considered asset return series and underlying factors. Hence, compared to the traditional factor models that impose a rigid linear relationship between the factors and considered variables, our deep neural network factor model considerably enhances the model flexibility. In order to analyze the implications of imposing sparsity in the parametrization of the deep neural network, we extend our deep learning results to an non-linear additive model setting.

Furthermore, we develop a novel data-dependent estimator of the covariance matrix of the residuals in the DNN-FM and its sparse version (SDNN-FM), which is necessary to construct a robust estimator for the covariance and precision matrix of the asset returns based on the DNN-FM and SDNN-FM. The estimator is based on a flexible adaptive thresholding method, which is robust to outliers in the innovations. We prove that the adaptive thresholding estimator is consistent under the $l_2$-norm. In addition, we provide consistency results of the corresponding covariance and precision matrix estimators based on both deep neural network specification. It is important to note that we obtain these consistency results under a very low signal-to-noise ratio assumption, which implies that our model setting is better suited to measure the spectral structure of datasets incorporating assets returns compared to standard linear factor models relying on the pervasiveness assumption.

In the Monte Carlo study, we analyze the finite sample properties of our DNN-FM and SDNN-FM using various simulation designs. The results confirm our large sample findings. Specifically, both models consistently estimate the true underlying non-linear functional form, which connects the observed factors with the considered time series, as the number of temporal periods increases. Moreover, the corresponding covariance and precision matrix estimators based on the DNN-FM and SDNN-FM are consistent as well. The simulation results further verify that the convergence rates of both estimators are independent of the number of incorporated factors. In contrast to that, competing methods are more sensitive to the design of the data generating process and their error rates are negatively affected if the number of included factors increases.

In an out-of-sample portfolio application, we investigate the efficiency of the DNN-FM and SDNN-FM in predicting high-dimensional empirical portfolios in a global minimum variance portfolio setting and compare the performance to alternative approaches that are commonly used in the literature. The forecasting results show that our DNN-FM and SDNN-FM lead to the lowest out-of-sample portfolio standard deviation when the number of periods is larger than the number of assets. At the same time, they avoid large changes in the portfolio constellation, which leads to a low portfolio turnover. This results in the highest out-of-sample Sharpe ratio across different portfolio sizes compared to all alternative estimators when transaction costs are taken into account. Moreover, the results illustrate that the advantage of both deep neural networks in terms of lowest portfolio standard deviation is especially pronounced during volatile periods, such as during the COVID-19 Crisis.

\if0\blind
{
\section*{Acknowledgments}
\addcontentsline{toc}{section}{Acknowledgments}
For helpful comments on an earlier draft of the paper we would like to thank Yuan Liao, Agostino Capponi, and Marcelo Medeiros. We also thank the participants of the 33rd (EC)$^2$ conference in Paris-2022, the 16th International Conference on Computational and Financial Econometrics-2022 in London, the Financial Econometrics Conference in Lancaster-2023, the 2023 North American Summer Meeting of the Econometric Society in Los Angeles, the 2023 Rochester Conference in
Factor Model Based Econometrics, the 2022 NBER-NSF Time Series Conference, as well as seminar participants at the ETH Zurich, KOF Swiss Economic Institute, the University of Exeter and the University of Konstanz, and University of Illinois, Urbana-Champaign.
} \fi

\setcounter{equation}{0}
\setcounter{lemma}{0}
\setcounter{table}{0}
\renewcommand{\theequation}{A.\arabic{equation}}
\renewcommand{\thelemma}{A.\arabic{lemma}}
\renewcommand{\thecorollary}{A.\arabic{corollary}}
\renewcommand{\thetable}{\thesection.\arabic{table}}

\appendix
\section*{Appendix}
\addcontentsline{toc}{section}{Appendix}
\section{Proofs}\label{sec_A_proofs}

{\bf Proof of Theorem \ref{t1}}.(i).
We start with Theorem \ref{t1}(b) and its proof on p.206  in \cite{Farrell2021}, with $\gamma=n^{d/(\beta+d)} (\log n)^8$,
\[ P \left\{ \frac{1}{n} \sum_{i=1}^n [ \hat{f}_j (X_i) - f_{0,j} (X_i)]^2 \ge C_1  [n^{-\beta/(\beta+d)} (\log n)^8 + \frac{\log(\log n)+\gamma}{n} ]
\right\} \le \exp (-\gamma) = 
\exp (- n^{d/(\beta+d)} (\log n)^8).\]
 and then take the union bound and take into account the number of outcome variables in the rate of convergence, set 
 $\gamma_1 =\gamma+ \log J$

The above inequality simplifies when we substitute $\gamma_1$ definition

\[
  P \{ \max_{1 \le j \le J} \frac{1}{n} \sum_{i=1}^n [ \hat{f}_j (X_i) - f_{0,j} (X_i)]^2 \ge C_1  [n^{-\beta/(\beta+d)} (\log n)^8 + \frac{\log(\log n)}{n} + \frac{\gamma_1}{n}]
\} \le J \exp (- \gamma_1) ,\]
which implies 
\begin{eqnarray*}
 P \{ \max_{1 \le j \le J} \frac{1}{n} \sum_{i=1}^n [ \hat{f}_j (X_i) - f_{0,j} (X_i)]^2 &\ge& C_1  [2 n^{-\beta/(\beta+d)} (\log n)^8 + \frac{\log(\log n)}{n} + \frac{\log J }{n}]\}
\\
 &\le & \exp (\log J - n^{d/(\beta+d)}( \log n)^8 - \log J) \\
 & =&
\exp (- n^{d/(\beta+d)} (\log n)^8).
\end{eqnarray*}
Define $r_{n1}:= 2 C_1 [n^{-\beta/(\beta+d)} (\log n)^8 + \frac{\log(\log n)}{n}]$, and $r_{n2}:= C_1 [\frac{\log J}{n}]$.

(ii). The asymptotic case follows from the tail probability in (i).

{\hfill \bf Q.E.D.}\\

{\bf Proof of Lemma \ref{mtl1}}. 

\ref{mtl1_i} Note that by Assumption \ref{as1}, $u_{j,i}$ are bounded, so we can assume for a positive constant $K>0$, $\max_{1 \le i \le n } \max_{1 \le j \le J} |u_{j,i} | \le K^{1/2}$,
 and independent across $i=1,\cdots, n$,  for each $j=1,\cdots, J$. So $u_{j,i}$ are independent sub Gaussian for each $j=1,\cdots, J$, across $i=1,\cdots, n$, and multiple of sub Gaussian random variable are sub exponential $u_{j,i} u_{k,i}$, and centered version is also subexponential by p.31-32 of \cite{v2019}, then  we can benefit from Corollary 2.8.3 of \cite{v2019}, and providing a  union bound, for $c>0$ a positive constant

\[ P \left(  \max_{1 \le j \le J} \max_{1 \le k \le J} | \frac{1}{n} \sum_{i=1}^n u_{j,i} u_{k,i} - E u_{j,i} u_{k,i} | \ge t 
\right) \le J^2 2 \exp \left( -c \min (\frac{t^2}{K^2}, \frac{t}{K}) n \right).\]

\noindent  Then the right side upper bound probability simplifies
\[ J^2 2 \exp \left( -c \min (\frac{t^2}{K^2}, \frac{t}{K}) n \right) = \exp \left(\log 2J^2 - c \min (\frac{t^2}{K^2}, \frac{t}{K}) n \right).\]
Set $t=C_2 \sqrt{\log J/n}$, with sufficiently large $n$, ($n^{1/2} \ge (C_2/K) (\log J)^{1/2}$) we have  $t^2/K^2 \le t/K$, we only analyze 
\[  J^2 2 \exp \left( -c \min (\frac{t^2}{K^2}, \frac{t}{K}) n \right) = \exp \left(\log 2J^2 - c \frac{t^2 n}{K^2}\right).\]
Set $c_1:= c C_2^2/K^2 - 2 >0$ to have 
\[ \log 2 + 2 \log J - c \frac{C_2^2}{K^2} \log J  = \log (\frac{2}{J^{c_1}}).\]
This provides the desired result.

{\hfill \bf Q.E.D.}\\

\ref{mtl1_ii}  First, $u_{j,i} u_{k,i}  - E u_{j,i} u_{k,i}$ is subexponential random variables  across $i=1,\cdots, n$ by Lemma 2.7.7 and p. 32 of \cite{v2019}. Then 
$ |u_{j,i} u_{k,i}  - E u_{j,i} u_{k,i}| $ is subexponential by Proposition  2.7.1 of \cite{v2019}, and by p.32 of \cite{v2019}
$|u_{j,i} u_{k,i}  - E u_{j,i} u_{k,i}| - E |u_{j,i} u_{k,i}  - E u_{j,i} u_{k,i}|$ is subexponential. Then apply Lemma \ref{mtl1}\ref{mtl1_i} proof above. 

{\hfill \bf Q.E.D.}\\
  
{\bf Proof of Lemma \ref{nl3}}. \ref{l3_i} Clearly, $u_{j,i} - \hat{u}_{j,i}= \hat{f}_j (X_i) - f_{0,j} (X_i)$. Then
 we simplify the problem as 
 \begin{equation}
 \max_{1 \le j \le J} \frac{1}{n} \sum_{i=1}^n (u_{j,i} - \hat{u}_{j,i})^2 = \max_{1 \le j \le J} \frac{1}{n} \sum_{i=1}^n (\hat{f}_j (X_i) - f_{0,j} (X_i))^2.\label{pl3-1_1}
  \end{equation} 
 Define the rate of convergences specifically, with $C_{1}$ as a positive constant
 \[ r_{n1}:= C_{1} [ n^{-\beta/(\beta+d)} (\log n)^8 + \frac{\log(\log n)}{n}],\]
 and 
 \[ r_{n2}:= C_{1} [ \frac{\log J}{n}].\]
 We first show a finite sample result.  By Theorem \ref{t1}(i)
 \begin{equation}
 P \left(  \max_{1 \le j \le J} \frac{1}{n} \sum_{i=1}^n (u_{j,i} - \hat{u}_{j,i})^2 >  r_{n1} + r_{n2}
 \right) \le \exp (- n^{\frac{d}{\beta+d}} (\log n)^8). \label{nl3-1}
 \end{equation}
 
 By (\ref{nl3-1}), with $n \to \infty$, we get the desired result as tail probability goes to zero in both cases, and by $l_n:= \max(r_{n1}, r_{n2})$.
 
{\hfill \bf Q.E.D.}\\
  
\ref{l3_ii} Use triangle inequality by seeing $\hat{u}_{j,i}= \hat{u}_{j,i}- u_{j,i} + u_{j,i}$, and in the same way for $\hat{u}_{k,i}$, and Cauchy-Schwartz inequality for the second inequality below

\begin{eqnarray*}
\max_{1 \le j \le J } \max_{1 \le k \le J} | \frac{1}{n} \sum_{i=1}^n (\hat{u}_{j,i} \hat{u}_{k,i} - u_{j,i} u_{k,i}) | & \le & 
\max_{1 \le j \le J } \max_{1 \le k \le J} | \frac{1}{n} \sum_{i=1}^n (\hat{u}_{j,i}-u_{j,i})( \hat{u}_{k,i} -u_{k,i}) | \\
& + & 2 \max_{1 \le j \le J } \max_{1 \le k \le J} | \frac{1}{n} \sum_{i=1}^n u_{j,i} (\hat{u}_{k,i} -  u_{k,i}) | \\
& \le & \max_{1 \le j \le J} \frac{1}{n} \sum_{i=1}^n (\hat{u}_{j,i}- u_{j,i})^2\\
& +& 2 \sqrt{ \max_{1 \le j \le J} \frac{1}{n} \sum_{i=1}^n u_{j,i}^2} \sqrt{ \max_{1 \le j \le J}
\frac{1}{n} \sum_{i=1}^n (\hat{u}_{j,i} - u_{j,i})^2}.
\end{eqnarray*}
Then apply Lemma \ref{mtl1}\ref{mtl1_i}, Lemma \ref{nl3}(i), Assumption 1, and  we have  $\max_{1 \le j \le J} E  u_{j,i}^2 \le C <\infty$, and by (\ref{rln})
\begin{eqnarray}
P ( \max_{1 \le j \le J } \max_{1 \le k \le J} | \frac{1}{n} \sum_{i=1}^n (\hat{u}_{j,i} \hat{u}_{k,i} - u_{j,i} u_{k,i})| & > & (2 C)^{1/2}  l_{n}^{1/2} + l_n
) \nonumber \\
& \le & \frac{2}{J^{c_1}}+ \exp ( - n^{d/(\beta+d)} (\log n)^8).\label{nl3-3}
\end{eqnarray}

Note that in  (\ref{nl3-3}) the tail probability goes to zero with $n \to \infty$ and we have the desired result, with Assumption \ref{asa3}, $l_n \to 0$ as $n \to \infty$.

{\hfill \bf Q.E.D.}\\

\ref{l3_iii} By triangle inequality
\begin{eqnarray*}
 \max_{1 \le j \le J} \max_{1 \le k \le J} | \frac{1}{n} \sum_{i=1}^n \hat{u}_{j,i} \hat{u}_{k,i} - E u_{j,i} u_{k,i}| 
&\le & \max_{1 \le j \le J} \max_{1 \le k \le J} | \frac{1}{n} \sum_{i=1}^n( \hat{u}_{j,i} \hat{u}_{k,i} -  u_{j,i} u_{k,i}) | \\
&+&
 \max_{1 \le j \le J} \max_{1 \le k \le J} | \frac{1}{n} \sum_{i=1}^n( u_{j,i} u_{k,i} - E u_{j,i} u_{k,i})|.
 \end{eqnarray*}
Then by Lemma \ref{mtl1}(i), Lemma \ref{nl3}\ref{l3_ii} in  (\ref{nl3-3})

\begin{eqnarray*}
P (  \max_{1 \le j \le J} \max_{1 \le k \le J} | \frac{1}{n} \sum_{i=1}^n \hat{u}_{j,i} \hat{u}_{k,i} - E u_{j,i} u_{k,i}| & > & 
(2 C)^{1/2} l_n^{1/2} + l_n + C_2 \frac{\sqrt{\log J}}{\sqrt{n}} ) \\
& \le & \frac{4}{J^{c_1}} + \exp ( - n^{d/(\beta+d)} (\log n)^8).
\end{eqnarray*}

When $n \to \infty$  the tail probability converges to zero and we obtain the desired result via Assumption \ref{asa3}, and via $l_n$ definition in (\ref{rln}) and since the slowest rate is $l_n^{1/2}$.

 
 {\hfill \bf Q.E.D.}\\
  
 The proof  of Theorem \ref{thm3} uses the following lemma and this will be used to in robust adaptive thresholding covariance matrix estimator.

 \begin{lemma}\label{la4}
 Under Assumptions \ref{as1}-\ref{asa4}
\[ P \left(C_L \le \min_{1 \le  j \le J, 1 \le k \le J} \hat{\theta}_{j,k} \le \max_{1 \le j \le J, 1 \le k \le J} \hat{\theta} \le C_u \right)
\ge 1 - o(1),\] 

with $C_U:=\max_{1 \le j \le J, 1 \le k \le J} | \sigma_{j,k}| +
   \max_{1 \le j \le J} \sigma_{j,j} < \infty$, and $C_L:=\min_{1 \le j \le J, 1 \le k \le J} E | u_{j,i} u_{k,i} - \sigma_{j,k}|=c>0.$
\end{lemma}

 {\bf Proof of Lemma \ref{la4}}.

 First define  a positive constant $C_* \ge (2 C)^{1/2} + 1 + C_2> 0$  and  define the following events, that we condition our proof, 
   \[ {\cal A}_1:= \{ \max_{1 \le k \le J} \max_{1 \le j \le J} | \hat{\sigma}_{j,k} - \sigma_{j,k}| \le C_* l_{n}^{1/2} \},\] 
 \[ {\cal A}_2:= \{ \max_{1 \le j \le J} \frac{1}{n} \sum_{i=1}^n (\hat{u}_{j,i} - u_{j,i})^2 \le l_n \}.\]
 \[ {\cal A}_3:= \{ \max_{1 \le j \le J} | \frac{1}{n} \sum_{i=1}^n u_{j,i}^2 - \sigma_{j,j} | \le C_{2} \sqrt{\log J/n} \}.\]
 \[ {\cal A}_4:= \{ \max_{ 1 \le j \le J} \max_{1 \le k \le J} \left|\frac{1}{n} \sum_{i=1}^n |u_{j,i} u_{k,i} - \sigma_{j,k}| -
 E |u_{j,i} u_{k,i} - \sigma_{j,k}|  \right| \le C_{2} \sqrt{\frac{\log J}{n}}\}.\]
 and then we will relax the condition and get the probabilistic result. 
 We start with the upper bound. By using the triangle inequality
 \begin{eqnarray}
 \max_{1 \le j \le J} \max_{1 \le k \le J} \hat{\theta}_{j,k} & = & \max_{1 \le j \le J} \max_{1 \le k \le J}  \frac{1}{n} \sum_{i=1}^n | \hat{u}_{j,i} \hat{u}_{k,i} - \hat{\sigma}_{j,k}| \le 
\max_{1 \le j \le J} \max_{1 \le k \le J}  \frac{1}{n} \sum_{i=1}^n | \hat{u}_{j,i} \hat{u}_{k,i} - \sigma_{j,k}| \nonumber \\
&+ &
\max_{1 \le j \le J} \max_{1 \le k \le J} \frac{1}{n}  \sum_{i=1}^n | \sigma_{j,k} - \hat{\sigma}_{j,k}| \nonumber \\
 & = & 
\max_{1 \le j \le J} \max_{1 \le k \le J} \frac{1}{n} \sum_{i=1}^n | \hat{u}_{j,i} \hat{u}_{k,i} - \sigma_{j,k}| + \max_{1 \le j \le J ,1 \le k \le J} | \sigma_{j,k} - \hat{\sigma}_{j,k}|.\label{la4-1}
\end{eqnarray}
Then condition on the event ${\cal A}_1$, and use it in second term on the right side of (\ref{la4-1})
 \begin{equation}
  \max_{1 \le j \le J ,1 \le k \le J} | \sigma_{j,k} - \hat{\sigma}_{j,k}|  \le C_* l_{n}^{1/2} = o(1),\label{la4-2}
 \end{equation}
   where the asymptotic negligibility is  by $l_{n}=o(1)$ by Assumption \ref{asa3} and definition in (\ref{rln}). Consider the first term on the right side of (\ref{la4-1}).
   \begin{eqnarray}
   \max_{1 \le j \le J} \max_{1 \le k \le J} \frac{1}{n} \sum_{i=1}^n | \hat{u}_{j,i} \hat{u}_{k,i} - \sigma_{j,k} | & \le & 
  \max_{1 \le j \le J} \max_{1 \le k \le J}  \frac{1}{n} \sum_{i=1}^n | (\hat{u}_{j,i} - u_{j,i}) (\hat{u}_{k,i} - u_{k,i})| \nonumber \\
   & + & \max_{1 \le j \le J} \max_{1 \le k \le J} \frac{1}{n} \sum_{i=1}^n | (\hat{u}_{j,i} - u_{j,i}) ( u_{k,i}) | +
   \max_{1 \le j \le J} \max_{1 \le k \le J}  \sum_{i=1}^n | (u_{j,i}) (\hat{u}_{k,i} - u_{k,i}) | \nonumber \\ 
 &  +& \max_{1 \le j \le J} \max_{1 \le k \le J} \frac{1}{n} \sum_{i=1}^n | u_{j,i} u_{k,i} - \sigma_{j,k}|
   .\label{la4-3}
      \end{eqnarray}
      
   Consider the first term on the right side of (\ref{la4-3}) via Cauchy-Schwartz inequality
      \begin{equation}
  \max_{1 \le j \le J} \max_{1 \le k \le J}    \sum_{i=1}^n | (\hat{u}_{j,i} - u_{j,i}) (\hat{u}_{k,i} - u_{k,i})|     \le \sqrt{\max_{1 \le j \le J}  \frac{1}{n} \sum_{i=1}^n (\hat{u}_{j,i} - u_{j,i})^2} \sqrt{\max_{1 \le k \le J} \frac{1}{n} \sum_{i=1}^n (\hat{u}_{k,i} - u_{k,i})^2}   \le l_n,\label{la4-4}   
      \end{equation}
      by event ${\cal A}_2$. Consider the second term on the right side of (\ref{la4-3})
     \begin{eqnarray}
\max_{1 \le j \le J} \max_{1 \le k \le J} \frac{1}{n} \sum_{i=1}^n | (\hat{u}_{j,i} - u_{j,i}) ( u_{k,i}) |  &\le & \sqrt{\max_{1 \le j \le J}  \frac{1}{n} \sum_{i=1}^n (\hat{u}_{j,i} - u_{j,i})^2} \sqrt{\max_{1 \le k \le J} \frac{1}{n} \sum_{i=1}^n u_{k,i}^2} \nonumber \\
&\le&  l_{n}^{1/2} \left\{ \max_{ 1 \le k \le J} \sigma_{k,k}  + C_{2} \sqrt{\frac{\log J}{n}}\right\}^{1/2},\label{la4-5}
\end{eqnarray}
where we use Cauchy-Schwartz for the first inequality, and we use events ${\cal A}_2 \cap {\cal A}_3$. Same analysis applies to third term on the right side of (\ref{la4-3}). Consider the fourth term on the right side of (\ref{la4-3})

\begin{equation}
\max_{1 \le j \le J} \max_{1 \le k \le J} \frac{1}{n} \sum_{i=1}^n | u_{j,i} u_{k,i} - \sigma_{j,k}|  \le 
\max_{1 \le j \le J} \max_{1 \le k \le J} \frac{1}{n} \sum_{i=1}^n |u_{j,i} u_{k,i}| + \max_{1 \le j \le J, 1 \le k \le J} |\sigma_{j,k}|.\label{la4-7}    
  \end{equation}    
   Use Cauchy-Schwartz inequality
  \begin{eqnarray}
\max_{1 \le j \le J} \max_{1 \le k \le J} \frac{1}{n} \sum_{i=1}^n |u_{j,i} u_{k,i}| & \le & \sqrt{
\max_{1 \le j \le J} \frac{1}{n} \sum_{i=1}^n u_{j,i}^2} \sqrt{ \max_{1 \le k \le J}\frac{1}{n} \sum_{i=1}^n u_{k,i}^2} 
\nonumber \\
& \le & [ \max_{1 \le j \le J} \sigma_{j,j} + C_{2} \sqrt{\frac{\log J}{n}}]^{1/2} [ \max_{1 \le k\le J} \sigma_{k,k} + C_2 \sqrt{\frac{\log J}{n}}]^{1/2}\label{la4-8}
\end{eqnarray}
via event ${\cal A}_3$. Combine (\ref{la4-4})-(\ref{la4-8}) in (\ref{la4-3}) on ${\cal A}_1 \cap {\cal A}_2 \cap {\cal A}_3$
\begin{eqnarray}
  \max_{1 \le j \le J, 1 \le k \le J}  \frac{1}{n} \sum_{i=1}^n | \hat{u}_{j,i} \hat{u}_{k,i} - \sigma_{j,k} | & \le &   
   l_n  + 2 l_{n}^{1/2} [ \max_{1 \le j \le J} \sigma_{j,j} + C_{2} \sqrt{\frac{\log J}{n}}]^{1/2} + \max_{1 \le j \le J , 1 \le k \le J} |\sigma_{j,k}| \nonumber \\
   & +& 
   [ \max_{1 \le j \le J} \sigma_{j,j} + C_{2} \sqrt{\frac{\log J}{n}}]^{1/2} [ \max_{1 \le k\le J} \sigma_{k,k} + C_{2} \sqrt{\frac{\log J}{n}}]^{1/2}.\label{la4-9}
   \end{eqnarray} 
   Then add (\ref{la4-2}) to (\ref{la4-9}), and   
   since $l_{n}=o(1)$ by Assumption \ref{asa3} with (\ref{rln})
   \begin{equation}
   \max_{1 \le j \le J, 1 \le k \le J} \hat{\theta}_{j,k} \le \max_{1 \le j \le J, 1 \le k \le J} | \sigma_{j,k}| +
   \max_{1 \le j \le J} \sigma_{j,j} + o(1).\label{la4-10}
   \end{equation}
   
   Note that ${\cal A}_1$ holds with probability approaching one, due to the following argument. 
   By Lemma \ref{mtl1}, (\ref{nl3-3}) and 
   \begin{eqnarray*}
   \max_{1 \le j \le J} \max_{1 \le k \le J} | \hat{\sigma}_{j,k} - \sigma_{j,k}| & \le & 
   \max_{1 \le j \le J } \max_{1 \le k \le J} | \frac{1}{n} \sum_{i=1}^n (u_{j,i} u_{k,i} - \sigma_{j,k})| \\
   & + & \max_{1 
   \le j \le J } \max_{1 \le k \le J} | \frac{1}{n} \sum_{i=1}^n (\hat{u}_{j,i} \hat{u}_{k,i} - u_{j,i} u_{k,i})| \\
   & \le &  (2 C)^{1/2} l_n^{1/2} + l_n + C_2 \sqrt{\frac{\log J}{n}} \\
   & \le & C_* l_n^{1/2},
   \end{eqnarray*}
   where the last inequality is by $C_*$, $l_n:=\max(r_{n1}, r_{n2})$ definitions in Theorem \ref{t1}.
   
   See that by defining $C_U:= \max_{1 \le j \le J, 1 \le k \le J} | \sigma_{j,k}| +
   \max_{1 \le j \le J} \sigma_{j,j} < \infty$ and using Lemma \ref{mtl1}-\ref{nl3}
   we have 
   \[ P ({\cal A}_1 \cap {\cal A}_2 \cap   {\cal A}_3) \ge 1 -  o(1).\]

   We analyze the lower bound in our problem.    We condition on the events ${\cal A}_1, {\cal A}_2, {\cal A}_3, {\cal A}_4$. Then we relax this assumption at the end of the proof here.
   We start with the following triangle inequality and use $\hat{\theta}_{j,k}$ definition
   \begin{eqnarray}
 \frac{1}{n} \sum_{i=1}^n | u_{j,i} u_{k,i} - \sigma_{j,k}| & \le & \frac{1}{n} \sum_{i=1}^n | u_{j,i} u_{k,i} - \hat{u}_{j,i} \hat{u}_{k,i}| 
 + \frac{1}{n} \sum_{i=1}^n | \hat{u}_{j,i} \hat{u}_{k,i} - \hat{\sigma}_{j,k}| + \frac{1}{n} \sum_{i=1}^n | \hat{\sigma}_{j,k} - \sigma_{j,k}|
 \nonumber \\
 & \le & \frac{1}{n} \sum_{i=1}^n | u_{j,i} u_{k,i} - \hat{u}_{j,i} \hat{u}_{k,i}| 
 + \hat{\theta}_{j,k}+ \max_{1 \le j \le J, 1 \le k \le J} | \hat{\sigma}_{j,k} - \sigma_{j,k}|.\label{la4-11}
  \end{eqnarray}

  The term on the left side of (\ref{la4-11}) is lower bounded by 
  \begin{eqnarray}
    \frac{1}{n} \sum_{i=1}^n | u_{j,i} u_{k,i} - \sigma_{j,k}|   &\ge& E | u_{j,i} u_{k,i} - \sigma_{j,k}| - C_{2} \sqrt{\frac{\log J}{n}} \nonumber \\
    & \ge & c - C_{2} \sqrt{\frac{\log J}{n}},\label{la4-12}
      \end{eqnarray}
   where the first inequality is by event ${\cal A}_4$, and the second inequality is by Assumption \ref{asa4}. Then analyze the right side of (\ref{la4-11}) by  event ${\cal A}_1$
   \[ \max_{1 \le j \le J, 1 \le k \le J} | \hat{\sigma}_{j,k} - \sigma_{j,k}| \le C_* l_{n}^{1/2}.\]

   Consider the first term on the right side of (\ref{la4-11}) by triangle inequality and (\ref{la4-4})-(\ref{la4-5}) on ${\cal A}_2 \cap {\cal A}_3$
  \begin{eqnarray}
  \max_{j,k}  \frac{1}{n} \sum_{i=1}^n | u_{j,i} u_{k,i} - \hat{u}_{j,i} \hat{u}_{k,i}|  & \le & \max_{j,k} \frac{1}{n} \sum_{i=1}^n |( \hat{u}_{j,i} - u_{j,i})(\hat{u}_{k,i} - u_{k,i})| + 
  \max_{j,k} \frac{1}{n} \sum_{i=1}^n    | ( \hat{u}_{j,i} - u_{j,i})(u_{k,i})|    \nonumber \\
&   + & \max_{j,k} \frac{1}{n} \sum_{i=1}^n  | ( u_{j,i})(\hat{u}_{k,i} - u_{k,i})| \nonumber \\
   & \le & l_n  + 2  l_{n}^{1/2}  \{ \max_{1 \le j \le J} \sigma_{j,j} + C_{2} \sqrt{\log J/n}   \}^{1/2} .\label{la4-13}
    \end{eqnarray}  
    Combine (\ref{la4-12})(\ref{la4-13}) in (\ref{la4-11}), 
    \[
    \min_{1 \le j \le J} \min_{1 \le k \le J} \hat{\theta}_{j,k} \ge     c - o(1),
            \]
     by Assumption \ref{asa3}, since $l_{n}=o(1)$. Since we condition on $\cap_{l=1}^4 {\cal A}_l$
    \[ P (\cap_{l=1}^4 {\cal A}_l ) \ge 1 - o(1),\]
    by Lemmata \ref{mtl1}-\ref{nl3}. 
    
    {\hfill \bf Q.E.D.}\\
    
    {\bf Proof of Theorem} {\ref{thm3}.

     \ref{thm3_i} Define $b_n:= C_* l_{n}^{1/2}$. Define   a new positive  constant $C_{**}>0$ such that $C_{**} C_L > 2 C_{*}>0$,  we get 
     \begin{equation}
     C_L \omega_n > 2 b_n,\label{pt3-1}
     \end{equation}
     by (\ref{rwn}).

  Next define
  \[ {\cal B}_1:= \{ \max_{1 \le k \le J} \max_{1 \le j \le J} | \hat{\sigma}_{j,k} - \sigma_{j,k}| \le C_{*}  l_{n}^{1/2}\},\]
  \[ {\cal B}_2:= \{ \min_{1 \le k \le J} \min_{1 \le j \le J} \hat{\theta}_{j,k} \omega_n > 2 b_n \} ,\]
  \[ {\cal B}_3:= \{ \max_{1 \le k \le J} \max_{1 \le j \le J} \hat{\theta}_{j,k} \le C_U \}.\]
  
  Define also the event $E:= \{ \cap_{l=1}^3 {\cal B}_l \}$. Note that under $E$, with $b_n$ definition
  \begin{equation}
  | \hat{\sigma}_{j,k}| \ge \omega_n \hat{\theta}_{j,k} \quad {\mbox implies} \quad |\sigma_{j,k}| \ge b_n,\label{pt3-3}
  \end{equation}
   by ${\cal B}_1, {\cal B}_2$.
  To see this
  \[ b_n + |\sigma_{j,k} | \ge | \hat{\sigma}_{j,k}| \ge \omega_n \hat{\theta}_{j,k} > 2 b_n.\]
  Next
  \begin{equation}
  |\hat{\sigma}_{j,k} |< \omega_n \hat{\theta}_{j,k} \quad {\mbox implies} \quad |\sigma_{j,k}| < b_n + C_U \omega_n,\label{pt3-4}
  \end{equation}
 via Lemma \ref{la4}, and event ${\cal B}_1$
 \[\omega_n  C_U \ge \omega_n \hat{\theta}_{j,k} >  | \hat{\sigma}_{j,k}| > |\sigma_{j,k}| - b_n.\]

 Let $\|A\|_{l_2}$ be the spectral norm of matrix A.Then
 \begin{eqnarray*}
 \| \hat{\Sigma}_u^{Th} - \Sigma_u \|_{l_2} & \le & \max_{1 \le j \le J}  \sum_{k=1}^J |\hat{\sigma}_{j,k} \1_{ \{ | \hat{\sigma}_{j,k}| \ge \omega_n \hat{\theta}_{j,k} \}} - 
 \sigma_{j,k}| \\
 & \le & 
 \max_{1 \le j \le J}  \sum_{k=1}^J |\hat{\sigma}_{j,k}  - \sigma_{j,k}|  \1_{ \{ | \hat{\sigma}_{j,k}| \ge \omega_n \hat{\theta}_{j,k} \}} \\
 & + &  \max_{1 \le j \le J} \sum_{k=1}^J | \sigma_{j,k}| \1_{ \{ | \hat{\sigma}_{j,k}| < \omega_n \hat{\theta}_{j,k} \}} \\
 & \le &  \max_{1 \le j \le J} \sum_{k=1}^J
 | \hat{\sigma}_{j,k} - \sigma_{j,k}| \1_{ \{ |\sigma_{j,k} | \ge b_n \}} + \max_{1 \le j \le J} \sum_{k=1}^J |\sigma_{j,k}| \1_{ \{ |\sigma_{j,k} | 
 < b_n + C_U \omega_n\}} \\
 & \le & b_n s_n + (b_n + C_U \omega_n )s_n  \le (C_L + C_U) \omega_n s_n = C_3 \omega_n s_n
    \end{eqnarray*}
    where the first inequality is due to $\| A \|_{l_2} \le \| A \|_{l_{\infty}}$ for matrix norms, with $A$ being a symmetric matrix, and the third inequality is by (\ref{pt3-3})(\ref{pt3-4}), and the fourth inequality is by ${\cal B}_1$ and the sparsity definition, and the last inequality is due to (\ref{pt3-1}).  The last equality is $C_3:= C_L + C_U$.
    
    Note that by (\ref{pt3-1})
        \[  P ({\cal B}_2 )= P (\min_{1 \le j \le J, 1 \le k \le J} \hat{\theta}_{j,k} \omega_n> 2b_n) \ge P (\min_{1 \le j \le J, 1 \le k \le J} \hat{\theta}_{j.k}    > C_L).\]
    Then by Lemma \ref{la4}, and Lemma \ref{nl3}
    we have 
    \[ P [E] \ge 1 - o(1).\]
    
    {\hfill \bf Q.E.D.}\\
 
\ref{thm3_ii} Given $Eigmin (\Sigma_u) \ge c > 0$, and \ref{thm3_i}, the proof follows exactly as in proof of  Theorem 2.1(ii) of \cite{fan2011}.
 
{\hfill \bf Q.E.D.}\\

{\bf Proof of Lemma \ref{l3a}}.

  There are three steps in this proof. 
  
  {\bf Step 1}.
  
 We start with definitions. 
 From the definition of the covariance matrix, we can simplify  the estimator for the covariance matrix for function of factors
 \[ \hat{\Sigma}^f= \frac{1}{n} \sum_{i=1}^n \hat{f} (X_i) \hat{f} (X_i)' - \bar{f} (X_i) \bar{f} (X_i)',\]
 where $(j,k)$ th element can be written as 
 \begin{equation}
 \hat{\Sigma}_{j,k}^f = \frac{1}{n} \sum_{i=1}^n \hat{f}_j (X_i) \hat{f}_k (X_i) - \bar{f}_j (X_i) \bar{f}_k (X_i),\label{def1}
 \end{equation}
 with $\bar{f}_j (X_i):= \frac{1}{n} \sum_{i=1}^n \hat{f}_j (X_i), \bar{f}_k (X_i):= \frac{1}{n} \sum_{i=1}^n \hat{f}_k (X_i)$. Now define the infeasible covariance matrix estimator for function of factors
 \begin{equation}
 \bar{\Sigma}^f:= \frac{1}{n} \sum_{i=1}^n [ f_0 (X_i) - \bar{f}_0 (X_i)][ f_0 (X_i) - \bar{f}_0 (X_i)]' =
 \frac{1}{n} \sum_{i=1}^n f_0 (X_i) f_0 (X_i)' - \bar{f}_0 (X_i) \bar{f}_0 (X_i)',\label{def2}
  \end{equation}
 with $\bar{f}_0 (X_i):=\frac{1}{n} \sum_{i=1}^n f_0 (X_i): J \times 1$, and 
 \[ f_0 (X_i):= (f_{0,1} (X_i), \cdots, f_{0,j} (X_i), \cdots, f_{0,J } (X_i))': \quad J \times 1.\]
 The $(j,k)$ th element of $\bar{\Sigma}^f$ is
 \[ \bar{\Sigma}_{j,k}^f:= \frac{1}{n} \sum_{i=1}^n f_{0,j} (X_i) f_{0,k} (X_i) - \bar{f}_{0,j} (X_i) \bar{f}_{0,k} (X_i),\]
 with $\bar{f}_{0,j} (X_i):= \frac{1}{n} \sum_{i=1}^n f_{0,j} (X_i)$, and $\bar{f}_{0,k}$ is defined in the same way, $k$ th asset replacing $j$ th asset.

 {\bf Step 2}.
 
 Now we start with the following triangle inequality.
 
 \begin{equation}
\| \hat{\Sigma}^f - \Sigma^f \|_{\infty} \le \| \hat{\Sigma}^f - \bar{\Sigma}^f \|_{\infty} + \| \bar{\Sigma}^f - \Sigma^f \|_{\infty}.\label{pl3-1}
\end{equation}
In (\ref{pl3-1}) we start with the first term on the right side  and use definitions (\ref{def1})(\ref{def2}) with triangle inequality
{\small
\begin{eqnarray}
\| \hat{\Sigma}^f - \bar{\Sigma}^f \|_{\infty} & \le & \max_{1 \le j \le J, 1 \le k \le J} \left| \frac{1}{n} \sum_{i=1}^n \hat{f}_j (X_i) \hat{f}_k (X_i) 
- \frac{1}{n} \sum_{i=1}^n f_{0,j} (X_i) f_{0,k} (X_i)
\right| \nonumber \\
& + & 
\max_{1 \le j \le J, 1 \le k \le J}
\left|  ( \frac{1}{n} \sum_{i=1}^n \hat{f}_j (X_i)) ( \frac{1}{n} \sum_{i=1}^n \hat{f}_k (X_i))  - ( \frac{1}{n} \sum_{i=1}^n f_{0,j} (X_i)) 
( \frac{1}{n} \sum_{i=1}^n f_{0,k} (X_i)) \right|.\label{pl3-2}
\end{eqnarray}
}

{\bf Step 2a}

In (\ref{pl3-2}) we consider the first right side term, by using $\hat{f}_j (X_i) = [\hat{f}_j (X_i) - f_{0,j} (X_i) ]+ f_{0,j} (X_i)$, and repeating the same for 
$\hat{f}_k (X_i)$ and triangle inequality
{\small
\begin{eqnarray}
\max_{1 \le j \le J, 1 \le k \le J} \left| \frac{1}{n} \sum_{i=1}^n \hat{f}_j (X_i) \hat{f}_k (X_i) - \frac{1}{n} \sum_{i=1}^n f_{0,j} (X_i) f_{0,k} (X_i) 
\right| & \le & \max_{1 \le j \le J, 1 \le k \le J} \left| \frac{1}{n} \sum_{i=1}^n (\hat{f}_j (X_i) - f_{0,j} (X_i))(\hat{f}_k (X_i) - f_{0,k} (X_i))
\right| \nonumber \\
& + & \max_{1 \le j \le J, 1 \le k \le J} \left| \frac{1}{n} \sum_{i=1}^n (\hat{f}_j (X_i) - f_{0,j} (X_i))( f_{0,k} (X_i))
\right| \nonumber \\
& + & \max_{1 \le j \le J, 1 \le k \le J} \left| \frac{1}{n} \sum_{i=1}^n (f_{0,j} (X_i))(\hat{f}_k (X_i) - f_{0,k} (X_i))
\right|.\label{pl3-3} 
\end{eqnarray}}
 Next consider the first right side term in (\ref{pl3-3}), by Cauchy-Schwartz inequality
 \begin{eqnarray}
 \max_{1 \le j \le J, 1 \le k \le J} \left| \frac{1}{n} \sum_{i=1}^n (\hat{f}_j (X_i) - f_{0,j} (X_i))(\hat{f}_k (X_i) - f_{0,k} (X_i))
\right|& \le & \max_{1 \le j \le J} \sqrt{\frac{1}{n} \sum_{i=1}^n (\hat{f}_j (X_i) - f_{0,j} (X_i))^2} \nonumber \\
& \times & \max_{1 \le k \le J} \sqrt{\frac{1}{n} \sum_{i=1}^n (\hat{f}_k (X_i) - f_{0,k} (X_i))^2} \nonumber \\
&  =& O_p (l_{n}),\label{pl3-4}
 \end{eqnarray}
 where the rate is by Remark 2-Theorem \ref{t1} and Assumption \ref{asa3}. In the same way as in (\ref{pl3-4}) via Cauchy-Schwartz inequality, consider the second term on the right side of (\ref{pl3-3})
 \begin{eqnarray}
\max_{1 \le j \le J, 1 \le k \le J} \left| \frac{1}{n} \sum_{i=1}^n (\hat{f}_j (X_i) - f_{0,j} (X_i))( f_{0,k} (X_i))
\right| & \le & \max_{1 \le j \le J} \sqrt{\frac{1}{n} \sum_{i=1}^n (\hat{f}_j (X_i) - f_{0,j} (X_i))^2} \nonumber \\
& \times & \max_{1 \le k \le J} \sqrt{\frac{1}{n} \sum_{i=1}^n f_{0,k} (X_i)^2} = O_p (l_{n}^{1/2}),\label{pl3-5}
\end{eqnarray}
where we use Remark 2-Theorem \ref{t1}, Assumption \ref{asa3}, and uniformly bounded functions $f_{0,k} (X_i)$ in Theorem \ref{t1} statement for the rate. Next we use the same analysis for the third term on the right side of (\ref{pl3-3}) and combine (\ref{pl3-4})(\ref{pl3-5}) in (\ref{pl3-3}) to have 

\begin{equation}
\max_{1 \le j \le J, 1 \le k \le J} \left| \frac{1}{n} \sum_{i=1}^n \hat{f}_j (X_i) \hat{f}_k (X_i) - \frac{1}{n} \sum_{i=1}^n f_{0,j} (X_i) f_{0,k} (X_i) 
\right|  = O_p (l_{n}^{1/2}).\label{pl3-6}
  \end{equation}

  {\bf Step 2b}. We consider the second term on the right side of (\ref{pl3-2}). Add and subtract
  \begin{eqnarray}
  \left( \frac{1}{n} \sum_{i=1}^n \hat{f}_j (X_i)
  \right)\left( \frac{1}{n} \sum_{i=1}^n \hat{f}_k (X_i)
  \right) & - &   \left( \frac{1}{n} \sum_{i=1}^n f_{0,j} (X_i)
  \right)\left( \frac{1}{n} \sum_{i=1}^n f_{0,k} (X_i)
  \right)\nonumber \\
  & = & \left( \frac{1}{n} \sum_{i=1}^n(\hat{f}_j (X_i) -  f_{0,j} (X_i))
  \right)\left( \frac{1}{n} \sum_{i=1}^n(\hat{f}_k (X_i) -  f_{0,k} (X_i))
  \right) \nonumber \\
  & + & \left( \frac{1}{n} \sum_{i=1}^n(\hat{f}_j (X_i) -  f_{0,j} (X_i))
  \right)  \left( \frac{1}{n} \sum_{i=1}^n f_{0,k} (X_i)
  \right) \nonumber \\
  & + & \left( \frac{1}{n} \sum_{i=1}^n(\hat{f}_k (X_i) -  f_{0,k} (X_i))
  \right)  \left( \frac{1}{n} \sum_{i=1}^n f_{0,j} (X_i)
  \right).\label{pl3-7}
    \end{eqnarray}

    In (\ref{pl3-7}) by $l_1, l_2$ norm inequality we get  

\[
  \frac{1}{n} \sum_{i=1}^n (\hat{f}_j (X_i) - f_{0,j} (X_i)) \le 
  \frac{1}{n} \sum_{i=1}^n |(\hat{f}_j (X_i) - f_{0,j} (X_i))|    \le \sqrt{\frac{1}{n} \sum_{i=1}^n (\hat{f}_j (X_i) - f_{0,j} (X_i))^2
  }.\]
  The same analysis is applied to  $\frac{1}{n} \sum_{i=1}^n (\hat{f}_k (X_i) - f_{0,k} (X_i))$. The first term on the right side of  (\ref{pl3-7}) can be upper bounded as   
  {\small \begin{eqnarray}
 \max_{ 1 \le j \le J, 1 \le k \le J} \left| \left( \frac{1}{n} \sum_{i=1}^n(\hat{f}_j (X_i) -  f_{0,j} (X_i))
  \right)\left( \frac{1}{n} \sum_{i=1}^n(\hat{f}_k (X_i) -  f_{0,k} (X_i))
  \right)  \right| &\le& \max_{1 \le j \le J} \sqrt{\frac{1}{n} \sum_{i=1}^n (\hat{f}_j (X_i) - f_{0,j} (X_i))^2
  } \nonumber \\
  & \times & 
  \max_{1 \le k \le J} \sqrt{\frac{1}{n} \sum_{i=1}^n (\hat{f}_k (X_i) - f_{0,k} (X_i))^2
  } \nonumber \\
  &  =& O_p (l_{n}),\label{pl3-7a}
  \end{eqnarray}}
  by Remark 1-Theorem \ref{t1}, and Assumption \ref{asa3}. Then in (\ref{pl3-7}) 
 \begin{eqnarray}
  \max_{1 \le j \le j, 1 \le k \le J} \left| 
  \left( \frac{1}{n} \sum_{i=1}^n(\hat{f}_j (X_i) -  f_{0,j} (X_i))
  \right)  \left( \frac{1}{n} \sum_{i=1}^n f_{0,k} (X_i)
  \right)   \right| & \le & \max_{1 \le j \le J} \sqrt{\frac{1}{n} \sum_{i=1}^n (\hat{f}_j (X_i) - f_{0,j} (X_i))^2
  } \nonumber \\
  & \times & \max_{1 \le k \le J} \left| \frac{1}{n} \sum_{i=1}^n f_{0,k} (X_i)
  \right| \nonumber \\
  &  =& O_p (l_{n}^{1/2}),\label{pl3-7b}
  \end{eqnarray}
  by Remark 2-Theorem \ref{t1}, Assumption \ref{asa3} and uniformly bounded $f_{0,k} (X_i)$ by Assumption. Same analysis in (\ref{pl3-7b}) applies to third term on the right side of (\ref{pl3-7})
   hence combine (\ref{pl3-7a})(\ref{pl3-7b}) in (\ref{pl3-7}) 
   \begin{equation}
   \max_{1 \le j \le J, 1 \le k \le J} \left| \left( \frac{1}{n} \sum_{i=1}^n \hat{f}_j (X_i)
   \right) \left( \frac{1}{n} \sum_{i=1}^n \hat{f}_k (X_i)   \right) - \left(\frac{1}{n} \sum_{i=1}^n f_{0,j} (X_i)
   \right) \left( \frac{1}{n} \sum_{i=1}^n f_{0,k} (X_i)  \right)  \right| = O_p (l_{n}^{1/2}).\label{pl3-8}
      \end{equation}
  Use (\ref{pl3-6})(\ref{pl3-8}) in (\ref{pl3-2}) to have
  \begin{equation}
  \| \hat{\Sigma}^f - \bar{\Sigma}^f \|_{\infty} = O_p (l_{n}^{1/2}).\label{pl3-9}
  \end{equation}
  
  {\bf Step 2c}. In (\ref{pl3-1}) consider the second term on the right side
  
  \begin{eqnarray}
  \| \bar{\Sigma}^f - \Sigma^f \|_{\infty} & \le & \max_{1 \le j \le J, 1 \le k \le J} \left| \frac{1}{n} \sum_{i=1}^n f_{0,j} (X_i) f_{0,k} (X_i) - E [f_{0,j} (X_i) f_{0,k} (X_i)]
  \right| \nonumber \\
  & + &  \max_{1 \le j \le J, 1 \le k \le J} \left| (\frac{1}{n} \sum_{i=1}^n f_{0,j} (X_i)) ( \frac{1}{n} \sum_{i=1}^n f_{0,k} (X_i)) - E [f_{0,j} (X_i)] E [ f_{0,k} (X_i)] \right|.\label{pl3-10}
\end{eqnarray}

Take the first term on the right side of (\ref{pl3-10})
\begin{equation}
\max_{1 \le j \le J, 1 \le k \le J} \left| \frac{1}{n} \sum_{i=1}^n f_{0,j} (X_i) f_{0,k} (X_i) - E [f_{0,j} (X_i) f_{0,k} (X_i)]
  \right|  = O_p (\sqrt{\frac{\log J}{n}}),\label{pl3-11}
\end{equation}
by the proof of Lemma \ref{mtl1} \ref{mtl1_i} since $f_{0,j}(X_i), f_{0,k} (X_i)$ are uniformly bounded (subgaussian) hence their product is subexponential.
Consider the second term on the right side of (\ref{pl3-10}) by adding and subtracting, and triangle inequality

{\small
\begin{eqnarray}
&& \max_{1 \le j \le J, 1 \le k \le J} \left| (\frac{1}{n} \sum_{i=1}^n f_{0,j} (X_i)) ( \frac{1}{n} \sum_{i=1}^n f_{0,k} (X_i)) - E [f_{0,j} (X_i)] E [ f_{0,k} (X_i)] \right|  \nonumber \\
&\le &  
\max_{ 1 \le j \le J, 1 \le k \le J} \left| \left( \frac{1}{n} \sum_{i=1}^n (f_{0,j} (X_i) - E [f_{0,j} (X_i)])\right) \left( \frac{1}{n} \sum_{i=1}^n (f_{0,k} (X_i) - E [f_{0,k} (X_i)])\right)\right|
\nonumber \\
& + & 
\max_{ 1 \le j \le J, 1 \le k \le J} \left| \left( \frac{1}{n} \sum_{i=1}^n (f_{0,j} (X_i) - E [f_{0,j} (X_i)])\right) E [ f_{0,k} (X_i)] \right| \nonumber \\
& + & \max_{ 1 \le j \le J, 1 \le k \le J} \left| \left( \frac{1}{n} \sum_{i=1}^n (f_{0,k} (X_i) - E [f_{0,k} (X_i)])\right) E [ f_{0,j} (X_i)] \right|.\label{pl3-12}
\end{eqnarray}}

  In (\ref{pl3-12}) consider the first term on the right side
{\small  
  \begin{eqnarray}
   \max_{ 1 \le j \le J, 1 \le k \le J} | \left( \frac{1}{n} \sum_{i=1}^n (f_{0,j} (X_i) - E [f_{0,j} (X_i)])\right) &\times &\left( \frac{1}{n} \sum_{i=1}^n (f_{0,k} (X_i) - E [f_{0,k} (X_i)])\right)| \nonumber \\
&\le& \max_{1 \le j \le J} \left| \frac{1}{n} \sum_{i=1}^n (f_{0,j} (X_i) - E [ f_{0,j} (X_i)]) 
\right| \nonumber \\
& \times & \max_{1 \le k \le J}
\left| \frac{1}{n} \sum_{i=1}^n (f_{0,k} (X_i) - E [ f_{0,k} (X_i)]) 
\right|.\label{pl3-13}
\end{eqnarray}}

  By Hoeffding inequality in (2.11) of \cite{w2019}, since our functions are uniformly bounded (subgaussian), then taking the union bound 
  with $t_2 = 2 F \sqrt{ \log 2J/n}$
    \begin{eqnarray}
  P \left[ \max_{1 \le j \le J} \left| \frac{1}{n} \sum_{i=1}^n f_{0,j} (X_i) - E f_{0,j} (X_i)
  \right| \ge t_2 \right] & \le & 
  2 J \exp \left(\frac{-2n}{4 F^2} (t_2^2)\right) \nonumber \\
  & = & \frac{1}{2J}.\label{pl3-14}
  \end{eqnarray}
  Second term on the right side of (\ref{pl3-13}) is handled in the same way as in (\ref{pl3-14}) hence
  \begin{equation}
   \max_{ 1 \le j \le J, 1 \le k \le J} \left| \left( \frac{1}{n} \sum_{i=1}^n (f_{0,j} (X_i) - E [f_{0,j} (X_i)])\right) \left( \frac{1}{n} \sum_{i=1}^n (f_{0,k} (X_i) - E [f_{0,k} (X_i)])\right) \right|  = O_p \left(\frac{\log J}{n}\right).\label{pl3-15}
  \end{equation}
  Since our functions are uniformly bounded we have $E| f_{0,k} (X_i)| \le F$, second term on the right side of (\ref{pl3-12}) uses (\ref{pl3-14})
  \begin{equation}
  \max_{ 1 \le j \le J, 1 \le k \le J} \left| \left( \frac{1}{n} \sum_{i=1}^n (f_{0,j} (X_i) - E [f_{0,j} (X_i)])\right) E [ f_{0,k} (X_i)] \right|  = O_p  \left( \sqrt{\frac{\log J}{n}}
  \right).\label{pl3-15a}
  \end{equation}
  Third term on the right side of (\ref{pl3-12}) follows the same analysis in (\ref{pl3-15a}) above hence via second term on the right side of (\ref{pl3-10}) and the analysis in (\ref{pl3-13})-(\ref{pl3-15a})
  \begin{equation}
  \max_{1 \le j \le J, 1 \le k \le J} \left| (\frac{1}{n} \sum_{i=1}^n f_{0,j} (X_i)) ( \frac{1}{n} \sum_{i=1}^n f_{0,k} (X_i)) - E [f_{0,j} (X_i)] E [ f_{0,k} (X_i)]  \right| = O_p \left( \sqrt{\frac{\log J}{n}} 
\right).\label{pl3-15b}
  \end{equation}
  
  {\bf Step 3}. Use (\ref{pl3-11}) and (\ref{pl3-15b}) in (\ref{pl3-10}) to have 
  \begin{equation}
  \| \bar{\Sigma}^f - \Sigma \|_{\infty} = O_p \left( \frac{\sqrt{\log J}}{\sqrt{n}} \right).\label{pl3-16}
   \end{equation}
  Since rate $l_{n}^{1/2}$ is always slower than or equal to  $\sqrt{\log J/n}$ due to Assumption \ref{asa3}, and see Remark 2-Theorem \ref{t1} as well, combine (\ref{pl3-9}) and (\ref{pl3-16}) in (\ref{pl3-1}) to have 
  \[ \| \hat{\Sigma}^f - \Sigma \|_{\infty} = O_p \left(  l_{n}^{1/2} \right).\]

{\hfill \bf Q.E.D.}\\
   
  {\bf Proof of Theorem \ref{covret}}. 
  Clearly  by definitions of $\hat{\Sigma}_y, \Sigma_y$,
  \begin{equation}
   \| \hat{\Sigma}_y  - \Sigma_y \|_{\infty} \le \| \hat{\Sigma}^f - \Sigma^f \|_{\infty} + \| \hat{\Sigma}_u^{Th} - \Sigma_u \|_{\infty}.\label{pt3a-1}
   \end{equation}
  In the inequality above we analyze the second right side term, using the same technique as in the proof of Theorem \ref{thm3}. In that sense
   we have the following definitions, $\sigma_{j,k}:= E u_{j,i} u_{k,i}$, and by $b_n = C_{*} l_{n}^{1/2}$
  \[ {\cal B}_1:= \{ \max_{1 \le k \le J} \max_{1 \le j \le J} | \hat{\sigma}_{j,k} - \sigma_{j,k}| \le b_n \},\]
  \[ {\cal B}_2:= \{ \min_{1 \le k \le J} \min_{1 \le j \le J} \hat{\theta}_{j,k} \omega_n > 2 b_n \} ,\]
  \[ {\cal B}_3:= \{ \max_{1 \le k \le J} \max_{1 \le j \le J} \hat{\theta}_{j,k} \le C_U \}.\]
  
  Define also the event $E: \{ \cap_{l=1}^3 {\cal B}_l \}$. Under event $E$  
  \begin{eqnarray*}
 \| \hat{\Sigma}_u^{Th} - \Sigma_u \|_{\infty} & = & \max_{1 \le j \le J}  \max_{1 \le k \le J}  |\hat{\sigma}_{j,k} \1_{ \{ | \hat{\sigma}_{j,k}| \ge \omega_n \hat{\theta}_{j,k} \}} - 
 \sigma_{j,k}| \\
 & \le & 
 \max_{1 \le j \le J}  \max_{1 \le k \le J } |\hat{\sigma}_{j,k}  - \sigma_{j,k}|  \1_{ \{ | \hat{\sigma}_{j,k}| \ge \omega_n \hat{\theta}_{j,k} \}} \\
 & + &  \max_{1 \le j \le J} \max_{1 \le k \le J }  | \sigma_{j,k}| \1_{ \{ | \hat{\sigma}_{j,k}| < \omega_n \hat{\theta}_{j,k} \}} \\
 & \le &  \max_{1 \le j \le J} \max_{1 \le k \le J} 
 | \hat{\sigma}_{j,k} - \sigma_{j,k}| \1_{ \{ |\sigma_{j,k} | \ge b_n \}} + \max_{1 \le j \le J} \max_{1 \le k \le J} |\sigma_{j,k}| \1_{ \{ |\sigma_{j,k} | 
 < b_n + C_U \omega_n\}} \\
 & \le & b_n + (b_n + C_U \omega_n )  \le (C_L + C_U) \omega_n,
    \end{eqnarray*}  
  where the first equality is the definition, and the second inequality is by (\ref{pt3-3})(\ref{pt3-4}), and the third inequality is by ${\cal B}_1$ definition, and the last inequality is by (\ref{pt3-1}). Following the proof of Theorem \ref{thm3}
  \[ P ( E ) \ge 1 - o(1),\]
  so 
  \[ \| \hat{\Sigma}_u^{Th} - \Sigma_u \|_{\infty} = O_p ( \omega_n).\]
  
  Then using Lemma \ref{l3a}, and since $\omega_n = O (l_{n}^{1/2} )$  we have by the last result in (\ref{pt3a-1})
  \[ \| \hat{\Sigma}_y  - \Sigma_y \|_{\infty} = O _p ( \omega_n).\]
  
{\hfill \bf Q.E.D.}\\  
  
\vspace{0.5in}

  {\bf \Large Consistency of Estimate of Precision Matrix of Returns }\\
  
  This part of Appendix analyzes estimate for the precision matrix of returns. We need several Lemmata to begin with. Our results in this part of the paper only allow $J <<n$.

 \begin{lemma}\label{la5}
 Under Assumptions \ref{as1}-\ref{as2}, \ref{asa4}-\ref{as9} 
 \[\| (\hat{\Sigma}_u^{Th})^{-1}  \hat{\Sigma}^f - \Sigma_u^{-1} \Sigma^f \|_{l_2} = O_p (J \omega_n s_n).\]
\end{lemma}  
  
  {\bf Proof of \ref{la5}}. First by adding and subtracting $(\hat{\Sigma}_u^{Th})^{-1} \Sigma_f$ for the first inequality and then add and subtract 
  $\Sigma_u^{-1} (\hat{\Sigma}^f - \Sigma^f)$ for the second inequality
   and triangle inequality  
  \begin{eqnarray}
 \| (\hat{\Sigma}_u^{Th})^{-1}  \hat{\Sigma}^f - \Sigma_u^{-1} \Sigma^f \|_{l_2} & \le & 
\| (\hat{\Sigma}_u^{Th})^{-1} ( \hat{\Sigma}^f - \Sigma^f) \|_{l_2} \nonumber \\
& + &
 \| [(\hat{\Sigma}_u^{Th})^{-1}  - \Sigma_u^{-1} ] \Sigma^f \|_{l_2} \nonumber \\
& \le & \| [(\hat{\Sigma}_u^{Th})^{-1} - \Sigma_u^{-1}] [ \hat{\Sigma}^f - \Sigma^f] \|_{l_2} \\
& + & \|  [(\hat{\Sigma}_u^{Th})^{-1} - \Sigma_u^{-1}] \Sigma^f \|_{l_2} + \| \Sigma_u^{-1}
(\hat{\Sigma}^f - \Sigma^f)\|_{l_2}.\label{pla5-1}
\end{eqnarray}
Analyze each term. See that 
\begin{eqnarray}
\| [(\hat{\Sigma}_u^{Th})^{-1} - \Sigma_u^{-1}] [ \hat{\Sigma}^f - \Sigma^f] \|_{l_2}
& \le & \| [(\hat{\Sigma}_u^{Th})^{-1} - \Sigma_u^{-1}] \|_{l_2} \| [ \hat{\Sigma}^f - \Sigma^f] \|_{l_2} \nonumber \\
& \le & J \| [(\hat{\Sigma}_u^{Th})^{-1} - \Sigma_u^{-1}] \|_{l_2} \| [ \hat{\Sigma}^f - \Sigma^f] \|_{\infty} \nonumber \\
& = & J O_p (\omega_n s_n) O_p (l_{n}^{1/2}),\label{pla5-2}
\end{eqnarray}
by norm inequality that ties spectral norm to $\|.\|_{\infty}$ norm in p.365 of \cite{hj2013}, and by Theorem \ref{thm3}, and Lemma \ref{l3a}. Then in (\ref{pla5-1}) consider 
\begin{eqnarray}
\|  [(\hat{\Sigma}_u^{Th})^{-1} - \Sigma_u^{-1}] \Sigma^f \|_{l_2}  & \le & \|  [(\hat{\Sigma}_u^{Th})^{-1} - \Sigma_u^{-1}] \|_{l_2}
\| \Sigma^f \|_{l_2} \nonumber \\
& = & O_p (\omega_n s_n) O(J),\label{pla5-3}
\end{eqnarray} 
by Theorem \ref{thm3} and Assumption \ref{as8}since  $\| \Sigma^f \|_{l_2} = Eigmax (\Sigma^f) = O(J)$. Next, on the right side of (\ref{pla5-1}) 
\begin{equation}
\| \Sigma_u^{-1}
(\hat{\Sigma}^f - \Sigma^f)\|_{l_2}  \le  \| \Sigma_u^{-1} \|_{l_2} \| \hat{\Sigma}^f - \Sigma^f\|_{l_2}  \le J \| \Sigma_u^{-1} \|_{l_2} \| \hat{\Sigma}^f - \Sigma^f\|_{\infty}=  O_p (J l_{n}^{1/2}),\label{pla5-4}
   \end{equation}
where we use  Lemma \ref{l3a}, and norm inequality that ties spectral norm to $\|.\|_{\infty}$ norm in p.365 of \cite{hj2013}, and 
\begin{equation}
\| \Sigma_u^{-1} \|_{l_2} = Eigmax (\Sigma_u^{-1}) = \frac{1}{Eigmin(\Sigma_u)} \le \frac{1}{c} < \infty.\label{12a}
\end{equation}
by Assumption 1.  Note that $l_{n}^{1/2} \to 0$, and $\omega_n s_n \to 0$ by Assumption \ref{as9}, since 
$\omega_n = O (l_{n}^{1/2})$, so the rate in (\ref{pla5-3}) is the slowest.

{\hfill \bf Q.E.D.}\\

\begin{lemma}\label{la6}
 Under Assumptions \ref{as1}-\ref{as2},\ref{asa4}-\ref{as9}, with $e_n \to 0$ or $e_n =0$, with  $\delta_n \to \infty$
\leavevmode
\begin{enumerate}[label=(\roman*)]
	\item \[ \left\| \left[I_J + \Sigma_u^{-1} \Sigma^f \right]^{-1} \right\|_{l_2} =  O (\frac{\delta_n}{\delta_n + e_n}) = O( 1).\] \label{la6_i}
	
	\item \[ \left\| \left[ I_J + (\hat{\Sigma}_u^{Th})^{-1} \hat{\Sigma}^f \right]^{-1} \right\|_{l_2} =  O_p (\frac{\delta_n}{\delta_n + e_n}) = O_p( 1).\] \label{la6_ii}
\end{enumerate}

\end{lemma} 

{\bf Proof of Lemma \ref{la6}}. 

\ref{la6_i} We start the proof with following eigenvalue inequality in \cite{abamag2005}, which is on p.344  as Exercise 12.40a, with a proof, for $A, B$ symmetric matrices
\begin{equation}
Eigmin (A + B ) \ge Eigmin (A) + Eigmin (B).\label{mevalin}
\end{equation}
Now apply (\ref{mevalin}) with $A= I_J, B= \Sigma_u^{-1} \Sigma^f$ in the first inequality below
\begin{eqnarray}
Eigmin (I_J + \Sigma_u^{-1} \Sigma^f) & \ge & Eigmin (I_J) + Eigmin (\Sigma_u^{-1} \Sigma^f) \nonumber \\
& \ge & 1+ Eigmin (\Sigma_u^{-1}) Eigmin (\Sigma^f) \nonumber \\
& = & 1+ \frac{1}{Eigmax (\Sigma_u)} Eigmin (\Sigma^f) \nonumber \\
& \ge & 1+ \frac{ e_n }{C \delta_n},\label{pla6-1}
\end{eqnarray}
by Fact 10.22.23 of p.809 of \cite{bern2018}, and the rest is by Assumptions \ref{as8}-\ref{as9}. Note that 
\[ Eigmax [(I_J + \Sigma_u^{-1} \Sigma^f)^{-1}] = \frac{1}{Eigmin [(I_J + \Sigma_u^{-1} \Sigma^f)]} = O (\frac{1}{1+ \frac{e_n}{\delta_n}}) = O ( \frac{\delta_n}{\delta_n + e_n}).\]
\ref{la6_ii}
\[ \| \left[ I_J + (\hat{\Sigma}_u^{Th})^{-1} \hat{\Sigma}^f \right] -\left[I_J + \Sigma_u^{-1} \Sigma^f \right]\|_{l_2}
= \| (\hat{\Sigma}_u^{Th})^{-1} \hat{\Sigma}^f - \Sigma_u^{-1} \Sigma^f \|_{l_2} = O_p (J \omega_n s_n),\]
by Lemma \ref{la5}. Then by Assumption \ref{as9}, $J  \omega_n s_n \to 0$, and Assumption \ref{as8}\ref{as8_ii}, $1 + e_n/C \delta_n \to 1$ or exactly 1 depending on $e_n$, and  via Lemma A.1(i) of \cite{fan2011}
\begin{eqnarray}
 P \left[ Eigmin (I_J + (\hat{\Sigma}_u^{Th})^{-1} \hat{\Sigma}^f ) \ge0.5[ 1+ e_n/ C \delta_n ] \right]
&\ge& P \left[ \| \left[ I_J + (\hat{\Sigma}_u^{Th})^{-1} \hat{\Sigma}^f \right] -\left[I_J + \Sigma_u^{-1} \Sigma^f \right]\|_{l_2}
\le  0.5[1+ e_n/C \delta_n]
\right] \nonumber \\ 
&\ge &1 - o(1).\label{la49a}
\end{eqnarray}
Since 
\begin{eqnarray*}
 \left\| \left[I_J + (\hat{\Sigma}_u^{Th})^{-1} \hat{\Sigma}^f)\right]^{-1} \right\|_{l_2} & = & Eigmax \left[ I_J + (\hat{\Sigma}_u^{Th})^{-1} \hat{\Sigma}^f)\right]^{-1} \\
 & = & \frac{1}{Eigmin ( I_J + (\hat{\Sigma}_u^{Th})^{-1} \hat{\Sigma}^f)},
\end{eqnarray*}
and by (\ref{la49a})
\[ \left\| \left[I_p + (\hat{\Sigma}_u^{Th})^{-1} \hat{\Sigma}^f)\right]^{-1} \right\|_{l_2} = O_p (\frac{\delta_n}{\delta_n + e_n})=O_p (1),\]
since either $\frac{\delta_n}{\delta_n + e_n} =1, $ or $\frac{\delta_n}{ \delta_n + e_n }\to 1$ by Assumption \ref{as8}\ref{as8_ii}.

{\hfill \bf Q.E.D.}\\

\begin{lemma}\label{la7}
 Under Assumptions \ref{as1}-\ref{as2},\ref{asa4}-\ref{as9} 
 \leavevmode
 \begin{enumerate}[label=(\roman*)]
 	\item \[ \| \hat{\Sigma}^f  \|_{l_2} = O_p (J).\] \label{la7_i}
 	
 	\item \[ \| (\hat{\Sigma}_u^{Th})^{-1}  \|_{l_2} = O_p (1).\] \label{la7_ii}
\end{enumerate}
\end{lemma}

{\bf Proof of Lemma \ref{la7}}.

\ref{la7_i} \begin{eqnarray*}
\| \hat{\Sigma}^f \|_{l_2} &\le & \| \hat{\Sigma}^f - \Sigma^f \|_{l_2} + \| \Sigma^f \|_{l_2} \\
& \le & [ J \| \hat{\Sigma}^f - \Sigma_f \|_{\infty} ] + \| \Sigma^f \|_{l_2} \\
& = & O_p ( J l_{n}^{1/2}) + O (J) = O_p (J),\label{pla7-1}
\end{eqnarray*}
by p.365 of \cite{hj2013}, tying $\|.\|_{l_2}$ spectral norm to $\|.\|_{\infty}$ norm, and via Lemma \ref{l3a}, and $l_{n} \to 0$.

\ref{la7_ii}
\begin{eqnarray*}
\| (\hat{\Sigma}_u^{Th})^{-1} \|_{l_2} & \le & \| (\hat{\Sigma}_u^{Th})^{-1} - \Sigma_u^{-1} \|_{l_2} + \| \Sigma_u^{-1} \|_{l_2}\\
& = & O_p ( \omega_n s_n ) + O(1) \\
& = & O_p (1),
\end{eqnarray*}
by Theorem \ref{thm3}, Assumption \ref{as9} implies $\omega_n s_n \to 0$, and 
\[ \| \Sigma_u^{-1} \|_{l_2} = Eigmax (\Sigma_u^{-1}) = \frac{1}{Eigmin (\Sigma_u)} \le \frac{1}{c} < \infty.\]

{\hfill \bf Q.E.D.}\\

{\bf Proof of Theorem \ref{thm4}}

First define
\begin{equation}
\hat{G}:= [ I_J + (\hat{\Sigma}_u^{Th})^{-1} \hat{\Sigma}^f)]^{-1}.\label{g1}
\end{equation}

\begin{equation}
G:= [ I_J + \Sigma_u^{-1} \Sigma^f]^{-1}.\label{g2}
\end{equation}

Note that by Lemma \ref{la6} and definitions (\ref{g1})(\ref{g2})
\begin{equation}
\| \hat{G} \|_{l_2} = O_p (1).\label{ag1}
\end{equation}

\begin{equation}
 \| G \|_{l_2} = O (1).\label{ag2}
\end{equation}

Then by (\ref{s4-1})(\ref{s4-2})

\begin{equation}
\| \hat{\Sigma}_y^{-1} - \Sigma_y^{-1} \|_{l_2} \le \| (\hat{\Sigma}_u^{Th})^{-1} - \Sigma_u^{-1} \|_{l_2} 
+ \| (\hat{\Sigma}_u^{Th})^{-1}  \hat{\Sigma}^f \hat{G} (\hat{\Sigma}_u^{Th})^{-1}  - \Sigma_u^{-1} \Sigma^f  G \Sigma_u^{-1} \|_{l_2}.\label{pt4-1}
\end{equation}
We consider the second right side term in (\ref{pt4-1}), and one issue is the simplification of that term so that we can benefit from Lemmata here. In that respect,
add and subtract $\Sigma_u^{-1} \hat{\Sigma}^f \hat{G} (\hat{\Sigma}_u^{Th})^{-1}$ and use triangle inequality
\begin{eqnarray}
\| (\hat{\Sigma}_u^{Th})^{-1}  \hat{\Sigma}^f \hat{G} (\hat{\Sigma}_u^{Th})^{-1}  - \Sigma_u^{-1} \Sigma^f  G \Sigma_u^{-1} \|_{l_2} & \le & 
\| [(\hat{\Sigma}_u^{Th})^{-1}  \hat{\Sigma}^f - \Sigma_u^{-1} \hat{\Sigma}^f] \hat{G} (\hat{\Sigma}_u^{Th})^{-1} \|_{l_2} \nonumber \\
& + & \| \Sigma_u^{-1} \hat{\Sigma}^f \hat{G} (\hat{\Sigma}_u^{Th})^{-1} - \Sigma_u^{-1} \Sigma^f  G \Sigma_u^{-1} \|_{l_2}.\label{pt4-2}
\end{eqnarray}

Take the second term in (\ref{pt4-2}) add and subtract $\Sigma_u^{-1} \Sigma^f \hat{G} (\hat{\Sigma}_u^{Th})^{-1}$ and use triangle inequality

\begin{eqnarray}
\| \Sigma_u^{-1} \hat{\Sigma}^f \hat{G} (\hat{\Sigma}_u^{Th})^{-1} - \Sigma_u^{-1} \Sigma^f  G \Sigma_u^{-1} \|_{l_2} & \le & 
\| \Sigma_u^{-1} (\hat{\Sigma}^f - \Sigma^f) \hat{G} (\hat{\Sigma}_u^{Th})^{-1} \|_{l_2} \nonumber \\
& + & \| \Sigma_u^{-1} \Sigma^f \hat{G} (\hat{\Sigma}_u^{Th})^{-1} - \Sigma_u^{-1}  \Sigma^f G \Sigma_u^{-1} \|_{l_2}.\label{pt4-3}
\end{eqnarray}

Take the second term on the right side of (\ref{pt4-3}), add and subtract $\Sigma_u^{-1} \Sigma^f \hat{G} \Sigma_u^{-1} $ via triangle inequality
\begin{eqnarray}
\| \Sigma_u^{-1} \Sigma^f \hat{G} (\hat{\Sigma}_u^{Th})^{-1} - \Sigma_u^{-1}  \Sigma^f G \Sigma_u^{-1} \|_{l_2} & \le & 
\| \Sigma_u^{-1} \Sigma^f \hat{G} (\hat{\Sigma}_u^{Th})^{-1} - \Sigma_u^{-1} \Sigma^f \hat{G} \Sigma_u^{-1} \|_{l_2} \nonumber \\
& + & \| \Sigma_u^{-1} \Sigma^f \hat{G} \Sigma_u^{-1} - \Sigma_u^{-1}  \Sigma^f G \Sigma_u^{-1} \|_{l_2}.\label{pt4-4}
\end{eqnarray}

Combine (\ref{pt4-3})(\ref{pt4-4}) in (\ref{pt4-2})

\begin{eqnarray}
\| (\hat{\Sigma}_u^{Th})^{-1}  \hat{\Sigma}^f \hat{G} (\hat{\Sigma}_u^{Th})^{-1}  - \Sigma_u^{-1} \Sigma^f  G \Sigma_u^{-1} \|_{l_2}
& \le & \| [(\hat{\Sigma}_u^{Th})^{-1}   - \Sigma_u^{-1} ]\hat{\Sigma}^f \hat{G} (\hat{\Sigma}_u^{Th})^{-1} \|_{l_2} \nonumber \\
& + & \| \Sigma_u^{-1} (\hat{\Sigma}^f - \Sigma^f) \hat{G} (\hat{\Sigma}_u^{Th})^{-1} \|_{l_2} \nonumber \\
& + & \| \Sigma_u^{-1} \Sigma^f \hat{G} [(\hat{\Sigma}_u^{Th})^{-1} - \Sigma_u^{-1}] \|_{l_2}\nonumber \\
& + & \| \Sigma_u^{-1} \Sigma^f (\hat{G}- G)  \Sigma_u^{-1} \|_{l_2}.\label{pt4-5}
\end{eqnarray}

We get the rates for each term on the right side of (\ref{pt4-5}). Consider the first term on the right side of (\ref{pt4-5})
\begin{eqnarray}
 \| [(\hat{\Sigma}_u^{Th})^{-1}   - \Sigma_u^{-1} ]\hat{\Sigma}^f \hat{G} (\hat{\Sigma}_u^{Th})^{-1} \|_{l_2}
 & \le & \| [(\hat{\Sigma}_u^{Th})^{-1}   - \Sigma_u^{-1} ] \|_{l_2}
 \| \hat{\Sigma}^f \|_{l_2} \| \hat{G} \|_{l_2} \| (\hat{\Sigma}_u^{Th})^{-1}\|_{l_2} \nonumber \\
 & = & O_p ( \omega_n s_n ) O_p (J) O_p (1) O_p (1) = O_p ( J \omega_n s_n) = o_p (1),\label{pt4-6}
\end{eqnarray}
by Theorem \ref{thm3}, Lemma \ref{la5}, Lemma \ref{la7}, Assumption \ref{as9}, (\ref{ag1}). Analyze the second term on the right side of (\ref{pt4-5})

\begin{eqnarray}
\| \Sigma_u^{-1} (\hat{\Sigma}^f - \Sigma^f) \hat{G} (\hat{\Sigma}_u^{Th})^{-1} \|_{l_2}
& \le & \| \Sigma_u^{-1}\|_{l_2} \|(\hat{\Sigma}^f - \Sigma^f)\|_{l_2} \| \hat{G} \|_{l_2} \| (\hat{\Sigma}_u^{Th})^{-1} \|_{l_2} \nonumber \\
& \le &  \| \Sigma_u^{-1}\|_{l_2}[ J  \|(\hat{\Sigma}^f - \Sigma^f)\|_{\infty}] \| \hat{G} \|_{l_2} \| (\hat{\Sigma}_u^{Th})^{-1} \|_{l_2} \nonumber \\
& = & O (1) O_p (J \omega_n) O_p ( 1) O_p (1) = O_p ( J  \omega_n) = o_p (1),\label{pt4-7}
\end{eqnarray}
by (\ref{12a}), by the inequality tying spectral norm to $\|.\|_{\infty}$ norm in p.365 of \cite{hj2013}, Lemma \ref{l3a}, Lemma \ref{la6}\ref{la6_i}-$\hat{G}$ definition in (\ref{g1})(\ref{ag1}), Lemma \ref{la7}\ref{la7_ii}, Assumption \ref{as9}. Consider the third term on the right side of (\ref{pt4-5})

\begin{eqnarray}
\| \Sigma_u^{-1} \Sigma^f \hat{G} [(\hat{\Sigma}_u^{Th})^{-1} - \Sigma_u^{-1}] \|_{l_2} 
& \le & \| \Sigma_u^{-1}\|_{l_2} \| \Sigma^f\|_{l_2} \| \hat{G}\|_{l_2}  \|[(\hat{\Sigma}_u^{Th})^{-1} - \Sigma_u^{-1}] \|_{l_2} \nonumber \\
& = & O (1) O (J) O_p ( 1) O_p (\omega_n s_n) = O_p ( J \omega_n s_n) = o_p (1),\label{pt4-8}
\end{eqnarray}
by (\ref{12a}), Lemma \ref{la6}\ref{la6_i}- $\hat{G}$ definition in (\ref{g1})(\ref{ag1}), Theorem \ref{thm3} with Assumptions \ref{as8}-\ref{as9}.

\noindent The fourth term on the right side of (\ref{pt4-5}) is 
\begin{eqnarray}
\| \Sigma_u^{-1} \Sigma^f (\hat{G}- G)  \Sigma_u^{-1} \|_{l_2} & \le & \| \Sigma_u^{-1} \|_{l_2}^2  \| \Sigma^f \|_{l_2} \| \hat{G} - G \|_{l_2} \nonumber \\
& \le & \| \Sigma_u^{-1} \|_{l_2}^2  \| \Sigma^f \|_{l_2} \| \hat{G} \|_{l_2} \| G \|_{l_2} \| (\hat{\Sigma}_u^{Th})^{-1} \hat{\Sigma}^f - \Sigma_u^{-1} \Sigma^f \|_{l_2} \nonumber \\
& = & O(1) O (J) O_p (1) O (1) O_p (J \omega_n s_n) = O_p (J^2 \omega_n s_n) = o_p (1),\label{pt4-9}
\end{eqnarray}
where we use $\hat{G}, G$ definitions in (\ref{g1})(\ref{g2}) with  (\ref{ag1})(\ref{ag2})
\[ \|\hat{G} - G \|_{l_2} \le \| \hat{G} \|_{l_2} \| G \|_{l_2} \| (I_J + (\hat{\Sigma}_u^{Th})^{-1} \hat{\Sigma}^f) - (I_J + \Sigma_u^{-1} \Sigma^f )\|_{l_2},\]
to get the second inequality on the right side of (\ref{pt4-9}). For the rates we use (\ref{12a}), Lemma \ref{la5}, \ref{la6} and Assumptions \ref{as8}-\ref{as9}.
By $\omega_n$ definition, $\omega_n = O (l_n^{1/2} )$, so the slowest rate among (\ref{pt4-6})-(\ref{pt4-9}) is (\ref{pt4-9}). Then by (\ref{pt4-1})
\begin{eqnarray*}
\| \hat{\Sigma}_y^{-1} - \Sigma_y^{-1} \|_{l_2} &\le & \| (\hat{\Sigma}_u^{Th})^{-1} -  \Sigma_u^{-1} \|_{l_2} \\
&+& \| (\hat{\Sigma}_u^{Th})^{-1} \hat{\Sigma}^f \hat{G}  (\hat{\Sigma}_u^{Th})^{-1} - \Sigma_u^{-1} \Sigma^f G \Sigma_u^{-1} \|_{l_2} \\
& = & O_p (\omega_n s_n) + O_p ( J^2 \omega_n s_n) = O_p ( J^2 \omega_n s_n) = o_p (1),
\end{eqnarray*}
by Theorem \ref{thm3}, (\ref{pt4-9}).

{\hfill \bf Q.E.D.}\\

In the following, we show that under a slightly weaker condition, our framework covers the weak factor setting, which leads to a counterpart of Theorem \ref{thm4}. Based on the alternative weak factor Assumption \ref{as8_alt1}, we can replace Assumption \ref{as9} by the following weaker version

\renewcommand{\theassumwfactors}{W.6}
\begin{assumwfactors}\label{as9_alt1}
	Assume $\max ( J \omega_n s_n, \kappa_n^2 \omega_n s_n) \to 0$.
\end{assumwfactors}

\begin{corollary}\label{coll_wf1}
	Under Assumptions \ref{as1}-\ref{asa4}, Assumptions \ref{as8}\ref{as8_i}, \ref{as8_alt1} and \ref{as9_alt1}, we have 
	\[ \| \hat{\Sigma}_y^{-1} - \Sigma_y^{-1} \|_{l_2} = O_p (\kappa_n^2 \omega_n s_n) = o_p (1).\]
\end{corollary}

Proof of Corollary \ref{coll_wf1}. The results follows from the proof of Theorem \ref{thm4}, based on Lemma A.2 with the rate $r_n \omega_n s_n$, and Lemma A.4 (i) with the rate $r_n$. 

{\hfill \bf Q.E.D.}\\

Instead of using Assumption \ref{as8}\ref{as8_ii} or Assumption \ref{as8_alt1}, we can impose a very weak factor condition according to Assumption \ref{as8_alt2}. This allows us to use the following slightly weaker condition compared to Assumption \ref{as9}

\renewcommand{\theassumwfactors}{VW.6}
\begin{assumwfactors}\label{as9_alt2}
Assume $J \omega_n s_n \to 0$.
\end{assumwfactors}

This leads to the following corollary, 
\begin{corollary}\label{coll_wf2}
	Under \ref{as1}-\ref{asa4}, Assumptions \ref{as8}\ref{as8_i}, \ref{as8_alt2} and \ref{as9_alt2}, we obtain
	\[ \| \hat{\Sigma}_y^{-1} - \Sigma_y^{-1} \|_{l_2} = O_p (J \omega_n) = o_p (1).\]
\end{corollary}

Proof of Corollary \ref{coll_wf2}. This follows from Lemma A.2 with the rate $J  l_n^{1/2}$ since $r_n \to 0$, based on Assumption \ref{as8_alt2} and $\omega_n = O (l_n^{1/2})$, and Lemma A.4 (i) is with the rate $\max(J l_n^{1/2},r_n)$, and the proof of Theorem 4 follows from the slowest rate (\ref{pt4-7}) with the rate $J \omega_n$, since by definition $s_n < J$. 

{\hfill \bf Q.E.D.}\\

The case of $\kappa_n = O(1)$ can be handled in a similar way as in Corollary \ref{coll_wf2}, based on Assumption \ref{as9_alt2}, which leads to the same rate as in Corollary \ref{coll_wf2}.

\setcounter{equation}{0}
\setcounter{lemma}{0}
\setcounter{table}{0}
\renewcommand{\theequation}{B.\arabic{equation}}
\renewcommand{\thelemma}{B.\arabic{lemma}}
\renewcommand{\thetable}{\thesection.\arabic{table}}

\section{Non-linear Additive Factor Models}

This appendix helps us understand the non-linear additive factor model setup and provides its proofs.

We setup a general composite true function  and show how a non-linear but additive function be composite.

\subsection{General Composite Functions} \label{sec_b1_gcf}

We start with general composite functions to illustrate the general case, and then in the next subsection we analyze the subcase of non-linear additive functions.

In the following, we incorporate smoothness restrictions and a composite form to estimate the true function $f_{0,j}(.)$ by s-sparse deep neural networks. For each $j=1,\cdots, J$ 
\begin{equation}
 f_{0,j} (.) = g_{q,j}(.) \circ g_{q-1,j}(.) \cdots \circ g_{1,j}(.) \circ g_{0,j}(.),\label{compt}
 \end{equation}
where $f_{0,j}(.)$ is a composition of $q+1$ functions. Let $h=0,\cdots, q$, $g_{h,j}: [a_h, b_h]^{d_h} \to [a_{h+1}, b_{h+1}]^{d_{h+1}}$. 
Hence, each $g_{h,j}$ has $m=1,\cdots, d_h+1$ components:
\[ g_{h,j}= (g_{h1,j}, \cdots, g_{hm,j},\cdots, g_{hd_{h+1},j})'.\]

At each $g_{hm,j}$ we assume that it depends on maximal $t_h$ number of variables. In other words, each $g_{hm,j}$ is a $t_h$ variate function itself. To give an example, for any $j$ (suppressing $j$ only here for clarity), let $f_0 (x_1, x_2, x_3) = g_1(.) \circ g_0 (.)=g_{11} (g_{01} (x_1,x_3), g_{02} (x_2, x_3))$,
with $g_1 = g_{11}, d_2 =1$, $g_0 = (g_{01} (x_1, x_3), g_{02} (x_2, x_3))$ so $d_0=3, t_0=2, d_1=t_1=2,d_2=1$. Note that $d_0$ shows the dimension of the input variables. However, both $g_{01}, g_{02}$ depend on only 2 variables, hence $t_0=2$. We always demand that $t_h \le d_h$. This restriction is crucial, and is available in additive models which we thoroughly analyze in the subsequent sections. The previous example is taken from p. 1880 of \cite{sh2020}.

We impose the following smoothness restrictions on $g_{hm,j}$. Let $\beta_f$ denote the largest integer strictly smaller than $\beta$. A function has Hölder smoothness index $\beta$ if all partial derivatives up to order $\beta_f$ exist and are bounded, and the partial derivatives of order $\beta_f$ are $\beta - \beta_f$ Hölder. We impose each $g_{hm,j}$ has Hölder smoothness $\beta_h$. Also remember that $g_{hm,j}$ has $t_h$ variables, so each $g_{hm,j} \in {\cal C}_{t_h}^{\beta_h} ( [a_h, b_h]^{t_h}, K_h)$ and 
$ {\cal C}_{t_h}^{\beta_h} ( [a_h, b_h]^{t_h}, K_h)$ is the ball of $\beta_h$ Hölder functions with radius $K_h$ defined as
\begin{eqnarray*}
{\cal C}_{t_h}^{\beta_h} ( [a_h, b_h]^{t_h}, K_h) &:=& [ g_{hm,j}: [a_h, b_h]^{t_h} \to R: \\
&& \sum_{\alpha: \|\alpha \|_1 < \beta_h} \| \partial^{\alpha} g_{hm,j} \|_{\infty} 
+ \sum_{\alpha: \|\alpha\|_1= \beta_f }\sup_{ x,y \in [a_h, b_h]^{t_h}, x \neq y} \frac{\| \partial^{\alpha} g_{hm,j}(x) - \partial^{\alpha} g_{hm,j} (y)\|_1}{\|x-y\|_{\infty}^{\beta -
\beta_f}}\le K_h.]
\end{eqnarray*}

\noindent So we can have different flexible composite functions, where the smoothness will be the same for each $m=1,\cdots, d_{h+1},j=1,\cdots, J$, but differ across $h$.
We impose the following for the true function, $f_{0,j} \in {\cal G} (q, d, t, \beta, K)$ with 
\begin{eqnarray}
{\cal G} (q, d, t, \beta, K)&:=& \{ f_{0,j}= g_{q,j} \circ g_{h,j} \circ  \cdots \circ g_{0,j}: g_{h,j}=(g_{hm,j})_{m=1}
^{d_h +1}: [a_h, b_h]^{d_h} \to [a_{h+1}, b_{h+1}]^{d_{h+1}}\nonumber  \\
& & g_{hm,j} \in {\cal C}_{t_h}^{\beta_h} ( [a_h, b_h]^{t_h}, K), \, \text{for some} \quad |a_h |, |b_h| \le K \}.\label{smooth}
\end{eqnarray}
We define $d:=(d_0,\cdots, d_h,\cdots, d_{q+1})', t:=(t_0,\cdots, t_h, \cdots, t_q)$, $\beta:=( \beta_0, \cdots, \beta_h, \cdots, \beta_q)'$.
 There are two key terms to be defined: the effective smoothness index
\[ \beta_h^*:= \beta_h \, \Pi_{l=h+1}^q (\beta_l \wedge 1),\]
and  the rate which will be used in function approximation  error 
\noindent \[ \phi_n := \max_{h=0,\cdots, q} n^{\frac{-2\beta_h^*}{2 \beta_h^*+t_h}}.\]

\subsection{Non-linear Additive Functions} \label{sec_b2_naf}

Now we simplify above for additive non-linear functions. \textit{Section 4} of \cite{sh2020} obtains the following simplification.
Clearly in this case $d_0=d, t_0=1, d_1=t_1=d$, $d_2=1$ in (\ref{compt}). The dimension of $g_{0,j}(.)$ is $d$, however each scalar function is used in a sum of $d$ terms. See that 
$g_{0,j} : [0,1]^d \to R^d$, and $g_{1,j}(.) : R^d \to R$. Suppose that $f_{j,m}  \in {\cal C}_1^{\beta} ([0,1], K)$ for 
$m=1,\cdots, d$. It follows that
\[ f_{0,j}: [0,1]^d \stackrel{g_{0,j}}{\to}[-K,K]^d\stackrel{g_{1,j}}{\to}[-Kd, Kd].\]
For any $\gamma>1$, $g_{1,j} (X_i) \in {\cal C}_d^{\gamma} ([-K, K]^d, (K+1)d)$, and 
set the dimension of inputs $d,d,$ in each composite function $g_{0,j}, g_{1,j}$ and output dimension  1, as  $\tilde{d}:=(d,d,1)$ with $d_0=d, d_1=d, d_2=1$, and $\tilde{t}=(1,d)$.
Last, set the smoothness indicators for each composite as $\tilde{\beta}:=(\beta, (\beta \vee 2) d)$ for $g_0, g_1$, respectively.
So we assume $f_{0j} \in {\cal G} (1, \tilde{d}, \tilde{t}, \tilde{\beta}, (K+1) d)$, and the rate of  function approximation error 
\begin{equation}
\phi_n^*:= n^{-2\beta/(2 \beta+1)},\label{aer}
\end{equation}
since the summation function in composite function is infinitely smooth, so rate comes from $g_{0,j}$.

\subsection{Proofs for the Non-linear Additive Factor Model Setting} \label{sec_b3_proofs}

We start with the proof of Theorem \ref{thm1} here.

{\bf Proof of Theorem \ref{thm1}}.

(i).  The proof depends on two crucial steps. 
We now outline the  steps, and then show the proof. Step 1 is subdivided into two parts.
Step 1a will be our oracle inequality with subgaussian noise. Step 1a consists of Lemma \ref{l1}-\ref{l2}. Our Lemma \ref{l1} is the subgaussian counterpart of Supplement Lemma C.1 in \cite{sh2020}, and our Lemma \ref{l1a} provides a bound for the noise term which is described in (\ref{b1}) below. Lemma \ref{l1a} is the subgaussian counterpart of the inequality (II) on Supplement p.10 of \cite{sh2020}. Our Lemma \ref{l2} is the subgaussian counterpart of Lemma 4 in \cite{sh2020}. The proofs for gaussian noise in \cite{sh2020} do not carry over in a simple way to subgaussian case, hence we provide Lemma \ref{l1}-\ref{l2} here.
Step 1b provides an upper bound for covering numbers for the functions in the sparse network.
Step 2 is function approximation result that directly carries over from \cite{sh2020} with a minor extension.



Before the proof, we should note that the main technical issue is to obtain a bound for 
\begin{equation}
E \left[ \frac{2}{n} \sum_{i=1}^n  u_{j,i} \hat{f}_j (X_i)
\right].\label{b1}
\end{equation}

{\bf STEP 1a}.
 We will be interested in the following term which will be crucial for the bound in (\ref{b1}). For $q=1,\cdots, M_j$, $j=1,\cdots, J$
 \begin{equation}
\eta_{q,j}:= \frac{\sum_{i=1}^n u_{j,i} [ f_{q,j} (X_i) - f_{0,j} (X_i)]}{n^{1/2} \| f_{q,j} (X_i) - f_{0,j} (X_i) \|_n},\label{pt1-1}
\end{equation}
which is shown on p.13 of Supplement of \cite{sh2020} where $f_{q,j}(.)$ (i.e.,\ $f_{q,j}$ is not $f_{m,j}$, which is the part of the true model,  are the individual components of the sum in non-linear additive model) is defined as an approximation for the estimator, where 
\begin{equation}
 \| \hat{f}_j (X_i) - f_{q,j} (X_i) \|_{\infty} \le \delta ,\label{pt1-1a}
 \end{equation}
for $j=1,\cdots, J, q=1,\cdots, M_j$, with $\delta >0$. For our purposes, we will rewrite (\ref{pt1-1}) as 
\begin{equation}
\eta_{q,j} = \frac{ \sum_{i=1}^n u_{j,i}[ f_{q,j} (X_i) - f_{0,j} (X_i)]}{\sqrt{\sum_{i=1}^n (f_{q,j} (X_i) - f_{0,j} (X_i))^2}} = \sum_{i=1}^n u_{j,i} \Delta_{q,j} (X_i),\label{pt1-2}
\end{equation}
where 
\[ \Delta_{q,j} (X_i) := \frac{ f_{q,j} (X_i) - f_{0,j} (X_i)}{\sqrt{\sum_{i=1}^n [ f_{q,j} (X_i) - f_{0,j} (X_i)]^2}},\]
and set a $n \times 1$ vector, for each $q,j$ 
\[ \Delta_{q,j} (X):= (\Delta_{q,j} (X_1), \cdots, \Delta_{q,j} (X_n))'.\]
See that, for  each $q,j$ 
\begin{eqnarray}
\| \Delta_{q,j} (X) \|_2^2 &=& \sum_{i=1}^n \Delta_{q,j} (X_i)^2 \\
& = & \frac{ \sum_{i=1}^n [ f_{q,j} (X_i) - f_{0,j} (X_i)]^2}{\sum_{i=1}^n [ f_{q,j} (X_i) - f_{0,j} (X_i)]^2} =1.\label{pt1-3} 
\end{eqnarray}
Note that given $u_{j,i}$ as zero-mean, with variance 1, and iid subgaussian errors, which are independent of $X_i$ across $i=1,\cdots, n$. By that information
$E \eta_{q,j} = 0$, $var \eta_{q,j} =1$, for each $q=1, \cdots, M_j$. Set the Orlicz norm for the errors
\[ \max_{1 \le j \le J } \| u_{j,i} \|_{\psi_2} = C_{\psi} < \infty.\]
Now we state the following Lemma. Lemma \ref{l1} is for sub-gaussian noise, and 
  extends from Gaussian error-Lemma C.1 in \cite{sh2020}. 

\begin{lemma}\label{l1}

Under Assumptions \ref{asa1},\ref{as4}, for each $j=1,\cdots, J$
\[E \max_{1 \le q \le M_j } \eta_{q,j}^2 \le C_{s1} \log M_j + 2,\]
\end{lemma}
with $C_{s1} := \max (3, C_{\psi}^2/c ), c >0,$.

{\bf Proof of Lemma \ref{l1}}. Define $Z_j := \max_{1 \le q \le M_j} \eta_{q,j}^2$, then we have, for a given j, 
\begin{equation}
Z_j \le \sum_{q=1}^{M_j} \eta_{q,j}^2,\label{pt1-4}
\end{equation}
and 
\begin{equation}
E Z_j \le \sum_{q=1}^{M_j} E \eta_{q,j}^2 = M_j,\label{pt1-5} 
\end{equation}
via zero mean and unit variance for $\eta_{q,j}$. We show the proof for $M_j \ge 4$. The proof for $M_j \le 3$ is a simple algebraic inequality which does not use the distribution for $u_{j,i}$, and is shown in \cite{sh2020}, also will be shown at the end of the proof here. Note that  for $t>0$
\begin{equation*}
P [ \eta_{1,j}^2 \ge t] = P [ | \eta_{1,j} | \ge \sqrt{t}] = 2 P [ \eta_{1,j} \ge \sqrt{t}].
\end{equation*}
For any $T_j > 0$
\begin{eqnarray}
E Z_j & = & \int_0^{\infty} P [ Z_j \ge t ] dt \le T_j + \int_{T_j}^{\infty} P ( Z_j \ge t ) dt \nonumber \\
& \le & T_j + M_j \int_{T_j}^{\infty} P [ \eta_{1,j}^2 \ge t ] dt,\label{pt1-6}  
\end{eqnarray}
where we use Lemma 1.2.1-Integral identity of  \cite{v2019}, since $Z_j$ is nonnegative for the equality in (\ref{pt1-6}), and we use union bound for the last inequality and (\ref{pt1-4}). In (\ref{pt1-6}) consider by $\eta_{1,j}$ definition in (\ref{pt1-2})
\begin{equation}
\int_{T_j}^{\infty} P [ \eta_{1,j}^2 \ge t ] dt = \int_{T_j}^{\infty} P [ | \eta_{1,j} | \ge \sqrt{t} ] dt = \int_{T_j}^{\infty} P [ | \sum_{i=1}^n u_{j,i} \Delta_{1,j} (X_i) | \ge \sqrt{t}] dt.\label{pt1-7}
\end{equation}

Now we use General Hoeffding inequality, Theorem 2.6.3 of \cite{v2019}, since conditional on $X_i$ $u_{j,i} \Delta_{1,j} (X_i)$ is subgaussian, and $u_{j,i}$ is independent from $\Delta_{1,j} (X_i)$ across $i$, for a given $j$, and    for a positive constant $c>0$,

\begin{equation*}
P \left[ | \sum_{i=1}^n u_{j,i} \Delta_{1,j} (X_i) | \ge \sqrt{t}\right]  \le  2\exp \left(  \frac{-c t }{C_{\psi}^2 \| \Delta_{1,j} (X_i) \|_2^2} \right)
= 2 \exp \left(\frac{-c t }{C_{\psi}^2}\right),
\end{equation*}
by (\ref{pt1-3}) for the last equality. Now substitute this last inequality  in (\ref{pt1-6})
\begin{eqnarray}
E Z_j & \le & T_j + 2 M_j \int_{T_j}^{\infty} \exp \left( \frac{-c t }{C_{\psi}^2}
\right) dt \nonumber \\
& = & T_j + 2M_j \frac{C_{\psi}^2}{c} \exp \left( \frac{-c T_j }{C_{\psi}^2} 
\right) \nonumber \\
& = & \left( \frac{C_{\psi}^2}{c} 
\right) \log M_j + 2 M_j \frac{C_{\psi}^2}{c} \frac{1}{M_j} = \frac{C_{\psi}^2}{c} (\log M_j + 2),\label{pt1-8}
\end{eqnarray}
where we use $T_j = \frac{C_{\psi}^2}{c} \log M_j $ for the second equality. Combine (\ref{pt1-8}) with case of $1 \le M_j \le 3$ in p.14, proof of Lemma C.1 in Supplement of \cite{sh2020} which is $E Z_j \le M_j \le 3 \log M_j + 1$ to have the desired result.

{\hfill \bf Q.E.D.}\\

Now we provide another lemma, that will bound the noise given Lemma \ref{l1}. This is extension of the noise bound (II) on p.10 of Supplement of \cite{sh2020}, from gaussian error to subgaussian error. Note that we set $M_j= N_{n,j} (\delta, {\cal F}_j, \|.\|_{\infty})$ which are the covering numbers for the s-sparse deep network ${\cal F}_j$.
We condition on $\log M_j  \le n$. Case of $\log M_j \ge n$ will be discussed at the end of proof of Theorem \ref{thm1}.
 First, we set the in-sample prediction error as:
\begin{equation}
\hat{R}_n (\hat{f}_j (X_i) , f_{0,j} (X_i) ):= E \left[\frac{1}{n} \sum_{i=1}^n (\hat{f}_{j}(X_i) - f_{0,j} (X_i))^2\right].\label{eerror}
\end{equation}
The following noise bound is the extension of (II) in p.10 of Supplement of \cite{sh2020}, from gaussian noise to subgaussian noise. 
The proof technique for Lemma \ref{l1a} is the same as in \cite{sh2020}, given our new Lemma \ref{l1}.
To simplify the notation set $N_{n,j}:= N_{n,j} (\delta, {\cal F}_j, \| .\|_{\infty})$,

\begin{lemma}\label{l1a}
Under Assumptions \ref{asa1}, \ref{as4}, with $\log N_{n,j} \le n $, for all $j=1,\cdots, J$
\[\left|  E \left[ \frac{2}{n} u_{j,i} \hat{f}_j (X_i) 
\right]
\right|  \le  (2 + 2 \sqrt{C_{s1} + 2} ) \delta + 2 \sqrt{ \frac{\hat{R}_n ( \hat{f}_j (X_i), f_{0,j} (X_i)) (C_{s1} \log N_{n,j} + 2)}{n}}.\]
\end{lemma}

{\bf Proof of Lemma \ref{l1a}}. Note that by (C.5) of \cite{sh2020} we get the first inequality below
\begin{eqnarray}
\left|  E \left[ \frac{2}{n} u_{j,i} \hat{f}_j (X_i) 
\right]
\right| & \le & 2 \delta + \frac{2}{n^{1/2}} E \left[ ( \| \hat{f}_j (X_i) - f_{0,j} (X_i)  \|_n + \delta ) \| \eta_{m,j}| 
\right] \nonumber \\
& \le & 
2 \delta + \frac{2}{n^{1/2}} (\hat{R}_n (\hat{f}_j (X_i), f_{0,j} (X_i))^{1/2} + \delta) \sqrt{C_{s1} \log N_{n,j} + 2},\label{pl1a-1} 
\end{eqnarray}
where for the second inequality we use Cauchy-Schwartz inequality, with $E \eta_{q,j}^2 \le E \max_{1 \le q \le M_j}
\eta_{q,j}^2 $, and Lemma \ref{l1}. Then note that since we assume $\log N_{n,j} \le n$, 
\begin{equation}
\frac{2}{n^{1/2}} \delta \sqrt{ C_{s1} \log N_{n,j} + 2} \le 2 \delta  \sqrt{C_{s1} + 2} .\label{pl1a-2}
\end{equation}
Then  by (\ref{pl1a-1})(\ref{pl1a-2})

\begin{equation}
\left|  E \left[ \frac{2}{n} u_{j,i} \hat{f}_j (X_i) 
\right]
\right|  \le  (2 + 2 \sqrt{C_{s1} + 2} ) \delta + 2 \sqrt{ \frac{\hat{R}_n ( \hat{f}_j (X_i), f_{0,j} (X_i)) (C_{s1} \log N_{n,j} + 2)}{n}}.\label{pl1a-3}
\end{equation}

{\hfill \bf Q.E.D.}\\

Now we provide an oracle inequality which is the extension  of Lemma 4 of \cite{sh2020} from Gaussian noise to subgaussian noise. Also, we extend the result to maximum of $J$ functions.

\begin{lemma}\label{l2}
Under Assumptions \ref{asa1},\ref{as4}, and
 \[ \{ f_{0,j}(.) \} \cup {\cal F}_j \subset \{f_j(.): [0,1]^d \to [-F, F]\}\,  {\mbox for some}\, F \ge 1,\]
  with covering numbers
\[ {\cal N}_{n,j}  \ge 3 \quad  {\mbox for \, all}\,  \delta \in (0,1],\]
 then  
 for each $j=1,\cdots, J$
\[ \max_{1 \le j \le j } R (\hat{f}_j (X_i) , f_{0,j} (X_i)) \le 4 \left\{ \left[ \max_{1 \le j \le J} inf_{f_{0,j} \in {\cal F}_j} E [ (f_j(X_i) - f_{0,j} (X_i))^2] \right] + 
\frac{4 F^2 }{n } [C_{s2} \max_{1 \le  j \le J}\log N_{n,j} + 74] + C_{s3} \delta F 
\right\},\]
for constants $C_{s2} \ge 18, C_{s3} \ge 84$. These last two constants are derived from the constant $C_{s1} \ge 3$, and explained in the proof. 
\end{lemma}

Remark. Note that in subgaussian case of Lemma 4 of \cite{sh2020},  he has 18 in front of the covering numbers, since $C_1 \ge 3$ we have $C_{s2} := 2 C_{s1} + 12 \ge 18$. Our second constant  in the second term is $74$ and larger than $72$ in Lemma 4 of \cite{sh2020}, so there is a price to pay for a more general result in our case, but these differences will not matter when $n \to \infty$. Also in front of $\delta F$, Lemma 4 of \cite{sh2020} has 32 as a constant, and we have at least 84. 
Clearly the Orlicz norm plays a key role in our bound, since $C_{s1} = \max (3, \frac{C_{\psi}^2}{c})$, and $C_{\psi}$ represents maximum, across iid observations, of the Orlicz norm ($\psi_2$) of the noise.

{\bf Proof of Lemma \ref{l2}}. Our proof of Lemma
\ref{l2} has the structure of (III) of p.11 on Supplement  of \cite{sh2020}. The main difference is our Lemma \ref{l1}-\ref{l1a} here.
We start with the case of $\log N_{n,j} \le n$, for $j=1,\cdots, J$. The other case will be discussed at the end. 
By $\hat{f}_j(.)$ definition
\begin{equation}
E \left[ \frac{1}{n} \sum_{i=1}^n (Y_{j,i} - \hat{f}_j (X_i))^2
 \right] \le E \left[ \frac{1}{n} \sum_{i=1}^n (Y_{j,i} - f_j (X_i))^2
 \right].\label{pl2-1} \end{equation}
 
 Also by assumption
 \begin{equation}
 E  u_{j,i} f_j (X_i) =0.\label{pl2-3}
 \end{equation}

By the model, $Y_{j,i}$ in (\ref{mo})
\[ E \left[ \frac{1}{n}  \sum_{i=1}^n (u_{j,i} - (\hat{f}_j (X_i) - f_{0,j} (X_i) ))^2 \right] \le 
E \left[ \frac{1}{n}  \sum_{i=1}^n (u_{j,i} - (f_j (X_i) - f_{0,j} (X_i) ))^2 \right] 
.\]
Then rearranging simply and using (\ref{eerror})(\ref{pl2-3}), iid nature of $X_i$
\begin{equation}
\hat{R}_n ( \hat{f}_j (X_i), f_{0,j} (X_i) ) \le E \left[ \frac{2}{n} \sum_{i=1}^n u_{j,i} \hat{f}_j (X_i) 
\right] + E [ f_j (X_i) - f_{0,j} (X_i)]^2.\label{pl2-4}
\end{equation}
By Lemma \ref{l1a}
\begin{eqnarray*}
\hat{R}_n (\hat{f}_j (X_i), f_{0,j} (X_i)) & \le & E [ f_j (X_i) - f_{0,j} (X_i)]^2 \nonumber \\
& + &(2 + 2 \sqrt{C_{s1} + 2} ) \delta + 2 \sqrt{ \frac{\hat{R}_n ( \hat{f}_j (X_i), f_{0,j} (X_i)) (C_{s1} \log N_{n,j} + 2)}{n}}.
\end{eqnarray*}
For positive real numbers, $a,b,d$
set 
\[a = \hat{R}_n (\hat{f}_j (X_i), f_{0,j} (X_i)), \, b = \sqrt{(C_{s1} \log N_{n,j} + 2 )/n},\]
 
\[ d =   E [ f_j (X_i) - f_{0,j} (X_i)]^2 
 + (2 + 2 \sqrt{C_{s1} + 2} ) \delta.\]
 
  If $|a | \le 2 \sqrt{a}b + d $ then $  a \le 2 d + 4b^2$
 which implies, since $f_j$ can also be the minimizer, and $F \ge 1$
 \begin{equation}
 \hat{R}_n (\hat{f}_j (X_i), f_{0,j} (X_i)) \le   2 \left[ inf_{f_j \in {\cal F}_j} E [ f_j (X_i) - f_{0,j} (X_i)]^2  \right]
 + (2 + 2 \sqrt{C_{s1} + 2} ) 2 \delta
 + F^2 \frac{4 ( C_{s1} \log N_{n,j} + 2)}{n}.\label{pl2-5}
  \end{equation}
  
  Next, we use the upper bound in the inequality (I) on p. 10 of Supplement of \cite{sh2020}, which  uses only independence of the data $Y_i, X_i$ across $i$, but relates the risk to its empirical version
  \[ R (\hat{f}_j (X_i), f_{0,j} (X_i) ) \le 2 \left[ \hat{R}_n (\hat{f}_j (X_i), f_{0,j} (X_i)) + \frac{2 F^2}{n} (12 \log N_{n,j} + 70) + 26 \delta F\right].\]

Taking maximum over $j=1,\cdots , J$ and apply also (\ref{pl2-5})
\begin{eqnarray*}
 \max_{1 \le j \le J} R (\hat{f}_j (X_i), f_{0,j} (X_i)) &\le& 4 [ \max_{1 \le j \le J} \inf_{f_j \in {\cal F}_j} E [ f_j (X_i) - f_{0,j} (X_i)]^2] \\
 &+& [ 52 + (8 + 8 \sqrt{C_{s1} +2})] \delta F + 4 F^2 \frac{ (12 + 2 C_{s1}) \max_{1 \le j \le J} \log N_{n,j} + 74}{n}.
 \end{eqnarray*}

Set $C_{s2}:= 12 + 2 C_{s1}, C_{s3} \ge  84 \ge 52 + (8 + 8 \sqrt{C_1 +2})$, with $C_{s1} \ge 3$. Also we have an empirical minimizer as estimator so $\Delta_n=0$ in the proofs of \cite{sh2020}.

Case of $\log N_{n,j} \ge n$ is shown in proof of Lemma 4 in Supplement of \cite{sh2020}. As  
\[  \max_{1 \le j \le J} R (\hat{f}_j (X_i), f_{0,j} (X_i)) \le 4 F^2,\]
Lemma B.3 bound also holds for $\log N_{n,j} \ge n$.

{\hfill \bf Q.E.D.}\\

{\bf Step 1b}.  Next, we need to find upper bounds on the covering numbers. First, Lemma 5 of \cite{sh2020} with Remark 1 and proof of Theorem 2 in \cite{sh2020} puts an upper bound on $\log N_{n,j}$ in our  Lemma \ref{l2} and it is not related to $u_{j,i}$ data distribution. Let $C>0$ be  a positive constant
\begin{eqnarray*}
\max_{1 \le j \le J} \log N_{n,j} & \le &  C \max_{1 \le j \le J} \left[ (s_j +1) \log ( n (s_j +1)^L d)
\right] \\
& = & (\bar{s} + 1) \log ( n (\bar{s}+1)^L d ),
\end{eqnarray*}
where we use $p_0=d, p_{L+1}=1$ in our notation, where $p_0$ is the input dimension, and $p_{L+1}$ is the output dimension in \cite{sh2020}. We extend the result of \cite{sh2020} to uniform over $j=1,\cdots, J$ functions.  Then 
\begin{eqnarray}
\frac{\max_{1 \le j \le J} \log N_{n,j}}{n} & \le &\frac{ C (\bar{s} + 1) \log [ n (\bar{s}+1)^L d ]}{n}  =   \frac{C (\bar{s} + 1) \log [ \frac{n^L}{n^{L-1}} (\bar{s}+1)^L \frac{d^L}{d^{L-1}} ]}{n} \nonumber \\
& = & \frac{C (\bar{s} + 1) L  \log [ \frac{n}{n^{1-1/L}} (\bar{s}+1) \frac{d}{d^{1-1/L}} ]}{n}.\label{pt1-11}
\end{eqnarray}
Then since  Assumption  \ref{as5}, $\bar{s} \le C n \phi_n^* \log n$ we have 
\begin{equation}
\bar{s} +1 \le  C' n \phi_n^* \log n.\label{pt1-12}
\end{equation}
where $C'$ is another positive constant. Next, we have by the definition of $\phi_n^*:= n^{-2\beta/(2 \beta +1)}$ in (\ref{aer}) 
\[ n^{\frac{1}{L}} \phi_n^* \log n = n^{\frac{1}{L}} n^{-2\beta/(2 \beta + 1)} \log n \to 0,\]
by Assumption  \ref{as5}, $L \ge C \log_2 (n)$. Since $d$ is constant, with sufficiently large n, and using (\ref{pt1-12}) and the last result above 
\begin{equation}
 \log [ \frac{n}{n^{1-1/L}} (\bar{s}+1) \frac{d}{d^{1-1/L}} ]\le \log [ (n) C' n^{1/L} d^{1/L} \phi_n^* \log n  ] \le \log n.\label{pt1-13}
\end{equation}
So substitute (\ref{pt1-12})-(\ref{pt1-13}) into (\ref{pt1-11}), with sufficiently large n 
\begin{equation}
\frac{\max_{1 \le j \le J} \log N_{n,j}}{n} \le C' L \phi_n^* \log^2n.\label{pt1-14}
\end{equation}

{\bf Step 2}.
Next, we consider function approximation. Since 
\[ \inf_{f_j \in {\cal F}_j} E [ f_j (X_i) - f_{0,j} (X_i)]^2] \le \inf_{f_j \in {\cal F}_j} \| f_j (X_i) - f_{0,j} (X_i) \|_{\infty}^2,\]
by (26) of \cite{sh2020}, defining $C'>0$ a positive constant
\[ \inf_{f_j \in {\cal F}_j} \| f_j (X) - f_0 (X) \|_{\infty}^2 \le C' \phi_n^*,\]
we get 
\begin{equation}
\max_{1 \le j \le J} \inf_{f_j \in {\cal F}_j} \| f_j (X_i) - f_0 (X_i) \|_{\infty}^2 \le C' \phi_n^*,\label{pt1-15}
\end{equation}
since right hand side of the previous inequality, the one before (\ref{pt1-15}) does not depend on $j$.
Now combine (\ref{pt1-14})(\ref{pt1-15}) in Lemma \ref{l2} to have the desired result, with $\delta=1/n$, and $F \ge 1$ and function have same smoothness parameters  $\beta$ regardless of $j=1,\cdots, J$.

Apply (\ref{pt1-11})-(\ref{pt1-15}) to the right side of (\ref{pl2-5}) to get, by using $X_i$ being iid, for sufficiently large n  
\[ \max_{1 \le j \le J } E \left[\frac{1}{n} \sum_{i=1}^n [ \hat{f}_j (X_i) - f_{0,j} (X_i) ]^2 \right] = 
\max_{1 \le j \le J } E \left[ \hat{f}_j (X_i) - f_{0,j} (X_i) ]^2 \right]
\le C \phi_n^* L \log^2 n.\]

{\hfill \bf Q.E.D.}
\\

(ii). This is not a trivial extension from Theorem \ref{thm1}(i) since we have $J$ functions to estimate, and it is not obvious which concentration inequality and how to use them in the proof. 
 Here our maximal inequality result in (\ref{pt2-4}) can also apply  to other estimation problems in high dimensions.
 
  Use Theorem \ref{thm1}, Remark 1 to have 
 \begin{equation}
 \max_{1 \le j \le J} \frac{1}{n} \sum_{i=1}^n  E { [ \hat{f}_j (X_i) - f_{0,j} (X_i)]^2 } \le C n^{-2\beta/(2 \beta+1)} \log^3 n.\label{pt2-5}
  \end{equation}
  
   Since for all $i=1,\cdots, n, j=1, \cdots , J $
 \[ | \hat{f}_j (X_i) - f_{0,j} (X_i) | \le 2 F,\]
 we have by (2.11) of \cite{w2019} or one-sided version of p.454 of \cite{w2019}, with $t>0$
 \[ P \left( \frac{1}{n}  \sum_{i=1}^n [ \hat{f}_j (X_i) - f_{0,j} (X_i)]^2 - E { [ \hat{f}_j (X_i) - f_{0,j} (X_i)]^2 } \ge t 
  \right) \le \exp(\frac{-n t^2}{32 F^4}).\]
  Using the union bound, with $t_1>0$
   \begin{equation}
    P \left(  \max_{1 \le j \le J} \left[ \frac{1}{n}  \sum_{i=1}^n [ \hat{f}_j (X_i) - f_{0,j} (X_i)]^2 - E { [ \hat{f}_j (X_i) - f_{0,j} (X_i)]^2 }\right] \ge t_1 
  \right) \le \exp(\log J - \frac{-n t_1^2}{32 F^4}).\label{pt2-1}
  \end{equation} 
  Choose $t_1 = (\sqrt{32} F^2) \sqrt{t^2 + \log J/n}$ with choice of $t= c_1 \sqrt{ \log J/n}$ in (\ref{pt2-1})
 \begin{equation}
  P \left(  \max_{1 \le j \le J} \left[ \frac{1}{n}  \sum_{i=1}^n [ \hat{f}_j (X_i) - f_{0,j} (X_i)]^2 - E { [ \hat{f}_j (X_i) - f_{0,j} (X_i)]^2 }\right] \ge 
 (\sqrt{32} F^2) \sqrt{ (c_1^2 +1) \log J/n} \right) \le \frac{1}{J^{c_1^2}}.\label{pt2-2}
 \end{equation}
 Then note that by defining $Z_j:= \sum_{i=1}^n [ \hat{f}_j (X_i) - f_{0,j} (X_i)]^2, Z_j \ge 0$, we see that 
 \begin{equation}
 \max_{1 \le j \le J } [ Z_j - E Z_j ] \ge \max_{1 \le j \le J} [ Z_j - \max_{1 \le j \le J} E Z_j] = \max_{1 \le j \le J} Z_j - \max_{1 \le j \le J} E Z_j.\label{pt2-3}
 \end{equation} 
 (\ref{pt2-3}) implies that we can change the left side term of the probability in  (\ref{pt2-2}) and get the following maximal inequality
 \begin{equation}
  P \left(  \max_{1 \le j \le J}  \frac{1}{n}  \sum_{i=1}^n [ \hat{f}_j (X_i) - f_{0,j} (X_i)]^2 - \max_{1 \le j \le J} \frac{1}{n} \sum_{i=1}^n  E { [ \hat{f}_j (X_i) - f_{0,j} (X_i)]^2 } \ge 
 (\sqrt{32} F^2) \sqrt{(c_1^2 +1) \log J/n} \right) \le \frac{1}{J^{c_1^2}}.\label{pt2-4}
 \end{equation} 
 
 So, use (\ref{pt2-5}) in the second term on the left side of the probability in (\ref{pt2-4}) with $r_{n3}:= C n^{-2\beta/(2 \beta+1)} \log^3 n, r_{n4}:=\sqrt{32} F^2 \sqrt{(c_1^2 +1) \log J/n}$ definitions
 \[ P [ \max_{1 \le j \le J} \frac{1}{n} \sum_{i=1}^n [ \hat{f}_j (X_i) - f_{0,j} (X_i)]^2  \ge  r_{n3} + r_{n4}] \le \frac{1}{J^{c_1^2}}.\] 

{\hfill \bf Q.E.D.}\\

\bibliographystyle{chicagoa}
\bibliography{dlearnfin-2}
\end{document}